\documentclass[11pt]{article}
\usepackage[preprint]{acl}
\usepackage{times}
\usepackage{latexsym}
\usepackage{enumitem}
\usepackage{xcolor,colortbl}
\usepackage{pgf}
\usepackage{booktabs}
\usepackage{times}
\usepackage{bbm}
\usepackage{latexsym}
\usepackage{hyperref}
\usepackage{amsmath}
\usepackage{amssymb}
\usepackage{placeins}
\usepackage{nameref}
\usepackage{pdflscape}
\usepackage{lipsum}
\usepackage{xstring}
\usepackage{subcaption}
\usepackage{multirow}
\usepackage{pdfpages} 
\usepackage{makecell}
\usepackage{longtable}
\usepackage{tikz}
\usetikzlibrary{positioning, arrows.meta,calc}
\usepackage{fancyvrb}
\usepackage{fvextra} 
\usepackage{hyperref}
\usepackage{adjustbox}
\usepackage{pgf}        
\usepackage{graphicx}   

\usepackage{booktabs}
\usepackage[T1]{fontenc}
\usepackage{amsfonts}
\usepackage{amsmath} 
\usepackage{silence}
\usepackage{microtype}
\usepackage{inconsolata}
\usepackage{graphicx}
\usepackage{fancyhdr}

\usepackage{enumitem}
\usepackage{booktabs}
\newcommand{\hfciteprocess}[1]{%
  \IfSubStr{#1}{https://huggingface.co/datasets/}{%
    \StrBehind{#1}{https://huggingface.co/datasets/}[\hfname]%
  }{%
    \StrBehind{#1}{https://huggingface.co/}[\hfname]%
  }%
  \href{#1}{\hfname}%
}
\newcommand{\hfcite}[2]{%
  (%
  \begingroup
    \def\hf@first{1}%
    \renewcommand*{\do}[1]{%
      \ifnum\hf@first=1
        \hfciteprocess{##1}%
        \def\hf@first{0}%
      \else
        , \hfciteprocess{##1}%
      \fi
    }%
    \docsvlist{#1}%
  \endgroup:~%
  \citealp{#2})%
}

\newcommand{\papertitle}{On Representational and Functional Alignment of Training Data Attribution Methods for Language Models}

\title{\papertitle}
\author{
 \textbf{Loris Schoenegger\textsuperscript{1,2}},
 \textbf{Benjamin Roth\textsuperscript{1,3}}
\\
\\
 \textsuperscript{1}Faculty of Computer Science, University of Vienna, Vienna, Austria\\
 \textsuperscript{2}UniVie Doctoral School Computer Science, University of Vienna, Vienna, Austria\\
  \textsuperscript{3}Faculty of Philological and Cultural Studies, University of Vienna, Vienna, Austria\\
\\
 \small{
   \textbf{Correspondence:} \href{mailto:loris.schoenegger@univie.ac.at}{loris.schoenegger@univie.ac.at}
 }
}

\begin{document}
\maketitle
\begin{abstract}
Existing approaches to training data attribution derive influence estimates from different sources of information, including gradients, model-internal states, and surface-level similarity.
However, comparisons between attribution methods remain incomplete, often relying on indirect evaluations through downstream task performance.
In this work, we introduce a framework for directly comparing methods along three dimensions: \textit{functional alignment}, which measures correlation between full attribution vectors; \textit{representational alignment}, which captures structural similarity of the representations underlying attribution; and \textit{cross-method predictability}, which evaluates agreement on local pairwise ranking decisions.
We observe strong representational and moderate functional alignment within paradigms. Alignment across paradigms is low and task-dependent, yet stable across models.
Our results suggest that the similar downstream performance observed across paradigms in prior work can be attributed to agreement on the relative importance of documents, rather than to agreement on rankings of the full training data.
\end{abstract}

\section{Introduction}\label{introduction}
Training data attribution methods assign documents from a language model'’s training data scores proportional to their contribution to a given generation, to provide insights into model predictions.
Several \textbf{attribution paradigms} have proven popular for language models, differing in the source of information they base attribution on:
Some methods leverage \textit{loss gradients} \cite[e.g.,][]{pruthi_estimating_2020,koh_understanding_2017}, others \textit{model-internal states} \cite[e.g.,][]{li_influence_2025}, yet others only \textit{lexical overlap} \cite[e.g.,][]{liu_olmotrace_2025}, assuming this still provides useful signals for understanding model behavior. 

These paradigms follow fundamentally different attribution rationales (gradient-based methods estimate the effect of including a given training example on a test prediction through a first-order approximation, whereas representation-based methods assume that an example is influential if it is close to said test prediction in some feature space; \citealp{sun_enhancing_2026}), with empirical work reporting limited rank correlation between certain paradigms \cite{pezeshkpour_empirical_2021}. Nevertheless, prior work has demonstrated that cheap representation-based methods can serve as proxies for expensive gradient-based ones for tasks such as fact tracing 
or artifact detection  \cite{akyurek_towards_2022,pezeshkpour_empirical_2021,sun_enhancing_2026,liu_olmotrace_2025}.

\begin{figure}[htb]
    \centering
   \includegraphics[
  width=0.75\columnwidth,
  trim=0 0.1cm 0 0.1cm,
  clip
]{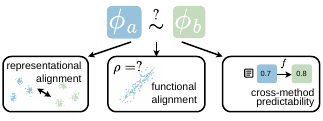}
    \caption{\textbf{Analytical Framework.} We directly compare attribution strategies by decomposing alignment into (1) similarity of underlying representations, (2) correlation of attributions, and (3) agreement on pairwise rankings.}
    \label{fig:setup}
\end{figure}

\noindent In this work, we investigate \textit{how well methods from different training data attribution paradigms align in their assessments of data influence}.
Prior work has largely neglected comparisons across paradigms and, when conducted, has compared solely in terms of downstream task performance \cite[e.g.,][]{jiao_date-lm_2025} or by the rank correlation between importance rankings over training instances \cite[e.g.,][]{pezeshkpour_empirical_2021}.
We argue that this generates an incomplete view of alignment and propose a novel evaluation framework that addresses three main limitations:
First, existing comparisons between attribution methods \cite{pezeshkpour_empirical_2021,brophy_adapting_2024,nguyen_bayesian_2023} depend heavily on evaluation task mechanics, limiting their transferability. For example, methods that work well for question answering may fail to capture aspects relevant to instruction tuning because of differences in learning objectives.
Second, these works only consider 
global measures of correlation between attributions (e.g., Spearman's $\rho$ between rankings induced by two attribution score vectors $\in \mathbb{R}^{|D|}$), which does not adequately reflect functional equivalence in many downstream applications such as data selection, where agreement on relative relevance judgments is sufficient (i.e., whether a given example is more influential than another).
Third, existing studies do not separate representational structure from downstream ranking decisions, making it difficult to determine whether differences between methods stem from how information is represented or from how it is processed.
We accordingly \textbf{decompose alignment into three dimensions}: \textit{functional alignment}, i.e., correlation between two full attribution vectors; \textit{representational alignment}, i.e., structural similarity of the underlying document representations used for attribution; and \textit{cross-method predictability}, i.e., agreement in pairwise relevance predictions rather than correlation between two rankings of the entire training dataset.
These measures provide complementary perspectives on alignment:
Functional alignment and cross-method predictability may serve to establish that two methods produce sufficiently similar importance rankings over training instances, while measures of representational alignment can generate insights into structural similarities of the  document representations used for attribution.

We leverage this framework to conduct a \textbf{systematic study of alignment} of 12 attribution methods across four paradigms, 5 tasks, and 7 LoRA-finetuned models spanning three model families.\footnote{We release our code at \href{https://doi.org/10.5281/zenodo.24214505}{doi.org/10.5281/zenodo.24214505}.} 
Specifically, \textbf{we provide insights into}:
(i) the alignment of attribution methods across different language modeling tasks, (ii) the stability of alignment across datasets and base models, (iii) whether methods can be employed as proxies for others, and (iv) factors that impact alignment at the instance level.
We find that methods within the same attribution paradigm tend to exhibit high representational but moderate functional alignment. Alignment across paradigms is overall low and highly task-dependent, but consistent across model families and sizes.
Our findings suggest that the relatively consistent downstream performance across paradigms in prior work may be driven by agreement on the relative importance of individual examples, rather than by agreement on example relevance across the full training data.
In a qualitative analysis, we find that cross-paradigm alignment appears to be shaped by a mix of prompting artifacts, lexical overlap, and task-dependent structural and behavioral similarities.
Overall, our results suggest that methods within a paradigm are partially interchangeable, but that different paradigms capture distinct, only partially aligned notions of data influence, suggesting that only a combination of different sources of information may yield a complete characterization of data influence.
Our main contributions are:
\begin{enumerate}[nosep, topsep=0pt, label=(\arabic*), left=0pt]
    \item A framework for measuring representational and functional alignment; and a test of cross-method predictability that assesses local rather than overall agreement on example influence.
    \item A systematic study of alignment across 7 base models, 5 tasks, and 12 attribution methods.
    \item A qualitative analysis of factors of alignment.
\end{enumerate}

\section{Related Work}
Works on \textit{input feature}- \cite[e.g.][]{jain_attention_2019, neely_song_2022, sun_evaluating_2025} and \textit{neuron} attribution \cite[e.g.][]{yu_revealing_2024} have compared methods in terms of vector similarity and explanation quality metrics. 
However, their findings are not directly applicable here because they target fundamentally different attribution mechanisms.
\paragraph{Indirect evaluation of alignment.}
Prior comparisons of \textit{training data} attribution methods are conceptual surveys without empirical evaluation \cite{hammoudeh_training_2024,deng_survey_2025}, or predominantly compare via benchmark performance on tasks such as hallucination tracing \cite{wu_enhancing_2024}, fact tracing \cite{akyurek_towards_2022}, or data cleaning \cite{pezeshkpour_empirical_2021,jiao_feasibility_2025,yu_revealing_2024,han_explaining_2020}.
Work on example-based explanation compares in terms of application-specific properties such as stability \cite{gu_iaeval_2023}, faithfulness, or plausibility \cite{hanawa_evaluation_2020}.
Others compare directly with ground-truth measures of influence obtained in training-based experiments \cite{bae_if_2022,grosse_studying_2023,park_trak_2023,mlodozeniec_influence_2024,hong_better_2025,ilyas_magic_2025,sun_enhancing_2026}. 
All of these approaches only quantify whether two methods perform similarly on average. They do not consider (1) whether methods behave similarly on the same instances (which we cover through cross-method predictability and functional alignment), and more generally (2) whether attribution strategies operate on similar representations (which we capture via representational alignment).

\paragraph{Functional alignment across methods.}
Prior work that \textit{directly} compares attribution methods has characterized alignment only in terms of correlation or set overlap among the top influential training examples, which corresponds to our definition of functional alignment. However, existing comparisons vary substantially in scope:
Some compare different implementations of the same attribution paradigm \cite{nguyen_bayesian_2023, brophy_adapting_2024}. 
Others evaluate the agreement of gradient-based methods with estimates derived from non-traditional attribution strategies, such as in-context probing \cite{jiao_feasibility_2025} and neuron attribution \cite{yu_revealing_2024}.
Our work is most aligned in scope with \citet{pezeshkpour_empirical_2021}, who compare embedding similarity with four gradient-based methods using BERT models. They find that certain gradient-based methods correlate highly with more expensive Hessian-based ones, but observe only modest correlation between similarity- and gradient-based methods.
We extend this line of work by introducing two additional perspectives on alignment.
We further consider new attribution methods from additional paradigms, new tasks, and more modern model architectures.

{
\setlength{\tabcolsep}{3pt}

\begin{table}[t]
\centering
\scriptsize
\begin{tabular}{|p{0.48\columnwidth}|p{0.48\columnwidth}|}
\hline

\textbf{Gradient-based representations}
&
\textbf{Model-internal representations}
\\
\hline

\begin{tabular}[t]{@{}p{0.2\linewidth} p{0.76\linewidth}@{}}
$r^{\text{id}_{X}}_{\nabla f}$ & Raw model loss gradients \\
$r^{L2}_{\nabla f}$ & L2-normalized gradients \\
$r^{aa}_{\nabla f}$ & Adam-adjusted gradients \cite{xia_less_2024} \\
$r_{s_{\text{test}}}$ & Hessian-adjusted gradients
\end{tabular}
&
\begin{tabular}[t]{@{}p{0.2\linewidth} p{0.76\linewidth}@{}}
$r_{\text{repsim}}$ & Final-layer last token   hidden
state \cite{li_influence_2025} \\
$r_{\text{mp}}$ & Mean-pooled final-layer hidden
states \\
$r_{\text{attn}}$ & Summary of norm. att. maps
\end{tabular}
\\

\hline

\textbf{Embedding similarity}
&
\textbf{Lexical overlap}
\\
\hline

\begin{tabular}[t]{@{}p{0.2\linewidth} p{0.76\linewidth}@{}}
$qwen$ &
\href{https://huggingface.co/Qwen/Qwen3-Embedding-8B}
{Qwen3-Embedding-8B} \\
$nemot$ &
\href{https://huggingface.co/nvidia/llama-embed-nemotron-8b}
{llama-embed-nemotron-8b}
\end{tabular}
&
\begin{tabular}[t]{@{}p{0.2\linewidth} p{0.76\linewidth}@{}}
$r_{\text{bm25}}$ & BM25-based vector \\
$r_{\text{jaccard}}$ & Jaccard-based vector \\
$r_{\text{tfidf}}$ & TF-IDF-based vector
\end{tabular}
\\

\hline
\end{tabular}

\caption{Summary of attribution methods studied.}
\label{tab:representations}
\end{table}
}

\section{Methodology and Experimental Setup}\label{sec:methodology}
We first describe 3 perspectives on alignment and define corresponding metrics. We then integrate different attribution methods under a common evaluation framework, before describing experiments targeting the following main questions: (1) \textit{How aligned are methods of different paradigms?} (2) \textit{Is alignment consistent across models and tasks?} (3) \textit{What drives alignment at the instance level?}
\subsection{Alignment Measures}
\subsubsection{Measuring Representational Alignment}\label{meth:rep}
We aim to capture the similarity between similarity structures induced by different representations used for training data attribution. 
For this, we adapt methodology from mechanistic interpretability, originally proposed for studying the alignment of model-internal representations: Following \citet{huh_position_2024}, we define representations as functions $r: \mathcal{Z} \to \mathbb{R}^d$ that encode individual training documents as vectors.
For illustration, consider two representations $r_i$ and $r_j$, implemented via two text embedding models (we introduce additional representations in Section \ref{sec:representations}).
Given a similarity kernel defined by an inner product over representations, we apply two alignment metrics. 
Centered Kernel Alignment \cite{kornblith_similarity_2019} yields a measure of \textit{global representational alignment}:
\begin{equation}
\resizebox{0.6\columnwidth}{!}{$
\text{CKA}(r_i,r_j) = \frac{\|r_i^\top r_j\|_F^2}{\|r_i^\top r_i\|_F \, \|r_j^\top r_j\|_F} \,,$}
\end{equation}
where $F$ is the Frobenius norm. 
We use Centered Kernel Nearest-Neighbor Alignment (CKNN) for \textit{local representational alignment},
which restricts comparison to $k$ neighbors \cite{huh_position_2024}.
\subsubsection{Measuring Functional Alignment}
We define functional alignment as the average overall agreement between two attribution methods across test instances.
Specifically, given two vectors of pointwise relevance scores over the training data $\phi_{r_i}(z_{\text{test}}), \phi_{r_j}(z_{\text{test}}) \in R^{|\mathcal{D}|}$, we quantify agreement using Spearman's $\rho$, Kendall's $\tau$, and rank-biased overlap (RBO, prioritizes agreement at the top of ranked lists). All metrics are computed per test instance and then averaged.
\subsubsection{Measuring Cross-Method Predictability}
Methods may diverge in their global rankings but still agree on the relative importance of individual examples.
We therefore further test if attributions from one method can be leveraged to emulate those of another. Specifically, we propose a test based on pairwise comparisons: we evaluate if the scores from method $i$ can act as features to successfully predict the binary pairwise rankings of method $j$.
Let $s^{(i)}_{z_{test}}(a)$ denote the score by method $i$ for document $a$ given $z_{test}$.
For each  $z_{test}$, we construct 2000 random pairs $(a \neq b)$, subsequently split into train and test sets (80/20).
We construct features based on score differences of method $i$, and labels based on the binary ranking of method $j$:

\begin{equation}
\label{eq:pairwise}
\begin{aligned}
    x_{ab,z_{test}} &= s^{(i)}_{z_{test}}(a) - s^{(i)}_{z_{test}}(b), \\
    y^{(j)}_{ab,z_{test}} &= \mathbb{I}\big[s^{(j)}_{z_{test}}(a) > s^{(j)}_{z_{test}}(b)\big].
\end{aligned}
\end{equation}

To quantify how well rankings produced by method $j$ can be predicted using features derived from method $i$ (\textit{cross-method predictability}), we report the mean held-out accuracy of instance-specific logistic regression models without intercept, where $\tilde{x}$ denotes the score differences $x$, standardized to zero mean and unit variance across pairs of a given test instance:

\begin{equation}
\resizebox{0.8\linewidth}{!}{$
    P\!\left(y^{(j)}_{ab,z_{\text{test}}} = 1\right)
    =
    \frac{1}{
        1 + \exp\!\left(
            -w_{z_{\text{test}}}\tilde{x}_{ab,z_{\text{test}}}
        \right)
    }
$}
\end{equation}

\subsection{Representations}\label{sec:representations}
Prior works categorize by attribution strategy \cite[e.g., \textit{retraining-} vs. \textit{gradient-based}:][]{hammoudeh_training_2024}, but only the former are tractable here. Instead, we categorize by  source of information (Table \ref{tab:representations}: \textit{gradients}, \textit{model-internal representations}, \textit{embeddings}, \textit{lexical overlap}), as each corresponds to different assumptions about 
data influence.
To enable the study of representational alignment, we require 
representation
functions $r$, so that documents can be compared via inner product (Section \ref{meth:rep}). For the study of functional alignment, we obtain point-wise scores $\phi(z_{\text{test}}, z)$ by dot product between $r(z_{\text{test}})$ and the representation of the example $r(z)$ for all methods except $r_{s_\text{test}}$ (for which scores are given by Equation \ref{eq:hif}). 

Note that there exist multiple valid definitions of relevance for gradient-based methods: one can retrieve the \textit{most} or \textit{least influential} samples (i.e., those with the largest or smallest absolute similarity scores) or the \textit{most helpful} or \textit{harmful} ones (those with the most positive or negative scores; the ones that either "help" with or "harm" loss most, \citealp[e.g.,][]{koh_understanding_2017}). 
We do so to ease interpretability, given that this strategy's anticipated remove-and-retrain effect on loss most closely matches that of the semantic- and lexical-similarity-based retrieval logics (we assume that training on similar examples is more likely to decrease loss than training on unrelated ones).

\subsubsection{Gradient-based Representations} 
Common to several gradient-based strategies is the definition of influence as the similarity between test-instance gradients and training-instance gradients \cite{hammoudeh_training_2024}.
We define three representations based on prior work:

\begin{equation*}
\resizebox{\columnwidth}{!}{$
\displaystyle
r_{\nabla f}^{id_X}(z) =
\nabla \mathcal{L}(z,\theta)
\refstepcounter{equation}\quad(\theequation)
\qquad
r_{\nabla f}^{L2}(z) =
\frac{\nabla \mathcal{L}(z,\theta)}
{\|\nabla \mathcal{L}(z,\theta)\|_2}
\refstepcounter{equation}\quad(\theequation)
$}
\end{equation*}
\begin{equation}
r_{\nabla f}^{aa}(z) =
aa(\nabla \mathcal{L}(z,\theta)).
\end{equation}

\noindent Representation $r^{id_{X}}_{\nabla f}$ is based on raw gradients, as used for estimation by \citet{pruthi_estimating_2020}. $r^{L2}_{\nabla f}$ represents unit-norm gradients, as utilized in \citet{yeh_first_2022}. In $r_{\nabla f}^{aa}$, we use \textit{Adam-adjusted} (normalized) gradients, as in \citet{xia_less_2024}.

Next, we consider point-wise influence via \textit{influence functions} as in \citet{koh_understanding_2017}: 
\begin{equation}
\resizebox{0.8\columnwidth}{!}{$
IF(z, z_{\text{test}}) =
-\underbrace{H_{\hat{\theta}}^{-1} \nabla \mathcal{L}(z_{\text{test}}, \hat{\theta})}_{s_{\text{test}}}
\cdot \nabla \mathcal{L}(z, \hat{\theta}).$}\label{eq:hif}
\end{equation}
Conceptually, $s_{\text{test}}$ incorporates second-order information on how changes in parameters affect test loss.
To study the effect of this re-weighting, we define the following representation. For implementation details, see Appendix \ref{appendix:rstest}:
\begin{equation}
\resizebox{0.45\columnwidth}{!}{$
r_{s_{\text{test}}}(z)= H_{\hat{\theta}}^{-1} \nabla_{\theta} L(z, \hat{\theta}).$}
\label{eq:rstest}
\end{equation}

\subsubsection{Model-internal Representations}
We define three representations based on internal model states: 
We define $r_{\text{repsim}}$ based on the influence estimation strategy in \citet{li_influence_2025}:
\begin{equation}
\resizebox{0.35\columnwidth}{!}{$
r_{\text{repsim}}(z) = H^{(L)}_{T,:},
$
}
\end{equation}
where $H^{(L)} \in \mathbb{R}^{n \times d}$ denotes the hidden states of the final transformer layer for an input sequence $z$, and $T$ is the index of the last non-padding token.  

We obtain $r_{\text{mp}}$ by mean pooling over the non-padding tokens in the final layer. Let $m \in \{0,1\}^n$ specify the attention mask, then: 
\begin{equation}
\resizebox{0.5\columnwidth}{!}{$
r_{\text{mp}}(z) = (m^\top H^{(L)})/(\mathbf{1}^\top m).$}
\end{equation}

\noindent Next, ${r}_{\text{attn}}$ is based on attention patterns. Here, documents are considered similar if they induce similar \textit{distributions} of attention values.
\citet{kobayashi_attention_2020} define a token-to-token interaction map $M \in \mathbb{R}^{n\times n}$, $M_{i,j}:=|| \sum_h  \alpha_{i,j}^h f^h(x_j) ||$, which we adapt to support grouped-query attention (GQA) utilized by Qwen models here (Appendix \ref{appendix:rattn}).
To obtain sequence-length invariant vectors suited for comparison across documents, we pad to the maximum sequence length $S$, and then compute the singular value decompositions
$M_{l} = U \, \Sigma \, V^\top, $$ \Sigma_{l} = \operatorname{diag}(\sigma_1^{(l)}, \dots, \sigma_{S}^{(l)}).$
The attention-based representation is the concatenation of the top $k=128$ singular values across all layers:
\begin{equation}
\resizebox{0.8\linewidth}{!}{$
\displaystyle
\mathbf{r}_{\text{attn}}(z) =
\left[
\sigma^{(1)}_{1}, \dots, \sigma^{(1)}_{k},
\dots,
\sigma^{(L)}_{1}, \dots, \sigma^{(L)}_{k}
\right].
$}
\label{eq:attn_representation}
\end{equation}

\subsubsection{Embedding-based Representations} 
We define two representations on embedding similarity using external models, $r_{nemot}$: \hfcite{https://huggingface.co/nvidia/llama-embed-nemotron-8b}{babakhin_llama-embed-nemotron-8b_2025} and $r_{qwen}$: \hfcite{https://huggingface.co/Qwen/Qwen3-Embedding-8B}{zhang_qwen3_2025}, selected based on performance on the MMTEB benchmark \cite{enevoldsen_mmteb_2024}.

\subsubsection{Lexical Overlap-based Representations}
Recent work \cite[e.g.,][]{liu_olmotrace_2025} retrieves lexically related training examples to address LLM-scale datasets. 
We therefore define a set of  representations in which two documents will be assigned high similarity scores if they are lexically related to the same documents in the training data.
Our approach is inspired by the representation strategy in \citet{lagus_low-rank_2019}, who show that by representing documents as matrices $A \in \mathbb{R}^{n \times d}$, where each row corresponds to a word vector, their similarity can be measured in terms of the squared Frobenius norm of their pairwise word similarity matrices ($\|A B^\top\|_F^2$).
This approach compares documents based on the covariance of word vectors.
To represent each document in terms of its similarity to other documents in the training data while allowing comparisons via inner-product similarity,
we first generate rankings for a given test instance $z_{\text{test}}$ using Jaccard similarity (for lexical overlap), TF-IDF (weighted lexical similarity), and BM25 (probabilistic relevance).
Let $d_i$ be the $i$-th training example, and $\phi(z,z_i)$ denote the relevance of $z_i$ for $z$. Then $r_{bm25}$, $r_{tfidf}$ and $r_{jaccard}$ are of the form
 \begin{equation}
\resizebox{0.8\linewidth}{!}{$
r_{\phi}(z) = \big[\phi(z, z_i), \phi(z, z_i), \dots, \phi(z, z_N)\big].
$}
 \end{equation} 
\subsection{Experiments}
\subsubsection{Modeling Alignment Across Paradigms}
We cluster representations using agglomerative hierarchical clustering  based on pairwise alignment scores to characterize alignment structure across attribution methods.
Let \( y_{ij} \) denote the alignment score (e.g., CKA) between representations \( i \) and \( j \). We consider a pseudo-dissimilarity measure \( d_{ij} = 1 - y_{ij} \) and perform agglomerative hierarchical clustering of representations using average linkage (UPGMA: \citealp{sokal1958statistical}).
Starting from individual representations, this approach builds a hierarchy by iteratively merging the two clusters with the smallest average pairwise dissimilarity.
This yields one hierarchical clustering per alignment score, which we visualize as dendrograms. 
\subsubsection{Stability of Clusterings}\label{sec:stab_exp}
To assess whether our method groupings are stable across different base models and tasks, and not a product of random variation, we compare dendrograms using cophenetic correlation \cite{sokal_comparison_1962}. For a given clustering, let $c_{ij}$ denote the cophenetic distance between representations $r_i$ and $r_j$, defined as the dissimilarity value at which $r_i$ and $r_j$ are first merged into the same cluster.
Given two clusterings over a set of representations, the cophenetic correlation between dendrograms following \citet{sokal_comparison_1962} is
\resizebox{0.7\linewidth}{!}{$
\rho_{\text{coph}}(A,B) =
\rho\big( \{c^{A}_{ij}\}_{i<j}, \{c^{B}_{ij}\}_{i<j} \big).\notag
$}
We measure the average stability across factors
\begin{equation}
\resizebox{0.85\columnwidth}{!}{$
\displaystyle
\bar{\rho}_{d,m}
=
\frac{1}{
    \binom{|\mathcal{D}|}{2}
    \binom{|\mathcal{M}|}{2}
}
\sum_{d_1 < d_2}
\sum_{m_1 < m_2}
\rho_{\mathrm{coph}}\!\big(
    C^{d_1,m_1},
    C^{d_2,m_2}
\big),
$}
\label{eq:mean_cophenetic}
\end{equation}
\noindent as well as two conditional averages:
(i) fixing the dataset, averaging over base models,\footnote{
\resizebox{!}{0.9\baselineskip}{$\displaystyle \overline{\rho}_{d}
=
\frac{1}{\binom{|\mathcal{M}|}{2}}
\sum_{m_1 < m_2}
\rho_{\mathrm{coph}}\!\big(
C^{d,m_1},
C^{d,m_2}
\big)
$}}
and (ii) fixing the base model, averaging over datasets.
Lastly, we measure the pairwise average cophenetic stability between individual alignment measures.\footnote{
\resizebox{!}{0.8\baselineskip}{$
\overline{\rho}_{a_1,a_2}=\frac{1}{\binom{|\mathcal{D}||\mathcal{M}|}{2}}\displaystyle
\sum_{\substack{d_1,m_1 < d_2,m_2}}\rho_{\mathrm{coph}}\!\big(C^{d_1,m_1}_{a_1},C^{d_2,m_2}_{a_2}\big)
$}}

\subsubsection{Modeling Factors Affecting Alignment}\label{sec:factors}
To assess the robustness of our analysis, specifically whether alignment scores are sensitive to the choice of base model and task, we fit linear mixed-effects models and assess contributions of both factors (and their interaction) via likelihood ratio tests over nested specifications.
The full model is given by:

\resizebox{0.9\linewidth}{!}{$
\mathcal{M}_{\text{full}}:~ 
\begin{aligned}
y&\sim C(b) * C(d) \equiv\\
y_{ijbd} &= \beta_0 + \beta_b + \beta_d + \beta_{bd} + u_{ij} + \epsilon_{ijbd}, \notag
\end{aligned}
$}
\noindent where \(y_{ijbd}\) denotes the alignment score for representation pair \(ij\), given base model \(b\) and dataset \(d\). We assume
\(u_{ij} \sim \mathcal{N}(0, \sigma^2_{\text{pair}})\),
\(\epsilon_{ijbd} \sim \mathcal{N}(0, \sigma^2_\epsilon)\).
We test for interaction effects via a likelihood ratio test between $\mathcal{M}_{\text{full}}$ and
$
\mathcal{M}_{\text{add}} : y \sim C(b) + C(d)
$
corresponding to
$H_0:\ \beta_{bd} = 0.$
We likewise test the effects of base model and dataset via likelihood ratio tests between full and reduced models ($\mathcal{M}_{-B}: y \sim C(d)$, $\mathcal{M}_{-D}: y \sim C(b)$),
corresponding to $H_0:\ \beta_b = 0$ and $H_0:\ \beta_d = 0$.
We test whether each fixed effect improves fit via likelihood ratio tests and quantify effect sizes as the reductions in marginal and conditional $R^2$.\footnote{
$
R^2_m =
\frac{\sigma^2_f}{\sigma^2_f + \sigma^2_{\text{pair}} + \sigma^2_\epsilon},
$~$
R^2_c =
\frac{\sigma^2_f + \sigma^2_{\text{pair}}}{\sigma^2_f + \sigma^2_{\text{pair}} + \sigma^2_\epsilon}
$; $\sigma^2_{f}
$ \& $\sigma^2_{\text{pair}}: 
$ var expl. by fixed \& rand-eff. 
\cite{nakagawa_general_2013}.}

\subsubsection{Qualitative Comparison of Local Neighborhoods Across Representations}
The proposed measures quantify alignment but do not identify sources of agreement or disagreement at the instance level. We therefore complement them with instance-level qualitative analysis to better understand what drives misalignment.
Given two representations $r_i,r_j$ and a document $z$, we retrieve the top-5 most similar examples to $z$ under each representation. For each dataset and representation pair, we compare the resulting local neighborhoods in two settings: (i) when the neighborhoods diverge most, and (ii) when they align best (see Appendix~\ref{appendix:experiment-4} for details).
We restrict this analysis to 4 of the 12 representations, selecting a representative subset using a greedy farthest-first procedure on the CKA dissimilarity matrix.
We inspect 60 pairs across the Olmo3 models.

\subsection{Models and Data}\label{sec:models-and-data}
We fine tune Qwen3-Base \cite{yang_qwen3_2025} at sizes 0.6B,~1.7B,~4B,~8B; Olmo-3-1025-7B \cite{olmo_olmo_2025}; OLMo-2-0425-1B \cite{walsh_2_2025}; and Llama-3.2-1B \cite{dubey_llama_2024} for one epoch using LoRA \cite{hu_lora_2021} on the following datasets (for details see Appendix \ref{appendix:finetuning}):

\begin{description}[leftmargin=0pt, labelsep=0.1em, nosep]
    \item [\textbf{SFT}]: the instruction-tuning dataset for OLMo2 based on \textit{Tülu3} (\href{https://huggingface.co/datasets/allenai/tulu-3-sft-olmo-2-mixture-0225}{allenai/tulu-3-sft-olmo-2-mixture-0225}: \citealp{lambert_tulu_2025-1});
    \item [\textbf{math}] (\href{https://huggingface.co/datasets/allenai/tulu-3-sft-personas-math}{allenai/tulu-3-sft-personas-math}), and \textbf{code}: subsets of \textit{Tülu3} targeting math and coding problems (\href{https://huggingface.co/datasets/allenai/tulu-3-sft-personas-code}{allenai/tulu-3-sft-personas-code});
    \item [\textbf{squad\_v2}]: reading comprehension for extractive question answering \cite{rajpurkar_know_2018};
    \item[\textbf{ftrace}]: benchmark dataset that perturbs documents to increase attribution difficulty \cite{jiao_date-lm_2025}.
\end{description}

We fine-tune 35 models on the full datasets and measure alignment across 1000 random documents per dataset and model combination.

\section{Results}
We report pairwise representational alignment (CKA), functional alignment (Spearman's $\rho$), and cross-method predictability averaged across models and tasks in Figure \ref{fig:alignment-main}, for Kendall's $\tau$, RBO, per model- and per task results see Appendix \ref{appendix:alignment}.  

\begin{figure}[htb]
    \centering
    
    \setlength{\tabcolsep}{0pt} 

    \begin{minipage}{\columnwidth}

    \newlength{\imgwidth}
    \setlength{\imgwidth}{\dimexpr\columnwidth-2em\relax}

    \begin{tabular}{@{}c@{}c@{}}
        \parbox[c]{2em}{
            \centering
            \scriptsize
                    \rotatebox[origin=c]{90}{%
            \shortstack{
                representational\\
                alignment (CKA)
            }
        }
    }
    &
        \parbox[c]{\imgwidth}{
            \centering
            \includegraphics[width=\imgwidth,
                trim=0.1cm 0.25cm 0.2cm 0.05cm, clip]{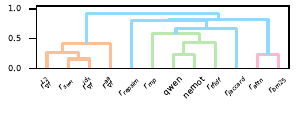}
        }
    \end{tabular}

    \vspace{-0.15em}

    \begin{tabular}{@{}c@{}c@{}}
        \parbox[c]{2em}{
            \centering
            \scriptsize
                    \rotatebox[origin=c]{90}{%
            \shortstack{
                functional\\
                alignment ($\rho$)
            }
        }
    }
    &
        \parbox[c]{\imgwidth}{
            \centering
            \includegraphics[width=\imgwidth,
                trim=0.1cm 0.25cm 0.2cm 0.05cm, clip]{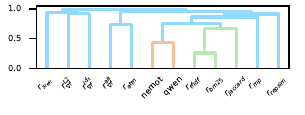}
        }
    \end{tabular}

    \vspace{-0.15em}

    \begin{tabular}{@{}c@{}c@{}}
        \parbox[c]{2em}{
            \centering
            \scriptsize
                    \rotatebox[origin=c]{90}{%
            \shortstack{
                cross-method\\
                predictability
            }
        }
    }
    &
        \parbox[c]{\imgwidth}{
            \centering
            \includegraphics[width=\imgwidth,
                trim=0.1cm 0.25cm 0.2cm 0.05cm, clip]{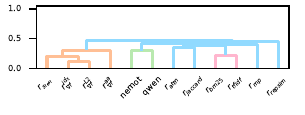}
        }
    \end{tabular}

    \end{minipage}

    \caption{\textbf{Hierarchical clustering of attribution methods.}
    Clusters largely align with attribution paradigms.}
    \label{fig:dendograms-main}
\end{figure}

\begin{figure*}[htb]
    \centering
    \begin{minipage}{\linewidth}
    \includegraphics[width=0.31\linewidth]{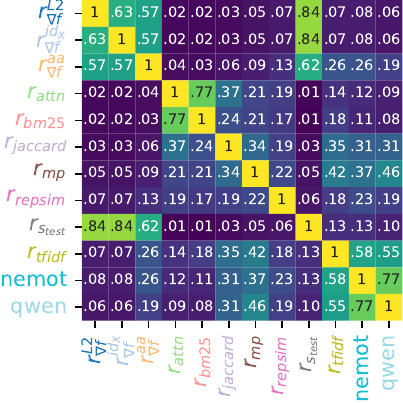}
    \hfill
    \includegraphics[width=0.31\linewidth]{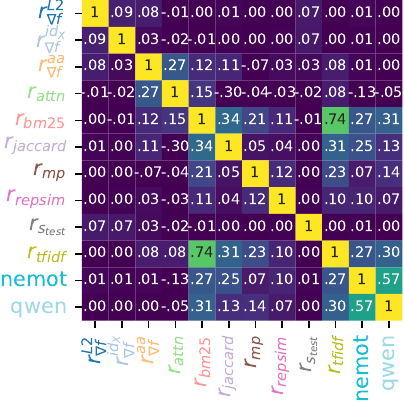}
    \hfill
    \includegraphics[width=0.31\linewidth]{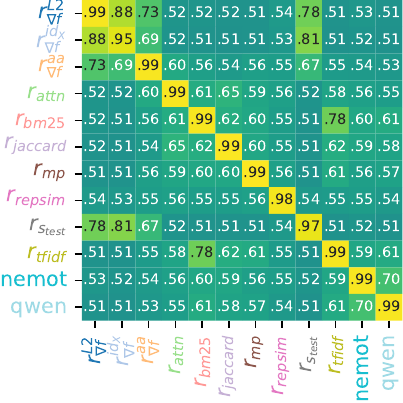}
    \end{minipage}
    \caption{\textbf{Pairwise alignment.} Left: representational alignment (CKA $\in [0,1]$), center: functional alignment (Spearman's $\rho$ $\in [-1,1]$), and right: cross-method predictability (acc $\in [0,1]$) averaged across tasks and models. Higher values suggest better alignment. For CKNNA@25, RBO, and Kendall's $\tau$, see Appendix \ref{appendix:alignment}.}
    \label{fig:alignment-main}
\end{figure*}

\subsection{Alignment Across Paradigms}
Regarding our first question (\emph{how aligned are methods from different paradigms?}), we find that attribution methods cluster by their underlying paradigms, exhibiting high internal agreement (e.g., $r_{\nabla f}^{L2}$ vs. $r_{\nabla f}^{id_X}$) but low cross-paradigm alignment (e.g., $r_{\nabla f}^{L2}$ vs. $r_{mp}$). This is reflected in the dendrograms in Figure \ref{fig:dendograms-main}, which show clear separation between gradient-based, embedding-based, and lexical similarity methods.
Overall, we observe clear within-paradigm alignment, as evidenced by a pronounced block-diagonal structure in the heatmaps of Figure \ref{fig:alignment-main} and Appendix \ref{appendix:alignment}. 
This is most pronounced for representational alignment (CKA, CKNNA@25) and cross-method predictability within gradient-based ($r_{s_{\text{test}}}$, $r_{\nabla f}^{aa}$, $r_{\nabla f}^{L2}$, $r_{\nabla f}^{id_X}$) and embedding-based attribution paradigms ($r_{nemot}$, $r_{qwen}$).
For functional alignment ($\rho$, $\tau$, RBO), however, this pattern only holds for embedding-based methods.
One notable exception is the pairing of BM25 and the attention-based method, which features relatively high representational alignment (yet still weak functional alignment and low cross-method predictability).

While we measure considerably lower rank correlations than \citet{pezeshkpour_empirical_2021} report for BERT models, our results broadly align with theirs for the strategies considered in both studies. Specifically, gradient- and Hessian-based methods exhibit greater functional alignment than cross-paradigm pairings, while correlation between gradient- and embedding-based methods is negligible.

Per-task predictability results (Appendix \ref{appendix:alignment}) show substantial variation within the same paradigm, for example, alignment between $r_{\nabla f}^{aa}$ and $r_{s_{\text{test}}}$ is about 11-20 pt lower on \textit{SFT} (0.56) than on other tasks. Alignment across paradigms, however, appears largely invariant to task.
\paragraph{Stability of clusterings.}
The average cophenetic stability $\bar{\rho}_{d,m}$ is high across metrics
(CKA: 0.80\unskip, CKNNA@25: 0.75\unskip,
$\rho$: 0.71\unskip,
$\tau$: 0.75\unskip,
$RBO$: 0.72\unskip,
 $acc_{\text{pred}}$: 0.79\unskip), indicating that dendrogram structure is largely preserved across conditions. 
 Stability across base models $\bar{\rho}_{b}$ and datasets $\bar{\rho}_{d}$ is consistently high as well ($> 0.7$, Appendix \ref{appendix:cophstab}).
Regarding cophenetic stability between alignment metrics (CKA, $\rho$, etc.), $\bar{\rho}_{a_1,a_2}$, we find high agreement between measures targeting the same quantity (representational alignment, functional alignment, or cross-method predictability). Measures targeting representational alignment (CKA, CKNNA@25) show strong consistency ($\bar{\rho}_{a_1,a_2} \approx 0.70$), while functional alignment measures ($\tau$, RBO, $\rho$) also exhibit moderate-to-high agreement ($\approx0.57-0.61$). In contrast, correlations across metrics targeting different quantities (e.g., CKA vs. $\rho$) are low ($\approx0.02 - 0.30$). Weak correlation to functional alignment confirms that cross-method predictability captures a distinct notion of alignment.

\subsection{Factors Affecting Alignment.} To answer our second question (\emph{is alignment consistent across models and tasks?}), we find that alignment patterns (rather than the clusterings discussed earlier) are broadly stable with respect to base model and dataset choice: while different choices induce statistically significant effects in likelihood-ratio tests, the practical effect sizes ($\Delta R^2$, Table \ref{tab:delta-r2-main} in Appendix \ref{appendix:lrt}) are minor. This demonstrates that the specific alignment properties (or lack thereof) between paradigms are fundamental, rather than artifacts of a particular experimental setup. Among these secondary factors, removing the dataset factor yields the largest decrease in explained variance across metrics, except CKA. Conversely, the base model choice shows a stronger effect only on global representational alignment (CKA), but not on its local counterpart (CKNNA@25). Interaction terms between the base model and dataset contribute comparatively little additional explanatory power.

\subsection{Qualitative Insights on Alignment}
To address our final question (\emph{what drives alignment at the instance level?}), we manually compare local neighborhoods of different representations to understand the qualitative sources of agreement and disagreement. 
We examine whether local neighborhoods differ across attribution representations in terms of 
(1) prompt and completion format, such as prompting style or step-by-step thinking,
(2) keyword overlap, and
(3) shared topics or themes.
For each pairing of attribution representations (e.g., $r_{\nabla f}^{id_X}$ vs. $r_{bm25}$), we analyze one test instance with minimal- and one test instance with maximal overlap between neighborhoods.
Below we summarize our results; for full annotations see Appendix \ref{appendix:experiment-4}.

\paragraph{Alignment depends on instruction tuning goals.}
Alignment patterns vary across tasks. For the datasets \textit{math} and \textit{code}, neighborhoods tend to align in terms of example structure, such as shared problem-decomposition patterns or prompting styles. In contrast, neighborhoods for \textit{squad\_v2} and \textit{ftrace} frequently align at the level of high-level question type, or semantic category (e.g., incomplete-information handling).
When BM25-based neighborhoods lack keyword overlap with the test instance, they also tend to have low overlap with other strategies. In contrast, neighborhoods of gradient- and embedding-based methods often consist of instances with similar structures or completion behaviors, despite limited keyword overlap.
\paragraph{Prompting artifacts drive misalignment.}
Recurring instruction specifications and persona-prompting templates introduce repeated structure across all datasets, increasing lexical overlap.
BM25 appears to be frequently driven by these artifacts. 
In contrast, representation-based methods appear to exploit more meaningful task-related similarities between instances (from an instruction-tuning perspective). 
On the other hand, test instances targeting abilities not directly related to task solving (such as safety refusals, indications of missing information, or reproduction of hard-coded facts such as knowledge-cutoff dates) often appear in the same neighborhoods under the representation-based methods ($r^{id_X}_{\nabla f}$, $r_{\mathrm{repsim}}$, and $r_{\mathrm{nemot}}$). This occurs even when the instances are neither semantically nor topically related: for example, some neighborhoods contain \textit{both} safety-related examples and examples on the cutoff date.

\section{Discussion}
\paragraph{Task dependency of alignment and limitations of lexical matching.}
The divergence across paradigms appears to stem from differences in how each method defines similarity. Lexical methods (e.g., BM25) primarily capture surface form and instruction overlap, and therefore diverge in their attributions from other strategies when structure is shared, but wording differs. In contrast, gradient- and embedding-based methods likely better capture structural and behavioral similarity, grouping instances by shared instruction and completion patterns rather than keyword overlap. Given that learning behavior rather than memorization is the central goal of instruction tuning tasks, this suggests that relying on lexical overlap alone may be a weak basis for attribution in instruction tuning.
\paragraph{Measuring functional alignment is insufficient.}
Our results show that the three proposed evaluation protocols capture distinct and only partially overlapping aspects of alignment. We further show that methods may exhibit low correlation in attributions across the full dataset yet strong pairwise agreement on example relevance, as well as high representational alignment. 
This implies that low average rank correlation does not necessarily preclude one method from serving as a useful substitute for another in a given downstream task.
Accordingly, relying solely on rank-correlation analyses may lead to the unwarranted disqualification of prospective methods, whereas our framework can provide more nuanced insights into alignment.

Additionally, our results may also partially explain the apparent mismatch between prior findings, where some report low cross-paradigm correlation, while others independently observe similar downstream-task performance of methods belonging to different paradigms. To assess whether this hypothesis holds, future work should design evaluation protocols spanning downstream tasks that vary in their reliance on local versus global agreement, and test whether cross-method predictability scores predict performance on individual tasks.

\paragraph{Mechanistic interpretation of alignment.}
Despite negligible functional alignment, we measure high representational alignment 
and cross-method predictability between gradient-similarity methods and the 
Hessian-based $r_{s_\text{test}}$. Decomposing alignment into the three suggested 
dimensions can provide a mechanistic interpretation of this apparent discrepancy:
High representational alignment suggests that 
the reweighting in $r_{s_\text{test}}$ (Equation {\ref{eq:rstest}}) does not 
fundamentally alter the representation structure. While it systematically changes score assignment, it retains information about pairwise relevance, as indicated by the high cross-method predictability.
Similarly, although representation-based methods may be expected to align closely with one another, our representational alignment results and qualitative analysis show that alignment is contingent on capturing the same task-specific features.

\paragraph{Alignment deteriorates cross-paradigm.}
While most pairings achieve above-chance cross-method predictability, accuracy is only consistently above 0.75 for within-paradigm comparisons, whereas the only cross-paradigm pairings scoring above 0.65 are those between lexical- and embedding-based methods.
Representational and functional alignment across paradigms is likewise weak.
Our findings suggest that practitioners seeking to use one method as a proxy for another (e.g., to pre-filter examples before applying more computationally expensive but also more accurate methods) should select methods of the same attribution paradigm.
At the same time, the observed lack of alignment suggests an opportunity for future work on \textit{example-based explanations} built on data attribution: combining orthogonal strategies may yield more comprehensive characterizations of data influence.

\section{Conclusion}
We introduce a framework for directly measuring alignment between training data attribution methods, which captures correlation of attribution rankings (\textit{functional alignment}), assesses structural similarity of the underlying representation spaces used for attribution (\textit{representational alignment}), and evaluates agreement on pairwise ranking decisions (\textit{cross-method predictability}). We show that this decomposition is useful for disentangling different notions of alignment that are not captured by existing evaluations based on measures of downstream performance.
Comparing 12 methods spanning four paradigms, we find high alignment within individual attribution paradigms (gradient similarity, model-internal representations, embedding similarity, and lexical overlap). In contrast, cross-paradigm alignment is consistently lower and task-dependent. This suggests that methods within a paradigm are partially interchangeable for downstream applications, whereas those across paradigms are not reliable proxies for one another.
Qualitatively, we find that this divergence stems from a fundamental difference in how paradigms define similarity: lexical methods capture surface-level prompting artifacts, whereas representation-based ones appear to capture behavioral patterns.

Overall, our results show that representational alignment, functional alignment, and cross-method predictability do not necessarily covary, emphasizing the need to distinguish between local and overall agreement in evaluation.
To this end, we recommend augmenting work on new attribution methods through systematic alignment analysis to contextualize new approaches within known paradigms.

\section*{Limitations}\label{limitations}
Our experiments consider text generation tasks. Alignment between specific paradigms may differ for other task formats, such as simpler classification setups. However, our framework is fully compatible with other task definitions and models outside NLP contexts.

In contrast to prior work, we intentionally forgo evaluating downstream performance in favor of directly testing alignment, which is required to establish a more fundamental understanding. Our focus was on developing complementary perspectives on alignment, rather than studying failure modes specific to individual tasks or replicating prior downstream evaluations.

Note that our analysis primarily provides recommendations on evaluation practices; our study is not designed to produce quality or performance characterizations of individual attribution methods that could directly inform the design of new attribution methods.

We restrict our analysis to base models with a maximum of 8B parameters due to the computational costs involved with obtaining and analyzing gradient-based influence estimates. We ensure robustness by validating our findings across seven models spanning parameter sizes from 0.6B to 8B and three model families. 
Likewise, in line with prior work \cite[e.g.,][]{kwon_datainf_2023,xia_less_2024}, we only consider LoRA-fine-tuned models in our analysis due to the high computational costs of gradient-based attribution. Alignment patterns may vary when attribution targets other (subsets) of model parameters.

While future work may explore alternative training data attribution paradigms, we have compiled a set of 12 methods across four paradigms, which represent a cross-section of the current literature.

\section*{Ethical Considerations}\label{risks}
Given that we document patterns of alignment across attribution paradigms, there is a risk that actors acting in bad faith could misuse this information to select an attribution strategy that aligns with their desired claims of data influence, despite prior negative results obtained with other methods. Our work mitigates this risk by demonstrating that single-paradigm attribution may provide an incomplete view of data influence.

\section*{Acknowledgments}
We would like to thank the anonymous reviewers for their helpful comments and suggestions.
This research has been funded by the Vienna Science and Technology Fund (WWTF)[10.47379/VRG19008] "Knowledge-infused Deep Learning for Natural Language Processing".
\bibliography{custom}

@InProceedings{jiao_date-lm_2025,
  author     = {Jiao, Cathy and Pan, Yijun and Xiao, Emily and Sheng, Daisy and Jain, Niket and Zhao, Hanzhang and Dasgupta, Ishita and Ma, Jiaqi W. and Xiong, Chenyan},
  booktitle  = {The {{Thirty-ninth Annual Conference}} on {{Neural Information Processing Systems Datasets}} and {{Benchmarks Track}}},
  title      = {{{DATE-LM}}: {{Benchmarking Data Attribution Evaluation}} for {{Large Language Models}}},
  year       = {2025},
  month      = oct,
  langid     = {english},
  shorttitle = {{{DATE-LM}}},
  url        = {https://openreview.net/forum?id=e2cD5xuHix},
  urldate    = {2025-12-02},
}

@Article{nguyen_bayesian_2023,
  author  = {Nguyen, Elisa and Seo, Minjoon and Oh, Seong Joon},
  journal = {Advances in Neural Information Processing Systems},
  title   = {A {{Bayesian Approach To Analysing Training Data Attribution In Deep Learning}}},
  year    = {2023},
  month   = dec,
  pages   = {64155--64180},
  volume  = {36},
  langid  = {english},
  url     = {https://proceedings.neurips.cc/paper_files/paper/2023/hash/ca774047bc3b46cc81e53ead34cd5d5a-Abstract-Conference.html},
  urldate = {2024-07-30},
}

@InProceedings{kobayashi_attention_2020,
  author     = {Kobayashi, Goro and Kuribayashi, Tatsuki and Yokoi, Sho and Inui, Kentaro},
  booktitle  = {Proceedings of the 2020 {{Conference}} on {{Empirical Methods}} in {{Natural Language Processing}} ({{EMNLP}})},
  title      = {Attention Is {{Not Only}} a {{Weight}}: {{Analyzing Transformers}} with {{Vector Norms}}},
  year       = {2020},
  address    = {Online},
  editor     = {Webber, Bonnie and Cohn, Trevor and He, Yulan and Liu, Yang},
  month      = nov,
  pages      = {7057--7075},
  publisher  = {Association for Computational Linguistics},
  doi        = {10.18653/v1/2020.emnlp-main.574},
  shorttitle = {Attention Is {{Not Only}} a {{Weight}}},
  urldate    = {2025-06-26},
}

@InProceedings{wu_enhancing_2024,
  author    = {Wu, Kangxi and Pang, Liang and Shen, Huawei and Cheng, Xueqi},
  booktitle = {Proceedings of the 2024 {{Conference}} on {{Empirical Methods}} in {{Natural Language Processing}}},
  title     = {Enhancing {{Training Data Attribution}} for {{Large Language Models}} with {{Fitting Error Consideration}}},
  year      = {2024},
  address   = {Miami, Florida, USA},
  editor    = {{Al-Onaizan}, Yaser and Bansal, Mohit and Chen, Yun-Nung},
  month     = nov,
  pages     = {14131--14143},
  publisher = {Association for Computational Linguistics},
  doi       = {10.18653/v1/2024.emnlp-main.782},
  urldate   = {2025-12-02},
}

@InProceedings{enevoldsen_mmteb_2024,
  author     = {Enevoldsen, Kenneth and Chung, Isaac and Kerboua, Imene and Kardos, M{\'a}rton and Mathur, Ashwin and Stap, David and Gala, Jay and Siblini, Wissam and Krzemi{\'n}ski, Dominik and Winata, Genta Indra and Sturua, Saba and Utpala, Saiteja and Ciancone, Mathieu and Schaeffer, Marion and Misra, Diganta and Dhakal, Shreeya and Rystr{\o}m, Jonathan and Solomatin, Roman and {\c C}a{\u g}atan, {\"O}mer Veysel and Kundu, Akash and Bernstorff, Martin and Xiao, Shitao and Sukhlecha, Akshita and Pahwa, Bhavish and Po{\'s}wiata, Rafa{\l} and Gv, Kranthi Kiran and Ashraf, Shawon and Auras, Daniel and Pl{\"u}ster, Bj{\"o}rn and Harries, Jan Philipp and Magne, Lo{\"i}c and Mohr, Isabelle and Zhu, Dawei and {Gisserot-Boukhlef}, Hippolyte and Aarsen, Tom and Kostkan, Jan and Wojtasik, Konrad and Lee, Taemin and Suppa, Marek and Zhang, Crystina and Rocca, Roberta and Hamdy, Mohammed and Michail, Andrianos and Yang, John and Faysse, Manuel and Vatolin, Aleksei and Thakur, Nandan and Dey, Manan and Vasani, Dipam and Chitale, Pranjal A. and Tedeschi, Simone and Tai, Nguyen and Snegirev, Artem and Hendriksen, Mariya and G{\"u}nther, Michael and Xia, Mengzhou and Shi, Weijia and L{\`u}, Xing Han and Clive, Jordan and K, Gayatri and Anna, Maksimova and Wehrli, Silvan and Tikhonova, Maria and Panchal, Henil Shalin and Abramov, Aleksandr and Ostendorff, Malte and Liu, Zheng and Clematide, Simon and Miranda, Lester James Validad and Fenogenova, Alena and Song, Guangyu and Safi, Ruqiya Bin and Li, Wen-Ding and Borghini, Alessia and Cassano, Federico and Hansen, Lasse and Hooker, Sara and Xiao, Chenghao and Adlakha, Vaibhav and Weller, Orion and Reddy, Siva and Muennighoff, Niklas},
  booktitle  = {The {{Thirteenth International Conference}} on {{Learning Representations}}},
  title      = {{{MMTEB}}: {{Massive Multilingual Text Embedding Benchmark}}},
  year       = {2024},
  month      = oct,
  langid     = {english},
  shorttitle = {{{MMTEB}}},
  url        = {https://openreview.net/forum?id=zl3pfz4VCV},
  urldate    = {2026-05-21},
}

@InProceedings{gu_iaeval_2023,
  author     = {Gu, Peijian and Shen, Yaozong and Wang, Lijie and Wang, Quan and Wu, Hua and Mao, Zhendong},
  booktitle  = {Findings of the {{Association}} for {{Computational Linguistics}}: {{EMNLP}} 2023},
  title      = {{{IAEval}}: {{A Comprehensive Evaluation}} of {{Instance Attribution}} on {{Natural Language Understanding}}},
  year       = {2023},
  address    = {Singapore},
  editor     = {Bouamor, Houda and Pino, Juan and Bali, Kalika},
  month      = dec,
  pages      = {11966--11977},
  publisher  = {Association for Computational Linguistics},
  doi        = {10.18653/v1/2023.findings-emnlp.801},
  shorttitle = {{{IAEval}}},
  urldate    = {2025-08-21},
}

@Misc{ilyas_magic_2025,
  author        = {Ilyas, Andrew and Engstrom, Logan},
  month         = apr,
  title         = {{{MAGIC}}: {{Near-Optimal Data Attribution}} for {{Deep Learning}}},
  year          = {2025},
  archiveprefix = {arXiv},
  doi           = {10.48550/arXiv.2504.16430},
  eprint        = {2504.16430},
  number        = {arXiv:2504.16430},
  primaryclass  = {cs.LG},
  publisher     = {arXiv},
  shorttitle    = {{{MAGIC}}},
  urldate       = {2026-05-19},
}

@InProceedings{yu_revealing_2024,
  author     = {Yu, Haeun and Atanasova, Pepa and Augenstein, Isabelle},
  booktitle  = {Proceedings of the 62nd {{Annual Meeting}} of the {{Association}} for {{Computational Linguistics}} ({{Volume}} 1: {{Long Papers}})},
  title      = {Revealing the {{Parametric Knowledge}} of {{Language Models}}: {{A Unified Framework}} for {{Attribution Methods}}},
  year       = {2024},
  address    = {Bangkok, Thailand},
  editor     = {Ku, Lun-Wei and Martins, Andre and Srikumar, Vivek},
  month      = aug,
  pages      = {8173--8186},
  publisher  = {Association for Computational Linguistics},
  doi        = {10.18653/v1/2024.acl-long.444},
  shorttitle = {Revealing the {{Parametric Knowledge}} of {{Language Models}}},
  urldate    = {2025-12-03},
}

@InProceedings{liu_olmotrace_2025,
  author     = {Liu, Jiacheng and Blanton, Taylor and Elazar, Yanai and Min, Sewon and Chen, Yen-Sung and {Chheda-Kothary}, Arnavi and Tran, Huy and Bischoff, Byron and Marsh, Eric and Schmitz, Michael and Trier, Cassidy and Sarnat, Aaron and James, Jenna and Borchardt, Jon and Kuehl, Bailey and Cheng, Evie Yu-Yen and Farley, Karen and Anderson, Taira and Albright, David and Schoenick, Carissa and Soldaini, Luca and Groeneveld, Dirk and Pang, Rock Yuren and Koh, Pang Wei and Smith, Noah A. and Lebrecht, Sophie and Choi, Yejin and Hajishirzi, Hannaneh and Farhadi, Ali and Dodge, Jesse},
  booktitle  = {Proceedings of the 63rd {{Annual Meeting}} of the {{Association}} for {{Computational Linguistics}} ({{Volume}} 3: {{System Demonstrations}})},
  title      = {{{OLMoTrace}}: {{Tracing Language Model Outputs Back}} to {{Trillions}} of {{Training Tokens}}},
  year       = {2025},
  address    = {Vienna, Austria},
  editor     = {Mishra, Pushkar and Muresan, Smaranda and Yu, Tao},
  month      = jul,
  pages      = {178--188},
  publisher  = {Association for Computational Linguistics},
  doi        = {10.18653/v1/2025.acl-demo.18},
  isbn       = {979-8-89176-253-4},
  shorttitle = {{{OLMoTrace}}},
  urldate    = {2026-03-11},
}

@InProceedings{xia_less_2024,
  author     = {Xia, Mengzhou and Malladi, Sadhika and Gururangan, Suchin and Arora, Sanjeev and Chen, Danqi},
  booktitle  = {Proceedings of the 41st {{International Conference}} on {{Machine Learning}}},
  title      = {{{LESS}}: {{Selecting Influential Data}} for {{Targeted Instruction Tuning}}},
  year       = {2024},
  month      = jul,
  pages      = {54104--54132},
  publisher  = {PMLR},
  issn       = {2640-3498},
  langid     = {english},
  shorttitle = {{{LESS}}},
  url        = {https://proceedings.mlr.press/v235/xia24c.html},
  urldate    = {2026-06-24},
}

@Misc{dubey_llama_2024,
  author        = {Dubey, Abhimanyu and Jauhri, Abhinav and Pandey, Abhinav and Kadian, Abhishek and {Al-Dahle}, Ahmad and Letman, Aiesha and Mathur, Akhil and Schelten, Alan and Yang, Amy and Fan, Angela and Goyal, Anirudh and Hartshorn, Anthony and Yang, Aobo and Mitra, Archi and Sravankumar, Archie and Korenev, Artem and Hinsvark, Arthur and Rao, Arun and Zhang, Aston and Rodriguez, Aurelien and Gregerson, Austen and Spataru, Ava and Roziere, Baptiste and Biron, Bethany and Tang, Binh and Chern, Bobbie and Caucheteux, Charlotte and Nayak, Chaya and Bi, Chloe and Marra, Chris and McConnell, Chris and Keller, Christian and Touret, Christophe and Wu, Chunyang and Wong, Corinne and Ferrer, Cristian Canton and Nikolaidis, Cyrus and Allonsius, Damien and Song, Daniel and Pintz, Danielle and Livshits, Danny and Esiobu, David and Choudhary, Dhruv and Mahajan, Dhruv and {Garcia-Olano}, Diego and Perino, Diego and Hupkes, Dieuwke and Lakomkin, Egor and AlBadawy, Ehab and Lobanova, Elina and Dinan, Emily and Smith, Eric Michael and Radenovic, Filip and Zhang, Frank and Synnaeve, Gabriel and Lee, Gabrielle and Anderson, Georgia Lewis and Nail, Graeme and Mialon, Gregoire and Pang, Guan and Cucurell, Guillem and Nguyen, Hailey and Korevaar, Hannah and Xu, Hu and Touvron, Hugo and Zarov, Iliyan and Ibarra, Imanol Arrieta and Kloumann, Isabel and Misra, Ishan and Evtimov, Ivan and Copet, Jade and Lee, Jaewon and Geffert, Jan and Vranes, Jana and Park, Jason and Mahadeokar, Jay and Shah, Jeet and {van der Linde}, Jelmer and Billock, Jennifer and Hong, Jenny and Lee, Jenya and Fu, Jeremy and Chi, Jianfeng and Huang, Jianyu and Liu, Jiawen and Wang, Jie and Yu, Jiecao and Bitton, Joanna and Spisak, Joe and Park, Jongsoo and Rocca, Joseph and Johnstun, Joshua and Saxe, Joshua and Jia, Junteng and Alwala, Kalyan Vasuden and Upasani, Kartikeya and Plawiak, Kate and Li, Ke and Heafield, Kenneth and Stone, Kevin and {El-Arini}, Khalid and Iyer, Krithika and Malik, Kshitiz and Chiu, Kuenley and Bhalla, Kunal and {Rantala-Yeary}, Lauren and {van der Maaten}, Laurens and Chen, Lawrence and Tan, Liang and Jenkins, Liz and Martin, Louis and Madaan, Lovish and Malo, Lubo and Blecher, Lukas and Landzaat, Lukas and {de Oliveira}, Luke and Muzzi, Madeline and Pasupuleti, Mahesh and Singh, Mannat and Paluri, Manohar and Kardas, Marcin and Oldham, Mathew and Rita, Mathieu and Pavlova, Maya and Kambadur, Melanie and Lewis, Mike and Si, Min and Singh, Mitesh Kumar and Hassan, Mona and Goyal, Naman and Torabi, Narjes and Bashlykov, Nikolay and Bogoychev, Nikolay and Chatterji, Niladri and Duchenne, Olivier and {\c C}elebi, Onur and Alrassy, Patrick and Zhang, Pengchuan and Li, Pengwei and Vasic, Petar and Weng, Peter and Bhargava, Prajjwal and Dubal, Pratik and Krishnan, Praveen and Koura, Punit Singh and Xu, Puxin and He, Qing and Dong, Qingxiao and Srinivasan, Ragavan and Ganapathy, Raj and Calderer, Ramon and Cabral, Ricardo Silveira and Stojnic, Robert and Raileanu, Roberta and Girdhar, Rohit and Patel, Rohit and Sauvestre, Romain and Polidoro, Ronnie and Sumbaly, Roshan and Taylor, Ross and Silva, Ruan and Hou, Rui and Wang, Rui and Hosseini, Saghar and Chennabasappa, Sahana and Singh, Sanjay and Bell, Sean and Kim, Seohyun Sonia and Edunov, Sergey and Nie, Shaoliang and Narang, Sharan and Raparthy, Sharath and Shen, Sheng and Wan, Shengye and Bhosale, Shruti and Zhang, Shun and Vandenhende, Simon and Batra, Soumya and Whitman, Spencer and Sootla, Sten and Collot, Stephane and Gururangan, Suchin and Borodinsky, Sydney and Herman, Tamar and Fowler, Tara and Sheasha, Tarek and Georgiou, Thomas and Scialom, Thomas and Speckbacher, Tobias and Mihaylov, Todor and Xiao, Tong and Karn, Ujjwal and Goswami, Vedanuj and Gupta, Vibhor and Ramanathan, Vignesh and Kerkez, Viktor and Gonguet, Vincent and Do, Virginie and Vogeti, Vish and Petrovic, Vladan and Chu, Weiwei and Xiong, Wenhan and Fu, Wenyin and Meers, Whitney and Martinet, Xavier and Wang, Xiaodong and Tan, Xiaoqing Ellen and Xie, Xinfeng and Jia, Xuchao and Wang, Xuewei and Goldschlag, Yaelle and Gaur, Yashesh and Babaei, Yasmine and Wen, Yi and Song, Yiwen and Zhang, Yuchen and Li, Yue and Mao, Yuning and Coudert, Zacharie Delpierre and Yan, Zheng and Chen, Zhengxing and Papakipos, Zoe and Singh, Aaditya and Grattafiori, Aaron and Jain, Abha and Kelsey, Adam and Shajnfeld, Adam and Gangidi, Adithya and Victoria, Adolfo and Goldstand, Ahuva and Menon, Ajay and Sharma, Ajay and Boesenberg, Alex and Vaughan, Alex and Baevski, Alexei and Feinstein, Allie and Kallet, Amanda and Sangani, Amit and Yunus, Anam and Lupu, Andrei and Alvarado, Andres and Caples, Andrew and Gu, Andrew and Ho, Andrew and Poulton, Andrew and Ryan, Andrew and Ramchandani, Ankit and Franco, Annie and Saraf, Aparajita and Chowdhury, Arkabandhu and Gabriel, Ashley and Bharambe, Ashwin and Eisenman, Assaf and Yazdan, Azadeh and James, Beau and Maurer, Ben and Leonhardi, Benjamin and Huang, Bernie and Loyd, Beth and De Paola, Beto and Paranjape, Bhargavi and Liu, Bing and Wu, Bo and Ni, Boyu and Hancock, Braden and Wasti, Bram and Spence, Brandon and Stojkovic, Brani and Gamido, Brian and Montalvo, Britt and Parker, Carl and Burton, Carly and Mejia, Catalina and Wang, Changhan and Kim, Changkyu and Zhou, Chao and Hu, Chester and Chu, Ching-Hsiang and Cai, Chris and Tindal, Chris and Feichtenhofer, Christoph and Civin, Damon and Beaty, Dana and Kreymer, Daniel and Li, Daniel and Wyatt, Danny and Adkins, David and Xu, David and Testuggine, Davide and David, Delia and Parikh, Devi and Liskovich, Diana and Foss, Didem and Wang, Dingkang and Le, Duc and Holland, Dustin and Dowling, Edward and Jamil, Eissa and Montgomery, Elaine and Presani, Eleonora and Hahn, Emily and Wood, Emily and Brinkman, Erik and Arcaute, Esteban and Dunbar, Evan and Smothers, Evan and Sun, Fei and Kreuk, Felix and Tian, Feng and Ozgenel, Firat and Caggioni, Francesco and Guzm{\'a}n, Francisco and Kanayet, Frank and Seide, Frank and Florez, Gabriela Medina and Schwarz, Gabriella and Badeer, Gada and Swee, Georgia and Halpern, Gil and Thattai, Govind and Herman, Grant and Sizov, Grigory and Guangyi and Zhang and Lakshminarayanan, Guna and Shojanazeri, Hamid and Zou, Han and Wang, Hannah and Zha, Hanwen and Habeeb, Haroun and Rudolph, Harrison and Suk, Helen and Aspegren, Henry and Goldman, Hunter and Damlaj, Ibrahim and Molybog, Igor and Tufanov, Igor and Veliche, Irina-Elena and Gat, Itai and Weissman, Jake and Geboski, James and Kohli, James and Asher, Japhet and Gaya, Jean-Baptiste and Marcus, Jeff and Tang, Jeff and Chan, Jennifer and Zhen, Jenny and Reizenstein, Jeremy and Teboul, Jeremy and Zhong, Jessica and Jin, Jian and Yang, Jingyi and Cummings, Joe and Carvill, Jon and Shepard, Jon and McPhie, Jonathan and Torres, Jonathan and Ginsburg, Josh and Wang, Junjie and Wu, Kai and U, Kam Hou and Saxena, Karan and Prasad, Karthik and Khandelwal, Kartikay and Zand, Katayoun and Matosich, Kathy and Veeraraghavan, Kaushik and Michelena, Kelly and Li, Keqian and Huang, Kun and Chawla, Kunal and Lakhotia, Kushal and Huang, Kyle and Chen, Lailin and Garg, Lakshya and A, Lavender and Silva, Leandro and Bell, Lee and Zhang, Lei and Guo, Liangpeng and Yu, Licheng and Moshkovich, Liron and Wehrstedt, Luca and Khabsa, Madian and Avalani, Manav and Bhatt, Manish and Tsimpoukelli, Maria and Mankus, Martynas and Hasson, Matan and Lennie, Matthew and Reso, Matthias and Groshev, Maxim and Naumov, Maxim and Lathi, Maya and Keneally, Meghan and Seltzer, Michael L. and Valko, Michal and Restrepo, Michelle and Patel, Mihir and Vyatskov, Mik and Samvelyan, Mikayel and Clark, Mike and Macey, Mike and Wang, Mike and Hermoso, Miquel Jubert and Metanat, Mo and Rastegari, Mohammad and Bansal, Munish and Santhanam, Nandhini and Parks, Natascha and White, Natasha and Bawa, Navyata and Singhal, Nayan and Egebo, Nick and Usunier, Nicolas and Laptev, Nikolay Pavlovich and Dong, Ning and Zhang, Ning and Cheng, Norman and Chernoguz, Oleg and Hart, Olivia and Salpekar, Omkar and Kalinli, Ozlem and Kent, Parkin and Parekh, Parth and Saab, Paul and Balaji, Pavan and Rittner, Pedro and Bontrager, Philip and Roux, Pierre and Dollar, Piotr and Zvyagina, Polina and Ratanchandani, Prashant and Yuvraj, Pritish and Liang, Qian and Alao, Rachad and Rodriguez, Rachel and Ayub, Rafi and Murthy, Raghotham and Nayani, Raghu and Mitra, Rahul and Li, Raymond and Hogan, Rebekkah and Battey, Robin and Wang, Rocky and Maheswari, Rohan and Howes, Russ and Rinott, Ruty and Bondu, Sai Jayesh and Datta, Samyak and Chugh, Sara and Hunt, Sara and Dhillon, Sargun and Sidorov, Sasha and Pan, Satadru and Verma, Saurabh and Yamamoto, Seiji and Ramaswamy, Sharadh and Lindsay, Shaun and Lindsay, Shaun and Feng, Sheng and Lin, Shenghao and Zha, Shengxin Cindy and Shankar, Shiva and Zhang, Shuqiang and Zhang, Shuqiang and Wang, Sinong and Agarwal, Sneha and Sajuyigbe, Soji and Chintala, Soumith and Max, Stephanie and Chen, Stephen and Kehoe, Steve and Satterfield, Steve and Govindaprasad, Sudarshan and Gupta, Sumit and Cho, Sungmin and Virk, Sunny and Subramanian, Suraj and Choudhury, Sy and Goldman, Sydney and Remez, Tal and Glaser, Tamar and Best, Tamara and Kohler, Thilo and Robinson, Thomas and Li, Tianhe and Zhang, Tianjun and Matthews, Tim and Chou, Timothy and Shaked, Tzook and Vontimitta, Varun and Ajayi, Victoria and Montanez, Victoria and Mohan, Vijai and Kumar, Vinay Satish and Mangla, Vishal and Albiero, V{\'i}tor and Ionescu, Vlad and Poenaru, Vlad and Mihailescu, Vlad Tiberiu and Ivanov, Vladimir and Li, Wei and Wang, Wenchen and Jiang, Wenwen and Bouaziz, Wes and Constable, Will and Tang, Xiaocheng and Wang, Xiaofang and Wu, Xiaojian and Wang, Xiaolan and Xia, Xide and Wu, Xilun and Gao, Xinbo and Chen, Yanjun and Hu, Ye and Jia, Ye and Qi, Ye and Li, Yenda and Zhang, Yilin and Zhang, Ying and Adi, Yossi and Nam, Youngjin and Yu and Wang and Hao, Yuchen and Qian, Yundi and He, Yuzi and Rait, Zach and DeVito, Zachary and Rosnbrick, Zef and Wen, Zhaoduo and Yang, Zhenyu and Zhao, Zhiwei},
  month         = aug,
  title         = {The {{Llama}} 3 {{Herd}} of {{Models}}},
  year          = {2024},
  archiveprefix = {arXiv},
  doi           = {10.48550/arXiv.2407.21783},
  eprint        = {2407.21783},
  number        = {arXiv:2407.21783},
  primaryclass  = {cs},
  publisher     = {arXiv},
  urldate       = {2024-08-27},
}

@Article{bae_if_2022,
  author  = {Bae, Juhan and Ng, Nathan and Lo, Alston and Ghassemi, Marzyeh and Grosse, Roger B.},
  journal = {Advances in Neural Information Processing Systems},
  title   = {If {{Influence Functions}} Are the {{Answer}}, {{Then What}} Is the {{Question}}?},
  year    = {2022},
  month   = dec,
  pages   = {17953--17967},
  volume  = {35},
  langid  = {english},
  url     = {https://proceedings.neurips.cc/paper\_files/paper/2022/hash/7234e0c36fdbcb23e7bd56b68838999b-Abstract-Conference.html},
  urldate = {2024-07-18},
}

@InProceedings{huh_position_2024,
  author     = {Huh, Minyoung and Cheung, Brian and Wang, Tongzhou and Isola, Phillip},
  booktitle  = {Proceedings of the 41st {{International Conference}} on {{Machine Learning}}},
  title      = {Position: {{The Platonic Representation Hypothesis}}},
  year       = {2024},
  month      = jul,
  pages      = {20617--20642},
  publisher  = {PMLR},
  issn       = {2640-3498},
  langid     = {english},
  shorttitle = {Position},
  url        = {https://proceedings.mlr.press/v235/huh24a.html},
  urldate    = {2026-02-24},
}

@Article{nakagawa_general_2013,
  author   = {Nakagawa, Shinichi and Schielzeth, Holger},
  journal  = {Methods in Ecology and Evolution},
  title    = {A General and Simple Method for Obtaining {{R2}} from Generalized Linear Mixed-Effects Models},
  year     = {2013},
  issn     = {2041-210X},
  number   = {2},
  pages    = {133--142},
  volume   = {4},
  doi      = {10.1111/j.2041-210x.2012.00261.x},
  langid   = {english},
  urldate  = {2026-04-24},
}

@Article{sokal1958statistical,
  author  = {Sokal, R. and Michener, C.},
  journal = {University of Kansas science bulletin},
  title   = {A Statistical Method for Evaluating Systematic Relationships},
  year    = {1958},
  month   = jan,
  url     = {https://www.semanticscholar.org/paper/A-statistical-method-for-evaluating-systematic-Sokal-Michener/0db093335bc3b9445fa5a1a5526d634921d7b59a},
  urldate = {2026-05-24},
}

@Misc{grosse_studying_2023,
  author        = {Grosse, Roger and Bae, Juhan and Anil, Cem and Elhage, Nelson and Tamkin, Alex and Tajdini, Amirhossein and Steiner, Benoit and Li, Dustin and Durmus, Esin and Perez, Ethan and Hubinger, Evan and Luko{\v s}i{\=u}t{\.e}, Kamil{\.e} and Nguyen, Karina and Joseph, Nicholas and McCandlish, Sam and Kaplan, Jared and Bowman, Samuel R.},
  month         = aug,
  title         = {Studying {{Large Language Model Generalization}} with {{Influence Functions}}},
  year          = {2023},
  archiveprefix = {arXiv},
  doi           = {10.48550/arXiv.2308.03296},
  eprint        = {2308.03296},
  number        = {arXiv:2308.03296},
  primaryclass  = {cs},
  publisher     = {arXiv},
  urldate       = {2025-11-17},
}

@Article{sun_enhancing_2026,
  author  = {Sun, Weiwei and Liu, Haokun and Kandpal, Nikhil and Raffel, Colin and Yang, Yiming},
  journal = {Advances in Neural Information Processing Systems},
  title   = {Enhancing {{Training Data Attribution}} with {{Representational Optimization}}},
  year    = {2026},
  month   = apr,
  pages   = {102274--102311},
  volume  = {38},
  langid  = {english},
  url     = {https://proceedings.neurips.cc/paper_files/paper/2025/hash/940a76f6cc2e23867421e1cf1c8da4f3-Abstract-Conference.html},
  urldate = {2026-05-19},
}

@InProceedings{lagus_low-rank_2019,
  author    = {Lagus, Jarkko and Sinkkonen, Janne and Klami, Arto},
  booktitle = {Proceedings of the 23rd {{Conference}} on {{Computational Natural Language Learning}} ({{CoNLL}})},
  title     = {Low-{{Rank Approximations}} of {{Second-Order Document Representations}}},
  year      = {2019},
  address   = {Hong Kong, China},
  editor    = {Bansal, Mohit and Villavicencio, Aline},
  month     = nov,
  pages     = {634--644},
  publisher = {Association for Computational Linguistics},
  doi       = {10.18653/v1/K19-1059},
  urldate   = {2026-03-10},
}

@Misc{yang_qwen3_2025,
  author        = {Yang, An and Li, Anfeng and Yang, Baosong and Zhang, Beichen and Hui, Binyuan and Zheng, Bo and Yu, Bowen and Gao, Chang and Huang, Chengen and Lv, Chenxu and Zheng, Chujie and Liu, Dayiheng and Zhou, Fan and Huang, Fei and Hu, Feng and Ge, Hao and Wei, Haoran and Lin, Huan and Tang, Jialong and Yang, Jian and Tu, Jianhong and Zhang, Jianwei and Yang, Jianxin and Yang, Jiaxi and Zhou, Jing and Zhou, Jingren and Lin, Junyang and Dang, Kai and Bao, Keqin and Yang, Kexin and Yu, Le and Deng, Lianghao and Li, Mei and Xue, Mingfeng and Li, Mingze and Zhang, Pei and Wang, Peng and Zhu, Qin and Men, Rui and Gao, Ruize and Liu, Shixuan and Luo, Shuang and Li, Tianhao and Tang, Tianyi and Yin, Wenbiao and Ren, Xingzhang and Wang, Xinyu and Zhang, Xinyu and Ren, Xuancheng and Fan, Yang and Su, Yang and Zhang, Yichang and Zhang, Yinger and Wan, Yu and Liu, Yuqiong and Wang, Zekun and Cui, Zeyu and Zhang, Zhenru and Zhou, Zhipeng and Qiu, Zihan},
  month         = may,
  title         = {Qwen3 {{Technical Report}}},
  year          = {2025},
  archiveprefix = {arXiv},
  doi           = {10.48550/arXiv.2505.09388},
  eprint        = {2505.09388},
  number        = {arXiv:2505.09388},
  primaryclass  = {cs},
  publisher     = {arXiv},
  urldate       = {2025-12-18},
}

@Misc{zhang_qwen3_2025,
  author     = {Zhang, Yanzhao and Li, Mingxin and Long, Dingkun and Zhang, Xin and Lin, Huan and Yang, Baosong and Xie, Pengjun and Yang, An and Liu, Dayiheng and Lin, Junyang and Huang, Fei and Zhou, Jingren},
  title      = {Qwen3 {{Embedding}}: {{Advancing Text Embedding}} and {{Reranking Through Foundation Models}}},
  year       = {2025},
  copyright  = {arXiv.org perpetual, non-exclusive license},
  doi        = {10.48550/ARXIV.2506.05176},
  publisher  = {arXiv},
  shorttitle = {Qwen3 {{Embedding}}},
  urldate    = {2026-05-21},
}

@InProceedings{sun_evaluating_2025,
  author    = {Sun, Jingyi and Atanasova, Pepa and Augenstein, Isabelle},
  booktitle = {Proceedings of the 2025 {{Conference}} of the {{Nations}} of the {{Americas Chapter}} of the {{Association}} for {{Computational Linguistics}}: {{Human Language Technologies}} ({{Volume}} 1: {{Long Papers}})},
  title     = {Evaluating {{Input Feature Explanations}} through a {{Unified Diagnostic Evaluation Framework}}},
  year      = {2025},
  address   = {Albuquerque, New Mexico},
  editor    = {Chiruzzo, Luis and Ritter, Alan and Wang, Lu},
  month     = apr,
  pages     = {10559--10577},
  publisher = {Association for Computational Linguistics},
  doi       = {10.18653/v1/2025.naacl-long.530},
  isbn      = {979-8-89176-189-6},
  urldate   = {2025-06-11},
}

@InProceedings{kornblith_similarity_2019,
  author    = {Kornblith, Simon and Norouzi, Mohammad and Lee, Honglak and Hinton, Geoffrey},
  booktitle = {Proceedings of the 36th {{International Conference}} on {{Machine Learning}}},
  title     = {Similarity of {{Neural Network Representations Revisited}}},
  year      = {2019},
  month     = may,
  pages     = {3519--3529},
  publisher = {PMLR},
  issn      = {2640-3498},
  langid    = {english},
  url       = {https://proceedings.mlr.press/v97/kornblith19a.html},
  urldate   = {2026-02-24},
}

@InProceedings{pruthi_estimating_2020,
  author    = {Pruthi, Garima and Liu, Frederick and Kale, Satyen and Sundararajan, Mukund},
  booktitle = {Advances in {{Neural Information Processing Systems}}},
  title     = {Estimating {{Training Data Influence}} by {{Tracing Gradient Descent}}},
  year      = {2020},
  pages     = {19920--19930},
  publisher = {Curran Associates, Inc.},
  volume    = {33},
  url       = {https://proceedings.neurips.cc/paper/2020/hash/e6385d39ec9394f2f3a354d9d2b88eec-Abstract.html},
  urldate   = {2024-07-18},
}

@InProceedings{hong_better_2025,
  author     = {Hong, Dat Minh and Mlodozeniec, Bruno Kacper and Eschenhagen, Runa and Turner, Richard E.},
  booktitle  = {Mechanistic {{Interpretability Workshop}} at {{NeurIPS}} 2025},
  title      = {Better {{Hessians Matter}}: {{Studying}} the {{Impact}} of {{Curvature Approximations}} in {{Influence Functions}}},
  year       = {2025},
  month      = sep,
  langid     = {english},
  shorttitle = {Better {{Hessians Matter}}},
  url        = {https://openreview.net/forum?id=383i8264it},
  urldate    = {2026-05-19},
}

@InProceedings{jain_attention_2019,
  author    = {Jain, Sarthak and Wallace, Byron C.},
  booktitle = {Proceedings of the 2019 {{Conference}} of the {{North American Chapter}} of the {{Association}} for {{Computational Linguistics}}: {{Human Language Technologies}}, {{Volume}} 1 ({{Long}} and {{Short Papers}})},
  title     = {Attention Is Not {{Explanation}}},
  year      = {2019},
  address   = {Minneapolis, Minnesota},
  editor    = {Burstein, Jill and Doran, Christy and Solorio, Thamar},
  month     = jun,
  pages     = {3543--3556},
  publisher = {Association for Computational Linguistics},
  doi       = {10.18653/v1/N19-1357},
  urldate   = {2025-01-13},
}

@Misc{babakhin_llama-embed-nemotron-8b_2025,
  author     = {Babakhin, Yauhen and Osmulski, Radek and Ak, Ronay and Moreira, Gabriel and Xu, Mengyao and Schifferer, Benedikt and Liu, Bo and Oldridge, Even},
  title      = {Llama-{{Embed-Nemotron-8B}}: {{A Universal Text Embedding Model}} for {{Multilingual}} and {{Cross-Lingual Tasks}}},
  year       = {2025},
  copyright  = {Creative Commons Attribution Non Commercial No Derivatives 4.0 International},
  doi        = {10.48550/ARXIV.2511.07025},
  publisher  = {arXiv},
  shorttitle = {Llama-{{Embed-Nemotron-8B}}},
  urldate    = {2026-05-21},
}

@Article{hammoudeh_training_2024,
  author     = {Hammoudeh, Zayd and Lowd, Daniel},
  journal    = {Machine Learning},
  title      = {Training Data Influence Analysis and Estimation: A Survey},
  year       = {2024},
  issn       = {1573-0565},
  month      = may,
  number     = {5},
  pages      = {2351--2403},
  volume     = {113},
  doi        = {10.1007/s10994-023-06495-7},
  langid     = {english},
  shorttitle = {Training Data Influence Analysis and Estimation},
  urldate    = {2024-07-18},
}

@InProceedings{mlodozeniec_influence_2024,
  author    = {Mlodozeniec, Bruno Kacper and Eschenhagen, Runa and Bae, Juhan and Immer, Alexander and Krueger, David and Turner, Richard E.},
  booktitle = {The {{Thirteenth International Conference}} on {{Learning Representations}}},
  title     = {Influence {{Functions}} for {{Scalable Data Attribution}} in {{Diffusion Models}}},
  year      = {2024},
  month     = oct,
  langid    = {english},
  url       = {https://openreview.net/forum?id=esYrEndGsr},
  urldate   = {2026-05-19},
}

@inproceedings{hu_lora_2021,
  title = {{{LoRA}}: {{Low-Rank Adaptation}} of {{Large Language Models}}},
  shorttitle = {{{LoRA}}},
  booktitle = {International {{Conference}} on {{Learning Representations}}},
  author = {Hu, Edward J. and Shen, Yelong and Wallis, Phillip and {Allen-Zhu}, Zeyuan and Li, Yuanzhi and Wang, Shean and Wang, Lu and Chen, Weizhu},
  year = 2021,
  month = oct,
  url = {https://openreview.net/forum?id=nZeVKeeFYf9},
  urldate = {2026-11-13},
  langid = {english}
}

@Article{yeh_first_2022,
  author  = {Yeh, Chih-Kuan and Taly, Ankur and Sundararajan, Mukund and Liu, Frederick and Ravikumar, Pradeep},
  journal = {Advances in Neural Information Processing Systems},
  title   = {First Is {{Better Than Last}} for {{Language Data Influence}}},
  year    = {2022},
  month   = dec,
  pages   = {32285--32298},
  volume  = {35},
  langid  = {english},
  url     = {https://proceedings.neurips.cc/paper_files/paper/2022/hash/d07022783ff6f7bf7a288c207b7dcbd1-Abstract-Conference.html},
  urldate = {2026-06-24},
}

@InCollection{neely_song_2022,
  author     = {Neely, Michael and Schouten, Stefan F. and Bleeker, Maurits and Lucic, Ana},
  booktitle  = {{{HHAI2022}}: {{Augmenting Human Intellect}}},
  title      = {A {{Song}} of ({{Dis}})Agreement: {{Evaluating}} the {{Evaluation}} of {{Explainable Artificial Intelligence}} in {{Natural Language Processing}}},
  doi        = {10.3233/FAIA220190},
  pages      = {60--78},
  publisher  = {IOS Press},
  copyright  = {All rights reserved},
  shorttitle = {A {{Song}} of ({{Dis}})Agreement},
  year       = {2022},
}

@Misc{olmo_olmo_2025,
  author        = {Ettinger, Allyson and Bertsch, Amanda and Kuehl, Bailey and Graham, David and Heineman, David and Groeneveld, Dirk and Brahman, Faeze and Timbers, Finbarr and Ivison, Hamish and Morrison, Jacob and Poznanski, Jake and Lo, Kyle and Soldaini, Luca and Jordan, Matt and Chen, Mayee and Noukhovitch, Michael and Lambert, Nathan and Walsh, Pete and Dasigi, Pradeep and Berry, Robert and Malik, Saumya and Shah, Saurabh and Geng, Scott and Arora, Shane and Gupta, Shashank and Anderson, Taira and Xiao, Teng and Murray, Tyler and Romero, Tyler and Graf, Victoria and Asai, Akari and Bhagia, Akshita and Wettig, Alexander and Liu, Alisa and Rangapur, Aman and Anastasiades, Chloe and Huang, Costa and Schwenk, Dustin and Trivedi, Harsh and Magnusson, Ian and Lochner, Jaron and Liu, Jiacheng and Miranda, Lester James V. and Sap, Maarten and Morgan, Malia and Schmitz, Michael and Guerquin, Michal and Wilson, Michael and Huff, Regan and Bras, Ronan Le and Xin, Rui and Shao, Rulin and Skjonsberg, Sam and Shen, Shannon Zejiang and Li, Shuyue Stella and Wilde, Tucker and Pyatkin, Valentina and Merrill, Will and Chang, Yapei and Gu, Yuling and Zeng, Zhiyuan and Sabharwal, Ashish and Zettlemoyer, Luke and Koh, Pang Wei and Farhadi, Ali and Smith, Noah A. and Hajishirzi, Hannaneh},
  month         = dec,
  title         = {Olmo 3},
  year          = {2025},
  archiveprefix = {arXiv},
  doi           = {10.48550/arXiv.2512.13961},
  eprint        = {2512.13961},
  number        = {arXiv:2512.13961},
  primaryclass  = {cs},
  publisher     = {arXiv},
  urldate       = {2026-03-06},
}

@InProceedings{koh_understanding_2017,
  author    = {Koh, Pang Wei and Liang, Percy},
  booktitle = {Proceedings of the 34th {{International Conference}} on {{Machine Learning}}},
  title     = {Understanding {{Black-box Predictions}} via {{Influence Functions}}},
  year      = {2017},
  month     = jul,
  pages     = {1885--1894},
  publisher = {PMLR},
  issn      = {2640-3498},
  langid    = {english},
  url       = {https://proceedings.mlr.press/v70/koh17a.html},
  urldate   = {2024-07-16},
}

@Misc{deng_survey_2025,
  author        = {Deng, Junwei and Hu, Yuzheng and Hu, Pingbang and Li, Ting-wei and Liu, Shixuan and Wang, Jiachen T. and Ley, Dan and Dai, Qirun and Huang, Benhao and Huang, Jin and Jiao, Cathy and Just, Hoang Anh and Pan, Yijun and Shen, Jingyan and Tu, Yiwen and Wang, Weiyi and Wang, Xinhe and Zhang, Shichang and Zhang, Shiyuan and Jia, Ruoxi and Lakkaraju, Himabindu and Peng, Hao and Tang, Weijing and Xiong, Chenyan and Zhao, Jieyu and Tong, Hanghang and Zhao, Han and Ma, Jiaqi W.},
  month         = sep,
  title         = {A {{Survey}} of {{Data Attribution}}: {{Methods}}, {{Applications}}, and {{Evaluation}} in the {{Era}} of {{Generative AI}}},
  year          = {2025},
  address       = {Rochester, NY},
  archiveprefix = {Social Science Research Network},
  doi           = {10.2139/ssrn.5451054},
  eprint        = {5451054},
  langid        = {english},
  number        = {5451054},
  publisher     = {Social Science Research Network},
  shorttitle    = {A {{Survey}} of {{Data Attribution}}},
  type          = {{{SSRN Scholarly Paper}}},
  urldate       = {2025-12-02},
}

@InProceedings{park_trak_2023,
  author     = {Park, Sung Min and Georgiev, Kristian and Ilyas, Andrew and Leclerc, Guillaume and Madry, Aleksander},
  booktitle  = {Proceedings of the 40th {{International Conference}} on {{Machine Learning}}},
  title      = {{{TRAK}}: {{Attributing Model Behavior}} at {{Scale}}},
  year       = {2023},
  month      = jul,
  pages      = {27074--27113},
  publisher  = {PMLR},
  issn       = {2640-3498},
  langid     = {english},
  shorttitle = {{{TRAK}}},
  url        = {https://proceedings.mlr.press/v202/park23c.html},
  urldate    = {2024-07-31},
}

@InProceedings{han_explaining_2020,
  author    = {Han, Xiaochuang and Wallace, Byron C. and Tsvetkov, Yulia},
  booktitle = {Proceedings of the 58th {{Annual Meeting}} of the {{Association}} for {{Computational Linguistics}}},
  title     = {Explaining {{Black Box Predictions}} and {{Unveiling Data Artifacts}} through {{Influence Functions}}},
  year      = {2020},
  address   = {Online},
  editor    = {Jurafsky, Dan and Chai, Joyce and Schluter, Natalie and Tetreault, Joel},
  month     = jul,
  pages     = {5553--5563},
  publisher = {Association for Computational Linguistics},
  doi       = {10.18653/v1/2020.acl-main.492},
  urldate   = {2024-07-18},
}

@InProceedings{pezeshkpour_empirical_2021,
  author    = {Pezeshkpour, Pouya and Jain, Sarthak and Wallace, Byron and Singh, Sameer},
  booktitle = {Proceedings of the 2021 {{Conference}} of the {{North American Chapter}} of the {{Association}} for {{Computational Linguistics}}: {{Human Language Technologies}}},
  title     = {An {{Empirical Comparison}} of {{Instance Attribution Methods}} for {{NLP}}},
  year      = {2021},
  address   = {Online},
  editor    = {Toutanova, Kristina and Rumshisky, Anna and Zettlemoyer, Luke and {Hakkani-Tur}, Dilek and Beltagy, Iz and Bethard, Steven and Cotterell, Ryan and Chakraborty, Tanmoy and Zhou, Yichao},
  month     = jun,
  pages     = {967--975},
  publisher = {Association for Computational Linguistics},
  doi       = {10.18653/v1/2021.naacl-main.75},
  urldate   = {2025-12-03},
}

@InProceedings{lambert_tulu_2025-1,
  author     = {Lambert, Nathan and Morrison, Jacob and Pyatkin, Valentina and Huang, Shengyi and Ivison, Hamish and Brahman, Faeze and Miranda, Lester James Validad and Liu, Alisa and Dziri, Nouha and Lyu, Xinxi and Gu, Yuling and Malik, Saumya and Graf, Victoria and Hwang, Jena D. and Yang, Jiangjiang and Bras, Ronan Le and Tafjord, Oyvind and Wilhelm, Christopher and Soldaini, Luca and Smith, Noah A. and Wang, Yizhong and Dasigi, Pradeep and Hajishirzi, Hannaneh},
  booktitle  = {Second {{Conference}} on {{Language Modeling}}},
  title      = {Tulu 3: {{Pushing Frontiers}} in {{Open Language Model Post-Training}}},
  year       = {2025},
  month      = aug,
  langid     = {english},
  shorttitle = {Tulu 3},
  url        = {https://openreview.net/forum?id=i1uGbfHHpH#discussion},
  urldate    = {2026-04-09},
}

@InProceedings{rajpurkar_know_2018,
  author     = {Rajpurkar, Pranav and Jia, Robin and Liang, Percy},
  booktitle  = {Proceedings of the 56th {{Annual Meeting}} of the {{Association}} for {{Computational Linguistics}} ({{Volume}} 2: {{Short Papers}})},
  title      = {Know {{What You Don}}'t {{Know}}: {{Unanswerable Questions}} for {{SQuAD}}},
  year       = {2018},
  address    = {Melbourne, Australia},
  editor     = {Gurevych, Iryna and Miyao, Yusuke},
  month      = jul,
  pages      = {784--789},
  publisher  = {Association for Computational Linguistics},
  doi        = {10.18653/v1/P18-2124},
  shorttitle = {Know {{What You Don}}'t {{Know}}},
  urldate    = {2026-03-05},
}

@InProceedings{hanawa_evaluation_2020,
  author    = {Hanawa, Kazuaki and Yokoi, Sho and Hara, Satoshi and Inui, Kentaro},
  booktitle = {International {{Conference}} on {{Learning Representations}}},
  title     = {Evaluation of {{Similarity-based Explanations}}},
  year      = {2020},
  month     = oct,
  langid    = {english},
  url       = {https://openreview.net/forum?id=9uvhpyQwzM_},
  urldate   = {2024-08-01},
}

@InProceedings{kwon_datainf_2023,
  author     = {Kwon, Yongchan and Wu, Eric and Wu, Kevin and Zou, James},
  booktitle  = {The {{Twelfth International Conference}} on {{Learning Representations}}},
  title      = {{{DataInf}}: {{Efficiently Estimating Data Influence}} in {{LoRA-tuned LLMs}} and {{Diffusion Models}}},
  year       = {2023},
  month      = oct,
  langid     = {english},
  shorttitle = {{{DataInf}}},
  url        = {https://openreview.net/forum?id=9m02ib92Wz},
  urldate    = {2026-02-05},
}

@InProceedings{li_influence_2025,
  author    = {Li, Zhe and Zhao, Wei and Li, Yige and Sun, Jun},
  booktitle = {Findings of the {{Association}} for {{Computational Linguistics}}: {{EMNLP}} 2025},
  title     = {Do {{Influence Functions Work}} on {{Large Language Models}}?},
  year      = {2025},
  address   = {Suzhou, China},
  editor    = {Christodoulopoulos, Christos and Chakraborty, Tanmoy and Rose, Carolyn and Peng, Violet},
  month     = nov,
  pages     = {14367--14382},
  publisher = {Association for Computational Linguistics},
  isbn      = {979-8-89176-335-7},
  url       = {https://aclanthology.org/2025.findings-emnlp.775/},
  urldate   = {2025-11-08},
}

@Article{sokal_comparison_1962,
  author     = {Sokal, Robert R. and Rohlf, F. James},
  journal    = {Taxon},
  title      = {The {{Comparison}} of {{Dendrograms}} by {{Objective Methods}}},
  year       = {1962},
  issn       = {0040-0262},
  number     = {2},
  pages      = {33--40},
  volume     = {11},
  doi        = {10.2307/1217208},
  eprint     = {1217208},
  eprinttype = {jstor},
  publisher  = {International Association for Plant Taxonomy (IAPT)},
  urldate    = {2026-04-30},
}

@InProceedings{walsh_2_2025,
  author    = {Walsh, Evan Pete and Soldaini, Luca and Groeneveld, Dirk and Lo, Kyle and Arora, Shane and Bhagia, Akshita and Gu, Yuling and Huang, Shengyi and Jordan, Matt and Lambert, Nathan and Schwenk, Dustin and Tafjord, Oyvind and Anderson, Taira and Atkinson, David and Brahman, Faeze and Clark, Christopher and Dasigi, Pradeep and Dziri, Nouha and Ettinger, Allyson and Guerquin, Michal and Heineman, David and Ivison, Hamish and Koh, Pang Wei and Liu, Jiacheng and Malik, Saumya and Merrill, William and Miranda, Lester James Validad and Morrison, Jacob and Murray, Tyler and Nam, Crystal and Poznanski, Jake and Pyatkin, Valentina and Rangapur, Aman and Schmitz, Michael and Skjonsberg, Sam and Wadden, David and Wilhelm, Christopher and Wilson, Michael and Zettlemoyer, Luke and Farhadi, Ali and Smith, Noah A. and Hajishirzi, Hannaneh},
  booktitle = {Second {{Conference}} on {{Language Modeling}}},
  title     = {2 {{OLMo}} 2 {{Furious}} ({{COLM}}'s {{Version}})},
  year      = {2025},
  month     = aug,
  langid    = {english},
  url       = {https://openreview.net/forum?id=2ezugTT9kU#discussion},
  urldate   = {2026-04-09},
}

@InProceedings{jiao_feasibility_2025,
  author    = {Jiao, Cathy and Gao, Weizhen and Raghunathan, Aditi and Xiong, Chenyan},
  booktitle = {Findings of the {{Association}} for {{Computational Linguistics}}: {{NAACL}} 2025},
  title     = {On the {{Feasibility}} of {{In-Context Probing}} for {{Data Attribution}}},
  year      = {2025},
  address   = {Albuquerque, New Mexico},
  editor    = {Chiruzzo, Luis and Ritter, Alan and Wang, Lu},
  month     = apr,
  pages     = {5140--5155},
  publisher = {Association for Computational Linguistics},
  doi       = {10.18653/v1/2025.findings-naacl.286},
  isbn      = {979-8-89176-195-7},
  urldate   = {2026-02-04},
}

@InProceedings{akyurek_towards_2022,
  author    = {Akyurek, Ekin and Bolukbasi, Tolga and Liu, Frederick and Xiong, Binbin and Tenney, Ian and Andreas, Jacob and Guu, Kelvin},
  booktitle = {Findings of the {{Association}} for {{Computational Linguistics}}: {{EMNLP}} 2022},
  title     = {Towards {{Tracing Knowledge}} in {{Language Models Back}} to the {{Training Data}}},
  year      = {2022},
  address   = {Abu Dhabi, United Arab Emirates},
  editor    = {Goldberg, Yoav and Kozareva, Zornitsa and Zhang, Yue},
  month     = dec,
  pages     = {2429--2446},
  publisher = {Association for Computational Linguistics},
  doi       = {10.18653/v1/2022.findings-emnlp.180},
  urldate   = {2024-08-13},
}

@Article{brophy_adapting_2024,
  author   = {Brophy, Jonathan and Hammoudeh, Zayd and Lowd, Daniel},
  journal  = {J. Mach. Learn. Res.},
  title    = {Adapting and Evaluating Influence-Estimation Methods for Gradient-Boosted Decision Trees},
  year     = {2024},
  issn     = {1532-4435},
  month    = mar,
  number   = {1},
  pages    = {154:7325--154:7372},
  volume   = {24},
  url      = {https://www.jmlr.org/papers/v24/22-0449.html},
}

\appendix
\newpage
\FloatBarrier
\section{Cophenetic Stability}\label{appendix:cophstab}
\FloatBarrier
\begin{table}[htb]
    \centering
    \resizebox{\linewidth}{!}{%
    \begin{tabular}{llllllll}
\toprule
 & \rotatebox{45}{Llama3.2-1B} & \rotatebox{45}{OLMo2-1B} & \rotatebox{45}{OLMo3-7B} & \rotatebox{45}{Qwen3-0.6B} & \rotatebox{45}{Qwen3-1.7B} & \rotatebox{45}{Qwen3-4B} & \rotatebox{45}{Qwen3-8B} \\
\midrule
CKA & {\cellcolor[HTML]{404588}} \color[HTML]{F1F1F1} 0.80 & {\cellcolor[HTML]{1E9D89}} \color[HTML]{F1F1F1} 0.82 & {\cellcolor[HTML]{424086}} \color[HTML]{F1F1F1} 0.80 & {\cellcolor[HTML]{FDE725}} \color[HTML]{000000} 0.84 & {\cellcolor[HTML]{4AC16D}} \color[HTML]{000000} 0.83 & {\cellcolor[HTML]{375B8D}} \color[HTML]{F1F1F1} 0.81 & {\cellcolor[HTML]{440154}} \color[HTML]{F1F1F1} 0.79 \\
CKNNA@25 & {\cellcolor[HTML]{FDE725}} \color[HTML]{000000} 0.86 & {\cellcolor[HTML]{375B8D}} \color[HTML]{F1F1F1} 0.76 & {\cellcolor[HTML]{46307E}} \color[HTML]{F1F1F1} 0.74 & {\cellcolor[HTML]{38598C}} \color[HTML]{F1F1F1} 0.76 & {\cellcolor[HTML]{472F7D}} \color[HTML]{F1F1F1} 0.74 & {\cellcolor[HTML]{440154}} \color[HTML]{F1F1F1} 0.72 & {\cellcolor[HTML]{404588}} \color[HTML]{F1F1F1} 0.75 \\
$\tau$ & {\cellcolor[HTML]{440154}} \color[HTML]{F1F1F1} 0.74 & {\cellcolor[HTML]{70CF57}} \color[HTML]{000000} 0.81 & {\cellcolor[HTML]{404588}} \color[HTML]{F1F1F1} 0.76 & {\cellcolor[HTML]{FDE725}} \color[HTML]{000000} 0.83 & {\cellcolor[HTML]{86D549}} \color[HTML]{000000} 0.81 & {\cellcolor[HTML]{27AD81}} \color[HTML]{F1F1F1} 0.79 & {\cellcolor[HTML]{482878}} \color[HTML]{F1F1F1} 0.75 \\
$acc_{\text{pred}}$ & {\cellcolor[HTML]{481668}} \color[HTML]{F1F1F1} 0.79 & {\cellcolor[HTML]{440154}} \color[HTML]{F1F1F1} 0.79 & {\cellcolor[HTML]{424186}} \color[HTML]{F1F1F1} 0.80 & {\cellcolor[HTML]{ADDC30}} \color[HTML]{000000} 0.83 & {\cellcolor[HTML]{FDE725}} \color[HTML]{000000} 0.83 & {\cellcolor[HTML]{3F4788}} \color[HTML]{F1F1F1} 0.80 & {\cellcolor[HTML]{481769}} \color[HTML]{F1F1F1} 0.79 \\
RBO & {\cellcolor[HTML]{39568C}} \color[HTML]{F1F1F1} 0.74 & {\cellcolor[HTML]{FDE725}} \color[HTML]{000000} 0.78 & {\cellcolor[HTML]{423F85}} \color[HTML]{F1F1F1} 0.73 & {\cellcolor[HTML]{37B878}} \color[HTML]{F1F1F1} 0.76 & {\cellcolor[HTML]{B5DE2B}} \color[HTML]{000000} 0.78 & {\cellcolor[HTML]{277F8E}} \color[HTML]{F1F1F1} 0.75 & {\cellcolor[HTML]{440154}} \color[HTML]{F1F1F1} 0.72 \\
$\rho$ & {\cellcolor[HTML]{414287}} \color[HTML]{F1F1F1} 0.72 & {\cellcolor[HTML]{C5E021}} \color[HTML]{000000} 0.78 & {\cellcolor[HTML]{481D6F}} \color[HTML]{F1F1F1} 0.71 & {\cellcolor[HTML]{FDE725}} \color[HTML]{000000} 0.79 & {\cellcolor[HTML]{37B878}} \color[HTML]{F1F1F1} 0.76 & {\cellcolor[HTML]{20928C}} \color[HTML]{F1F1F1} 0.75 & {\cellcolor[HTML]{440154}} \color[HTML]{F1F1F1} 0.71 \\
\bottomrule
\end{tabular}
}
    \caption{Average cophenetic correlation across tasks per base model.}
    \label{tab:cocorr-per-base-model}
\end{table}
\begin{table}[htb]
    \centering
    \resizebox{0.75\linewidth}{!}{%
    \begin{tabular}{llllll}
\toprule
 & code & ftrace & math & sft & squad\_v2 \\
\midrule
CKA & {\cellcolor[HTML]{27AD81}} \color[HTML]{F1F1F1} 0.92 & {\cellcolor[HTML]{86D549}} \color[HTML]{000000} 0.95 & {\cellcolor[HTML]{440154}} \color[HTML]{F1F1F1} 0.84 & {\cellcolor[HTML]{90D743}} \color[HTML]{000000} 0.95 & {\cellcolor[HTML]{FDE725}} \color[HTML]{000000} 0.97 \\
CKNNA@25 & {\cellcolor[HTML]{440154}} \color[HTML]{F1F1F1} 0.87 & {\cellcolor[HTML]{7FD34E}} \color[HTML]{000000} 0.96 & {\cellcolor[HTML]{24868E}} \color[HTML]{F1F1F1} 0.92 & {\cellcolor[HTML]{3DBC74}} \color[HTML]{F1F1F1} 0.95 & {\cellcolor[HTML]{FDE725}} \color[HTML]{000000} 0.98 \\
$\tau$ & {\cellcolor[HTML]{70CF57}} \color[HTML]{000000} 0.92 & {\cellcolor[HTML]{38B977}} \color[HTML]{F1F1F1} 0.90 & {\cellcolor[HTML]{440154}} \color[HTML]{F1F1F1} 0.77 & {\cellcolor[HTML]{37B878}} \color[HTML]{F1F1F1} 0.90 & {\cellcolor[HTML]{FDE725}} \color[HTML]{000000} 0.96 \\
$acc_{\text{pred}}$ & {\cellcolor[HTML]{FDE725}} \color[HTML]{000000} 0.99 & {\cellcolor[HTML]{1F988B}} \color[HTML]{F1F1F1} 0.90 & {\cellcolor[HTML]{440154}} \color[HTML]{F1F1F1} 0.80 & {\cellcolor[HTML]{3FBC73}} \color[HTML]{F1F1F1} 0.93 & {\cellcolor[HTML]{A5DB36}} \color[HTML]{000000} 0.96 \\
RBO & {\cellcolor[HTML]{70CF57}} \color[HTML]{000000} 0.91 & {\cellcolor[HTML]{20928C}} \color[HTML]{F1F1F1} 0.86 & {\cellcolor[HTML]{440154}} \color[HTML]{F1F1F1} 0.78 & {\cellcolor[HTML]{4CC26C}} \color[HTML]{000000} 0.90 & {\cellcolor[HTML]{FDE725}} \color[HTML]{000000} 0.94 \\
$\rho$ & {\cellcolor[HTML]{44BF70}} \color[HTML]{F1F1F1} 0.89 & {\cellcolor[HTML]{38B977}} \color[HTML]{F1F1F1} 0.88 & {\cellcolor[HTML]{440154}} \color[HTML]{F1F1F1} 0.74 & {\cellcolor[HTML]{48C16E}} \color[HTML]{F1F1F1} 0.89 & {\cellcolor[HTML]{FDE725}} \color[HTML]{000000} 0.95 \\
\bottomrule
\end{tabular}
}
    \caption{Average cophenetic correlation across base models per task.}
    \label{tab:cocorr-per-task}
\end{table}
\begin{table}[htb]
    \centering
    \resizebox{0.75\linewidth}{!}{%
    \begin{tabular}{lllllll}
\toprule
 & CKA & CKNNA@25 & $\tau$ & $acc_{\text{pred}}$ & RBO & $\rho$ \\
\midrule
CKA & {\cellcolor[HTML]{FDE725}} \color[HTML]{000000} 0.73 & {\cellcolor[HTML]{DFE318}} \color[HTML]{000000} 0.70 & {\cellcolor[HTML]{471063}} \color[HTML]{F1F1F1} 0.05 & {\cellcolor[HTML]{46075A}} \color[HTML]{F1F1F1} 0.03 & {\cellcolor[HTML]{440154}} \color[HTML]{F1F1F1} 0.02 & {\cellcolor[HTML]{471063}} \color[HTML]{F1F1F1} 0.05 \\
CKNNA@25 & {\cellcolor[HTML]{000000}} \color[HTML]{F1F1F1} {\cellcolor{white}} \color{white} nan & {\cellcolor[HTML]{C5E021}} \color[HTML]{000000} 0.67 & {\cellcolor[HTML]{482374}} \color[HTML]{F1F1F1} 0.09 & {\cellcolor[HTML]{481A6C}} \color[HTML]{F1F1F1} 0.07 & {\cellcolor[HTML]{471365}} \color[HTML]{F1F1F1} 0.06 & {\cellcolor[HTML]{482374}} \color[HTML]{F1F1F1} 0.09 \\
$\tau$ & {\cellcolor[HTML]{000000}} \color[HTML]{F1F1F1} {\cellcolor{white}} \color{white} nan & {\cellcolor[HTML]{000000}} \color[HTML]{F1F1F1} {\cellcolor{white}} \color{white} nan & {\cellcolor[HTML]{90D743}} \color[HTML]{000000} 0.61 & {\cellcolor[HTML]{2A768E}} \color[HTML]{F1F1F1} 0.30 & {\cellcolor[HTML]{69CD5B}} \color[HTML]{000000} 0.57 & {\cellcolor[HTML]{8ED645}} \color[HTML]{000000} 0.61 \\
$acc_{\text{pred}}$ & {\cellcolor[HTML]{000000}} \color[HTML]{F1F1F1} {\cellcolor{white}} \color{white} nan & {\cellcolor[HTML]{000000}} \color[HTML]{F1F1F1} {\cellcolor{white}} \color{white} nan & {\cellcolor[HTML]{000000}} \color[HTML]{F1F1F1} {\cellcolor{white}} \color{white} nan & {\cellcolor[HTML]{424186}} \color[HTML]{F1F1F1} 0.16 & {\cellcolor[HTML]{2C728E}} \color[HTML]{F1F1F1} 0.29 & {\cellcolor[HTML]{2B758E}} \color[HTML]{F1F1F1} 0.30 \\
RBO & {\cellcolor[HTML]{000000}} \color[HTML]{F1F1F1} {\cellcolor{white}} \color{white} nan & {\cellcolor[HTML]{000000}} \color[HTML]{F1F1F1} {\cellcolor{white}} \color{white} nan & {\cellcolor[HTML]{000000}} \color[HTML]{F1F1F1} {\cellcolor{white}} \color{white} nan & {\cellcolor[HTML]{000000}} \color[HTML]{F1F1F1} {\cellcolor{white}} \color{white} nan & {\cellcolor[HTML]{48C16E}} \color[HTML]{F1F1F1} 0.52 & {\cellcolor[HTML]{69CD5B}} \color[HTML]{000000} 0.57 \\
$\rho$ & {\cellcolor[HTML]{000000}} \color[HTML]{F1F1F1} {\cellcolor{white}} \color{white} nan & {\cellcolor[HTML]{000000}} \color[HTML]{F1F1F1} {\cellcolor{white}} \color{white} nan & {\cellcolor[HTML]{000000}} \color[HTML]{F1F1F1} {\cellcolor{white}} \color{white} nan & {\cellcolor[HTML]{000000}} \color[HTML]{F1F1F1} {\cellcolor{white}} \color{white} nan & {\cellcolor[HTML]{000000}} \color[HTML]{F1F1F1} {\cellcolor{white}} \color{white} nan & {\cellcolor[HTML]{8ED645}} \color[HTML]{000000} 0.61 \\
\bottomrule
\end{tabular}
}
    \caption{Average cophenetic correlation between alignment measures.}
    \label{tab:cocorr-cross-score}
\end{table}
\FloatBarrier
\section{Full Alignment Scores}\label{appendix:alignment}
\begin{minipage}{\columnwidth}
    \centering
    \includegraphics[width=0.7\columnwidth]{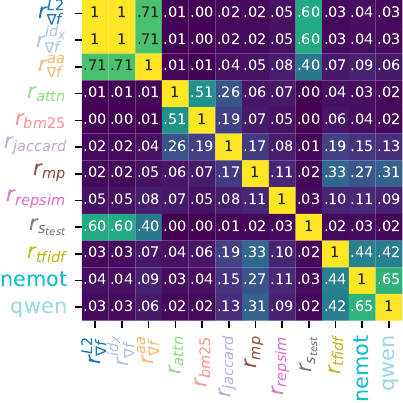}

    \includegraphics[width=0.7\columnwidth]{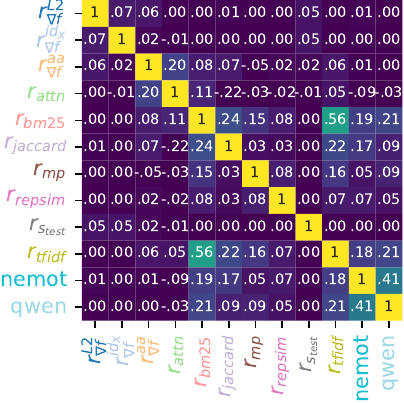}

    \includegraphics[width=0.7\columnwidth]{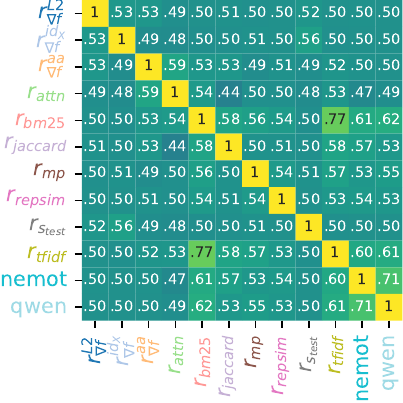}
    \captionsetup{hypcap=false}
    \captionof{figure}{Representational alignment (CKNNA@25: top), and functional alignment (Kendall's $\tau$: center, RBO: bottom), averaged across tasks and models.}
    \label{fig:alignment-appendix}
    \end{minipage}
   
\begin{figure}[htbp]
    \centering

    \includegraphics[width=0.64\linewidth]{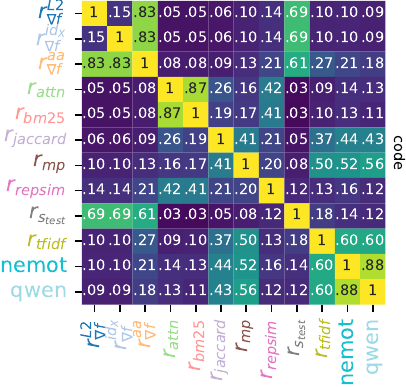}\\

    \includegraphics[width=0.64\linewidth]{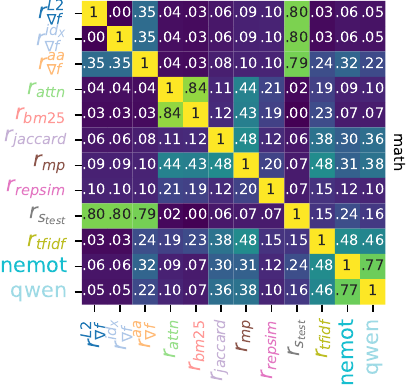}\\

    \includegraphics[width=0.64\linewidth]{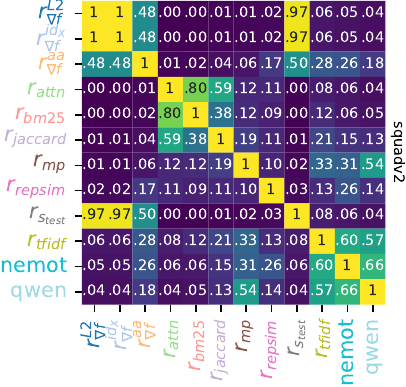}

    \includegraphics[width=0.64\linewidth]{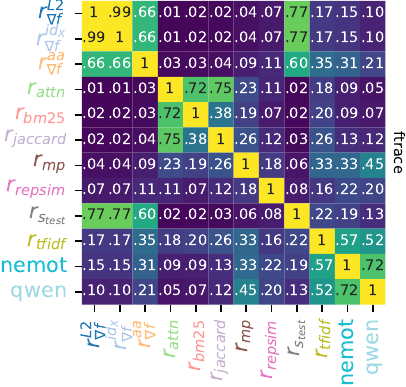}

    \includegraphics[width=0.64\linewidth]{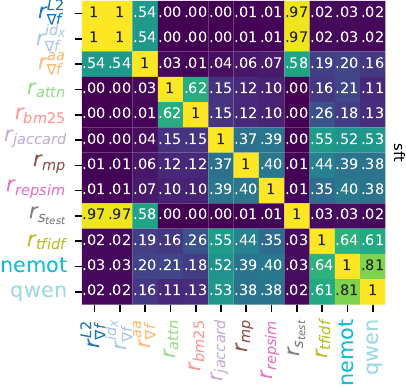}

    \caption{Per-task CKA results.}
    \label{fig:per-task-cka-appendix}
\end{figure}

\begin{figure}[htbp]
    \centering

    \includegraphics[width=0.64\linewidth]{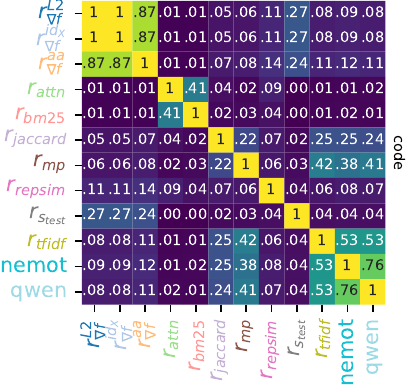}\\

    \includegraphics[width=0.64\linewidth]{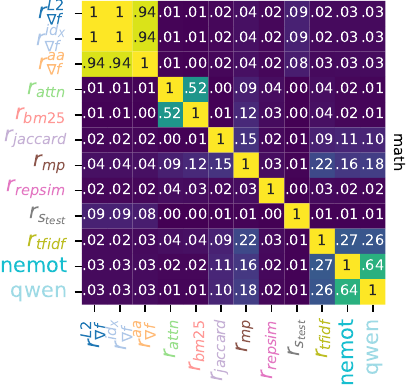}\\

    \includegraphics[width=0.64\linewidth]{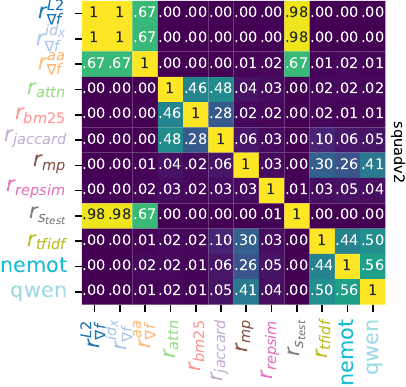}

    \includegraphics[width=0.64\linewidth]{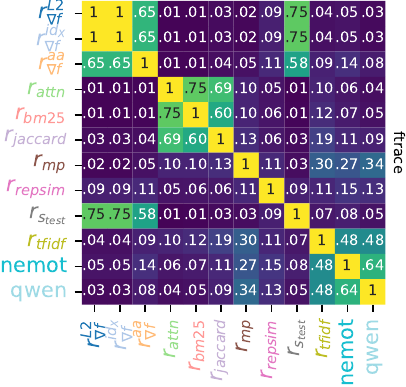}

    \includegraphics[width=0.64\linewidth]{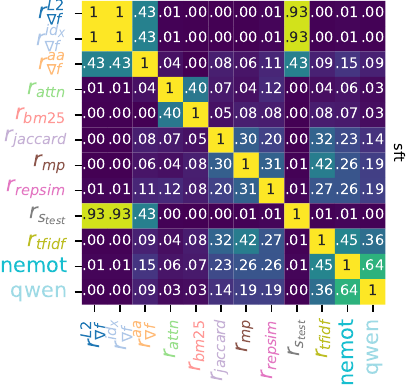}

    \caption{Per-task CKNNA@25 results.}
    \label{fig:per-task-cknna-appendix}
\end{figure}

\begin{figure}[htbp]
    \centering

    \includegraphics[width=0.64\linewidth]{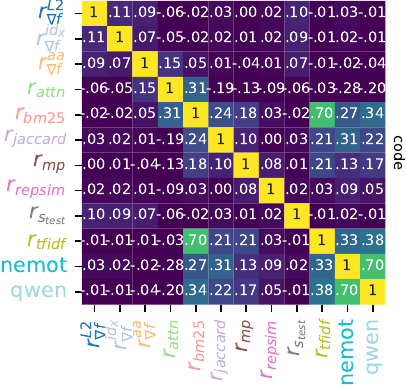}\\

    \includegraphics[width=0.64\linewidth]{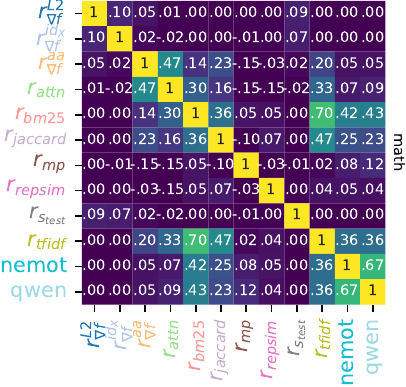}\\

    \includegraphics[width=0.64\linewidth]{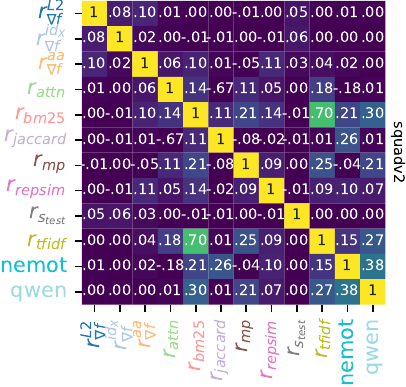}

    \includegraphics[width=0.64\linewidth]{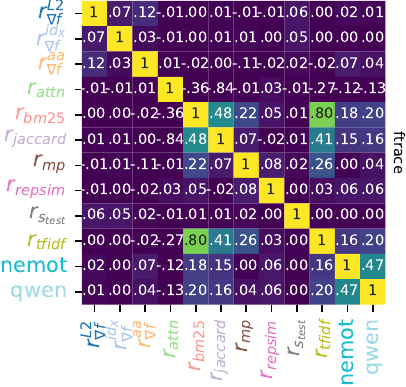}

    \includegraphics[width=0.64\linewidth]{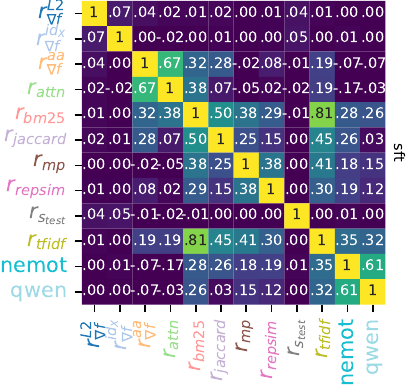}

    \caption{Per-task Spearman's $\rho$ results.}
    \label{fig:per-task-spearman-appendix}
\end{figure}

\begin{figure}[htbp]
    \centering

    \includegraphics[width=0.64\linewidth]{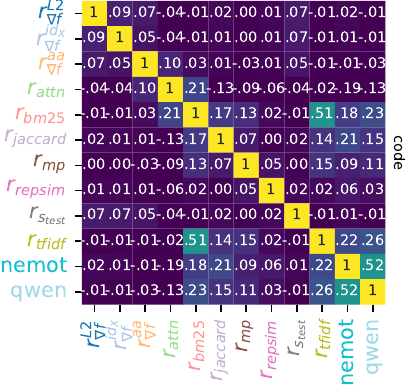}\\

    \includegraphics[width=0.64\linewidth]{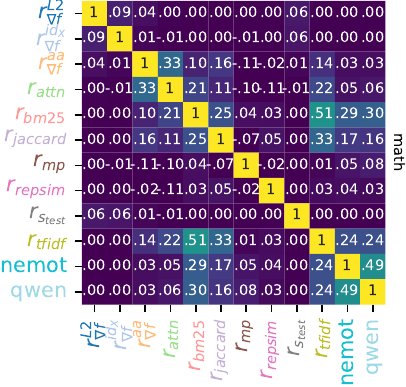}\\

    \includegraphics[width=0.64\linewidth]{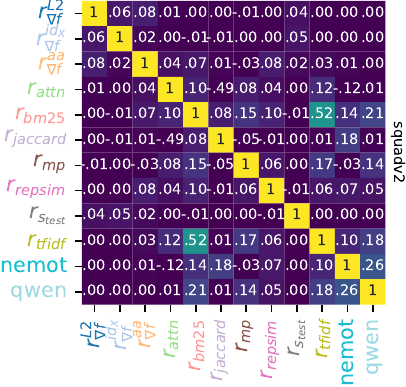}

    \includegraphics[width=0.64\linewidth]{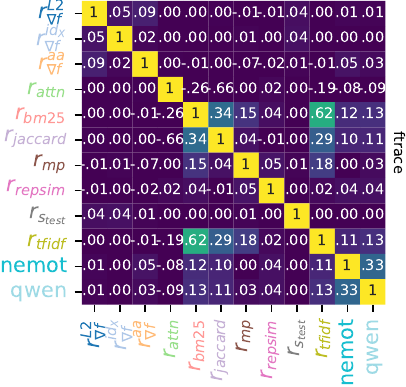}

    \includegraphics[width=0.64\linewidth]{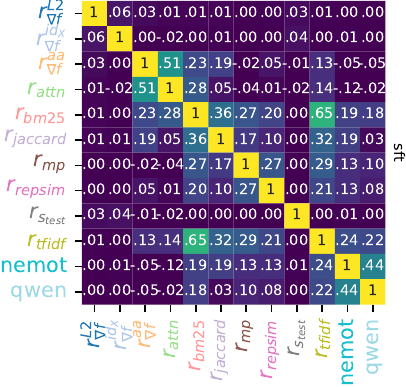}

    \caption{Per-task Kendall's $\tau$ results.}
    \label{fig:per-task-kendall-appendix}
\end{figure}
\begin{figure}[htbp]
    \centering

    \includegraphics[width=0.64\linewidth]{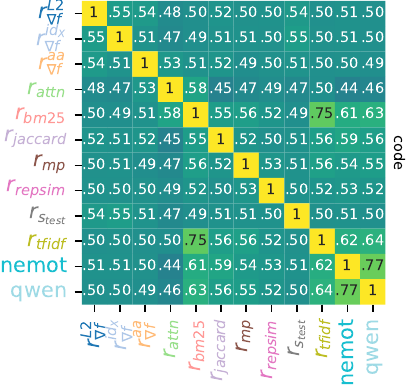}\\

    \includegraphics[width=0.64\linewidth]{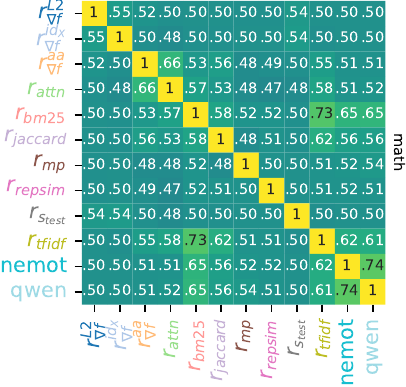}\\

    \includegraphics[width=0.64\linewidth]{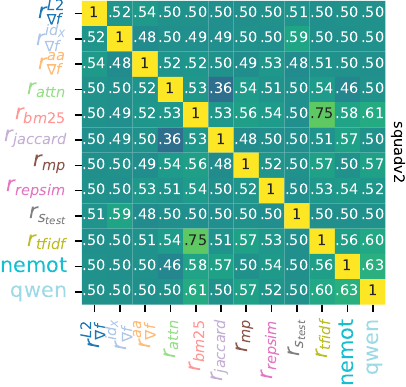}

    \includegraphics[width=0.64\linewidth]{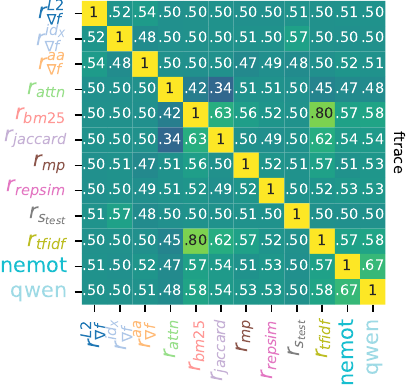}

    \includegraphics[width=0.64\linewidth]{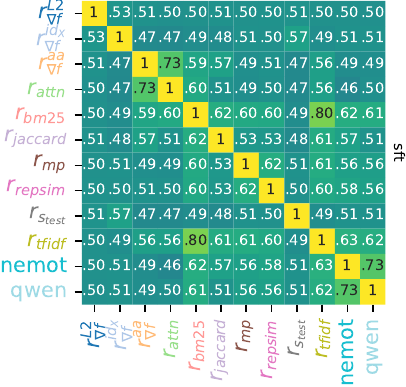}

    \caption{Per-task RBO results.}
    \label{fig:per-task-rbo-appendix}
\end{figure}
\begin{figure}[htbp]
    \centering

    \includegraphics[width=0.64\linewidth]{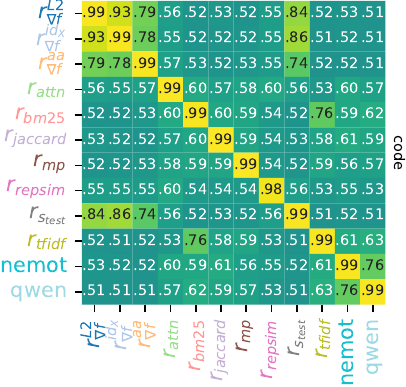}\\

    \includegraphics[width=0.64\linewidth]{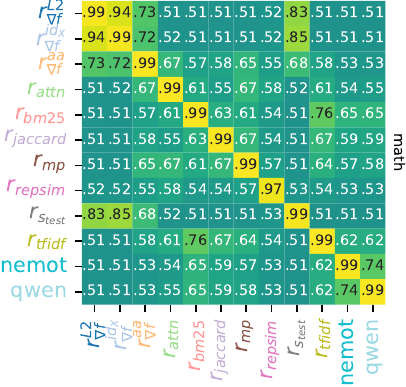}\\

    \includegraphics[width=0.64\linewidth]{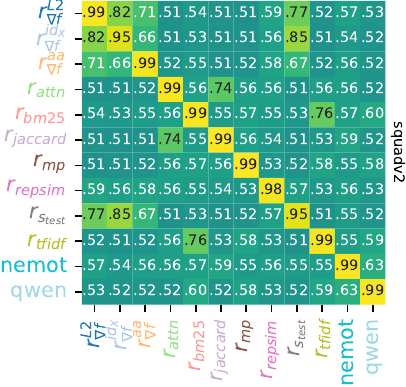}

    \includegraphics[width=0.64\linewidth]{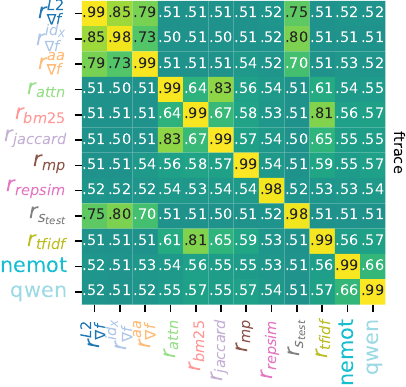}

    \includegraphics[width=0.64\linewidth]{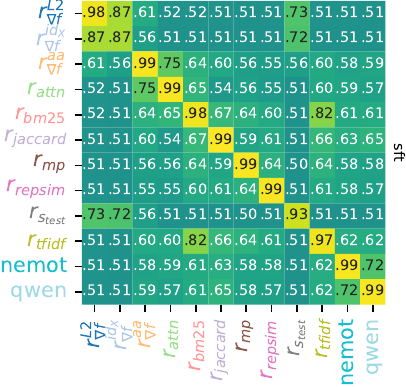}

    \caption{Per-task cross-method predictability results.}
    \label{fig:per-task-pwacc-appendix}
\end{figure}
\onecolumn
\noindent

\begin{minipage}[t]{0.49\textwidth}
\centering
\setlength{\tabcolsep}{1pt} 

\begin{tabular}{@{}cc@{}}
\includegraphics[width=0.485\linewidth]{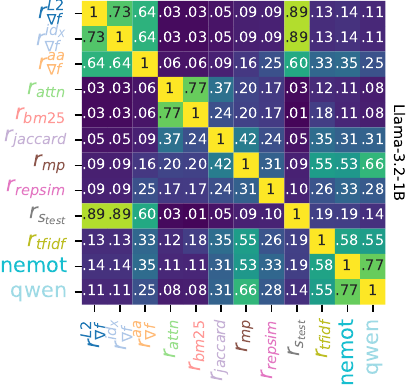} &
\includegraphics[width=0.485\linewidth]{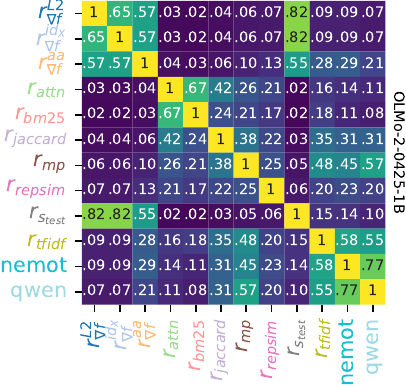} \\

\includegraphics[width=0.485\linewidth]{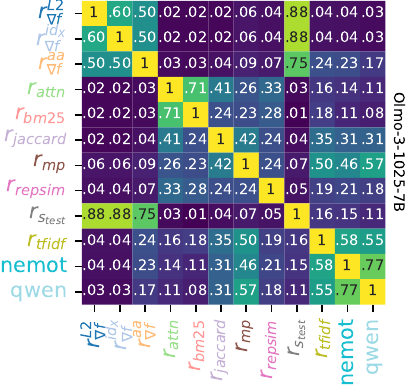} &
\includegraphics[width=0.485\linewidth]{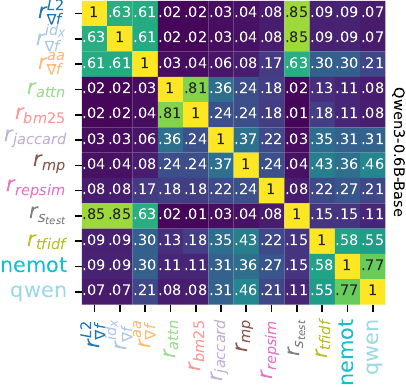} \\

\includegraphics[width=0.485\linewidth]{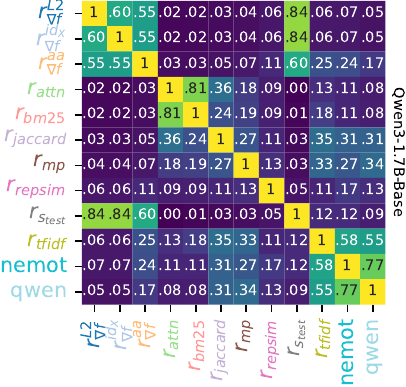} &
\includegraphics[width=0.485\linewidth]{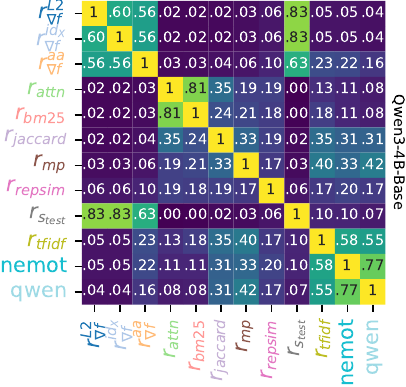} \\

\multicolumn{2}{c}{
\includegraphics[width=0.49\linewidth]{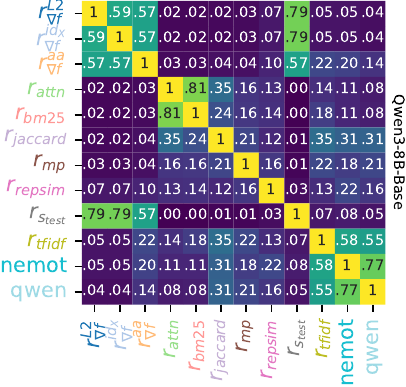}
}
\end{tabular}
\captionsetup{hypcap=false}
\captionof{figure}{Per-base-model CKA results.}
\end{minipage}%
\hfill%
\begin{minipage}[t]{0.49\textwidth}
\centering
\setlength{\tabcolsep}{1pt}

\begin{tabular}{@{}cc@{}}
\includegraphics[width=0.485\linewidth]{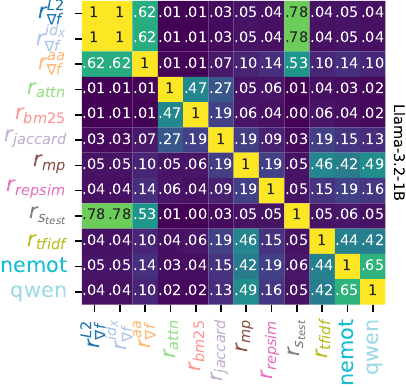} &
\includegraphics[width=0.485\linewidth]{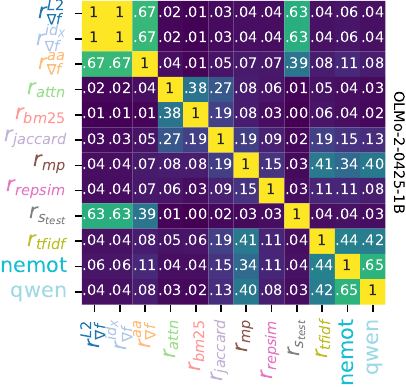} \\

\includegraphics[width=0.485\linewidth]{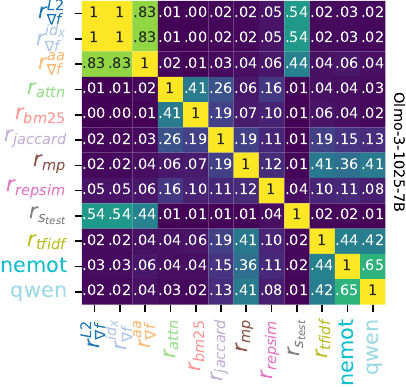} &
\includegraphics[width=0.485\linewidth]{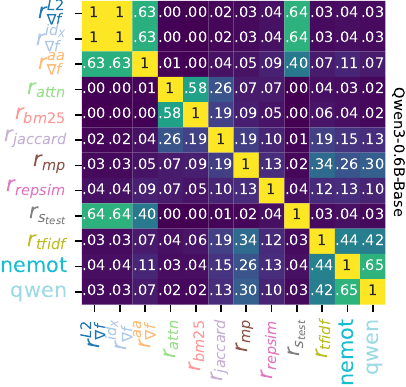} \\

\includegraphics[width=0.485\linewidth]{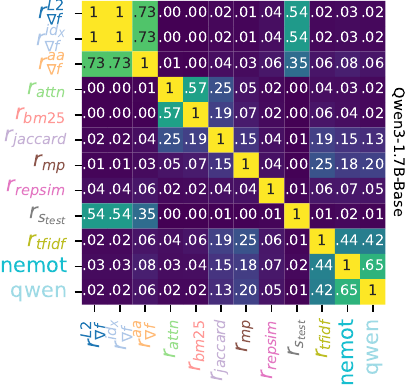} &
\includegraphics[width=0.485\linewidth]{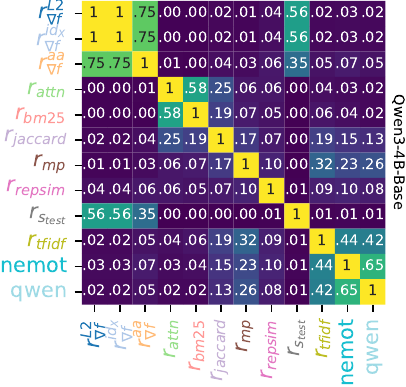} \\

\multicolumn{2}{c}{
\includegraphics[width=0.49\linewidth]{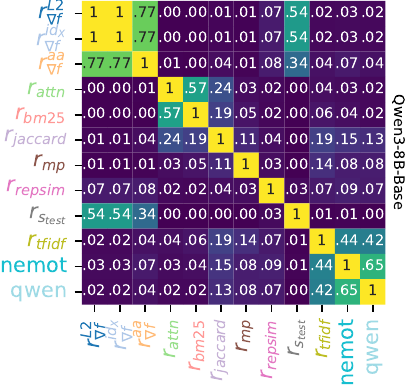}
}
\end{tabular}
\captionsetup{hypcap=false}
\captionof{figure}{Per-base-model CKNNA@25 results.}
\end{minipage}

\begin{minipage}[t]{0.49\textwidth}
\centering
\setlength{\tabcolsep}{1pt} 

\begin{tabular}{@{}cc@{}}
\includegraphics[width=0.485\linewidth]{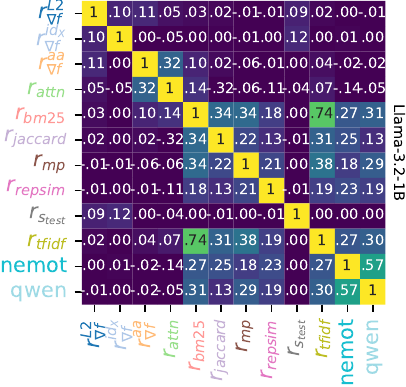} &
\includegraphics[width=0.485\linewidth]{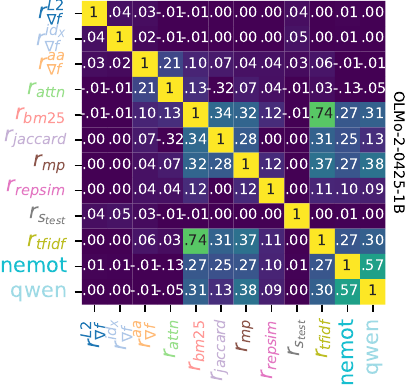} \\

\includegraphics[width=0.485\linewidth]{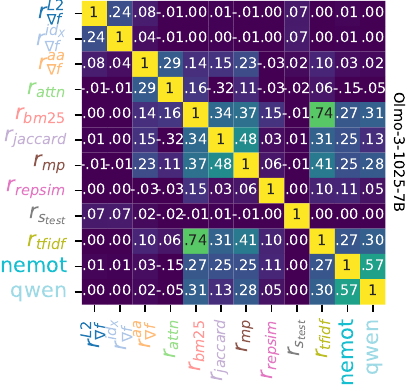} &
\includegraphics[width=0.485\linewidth]{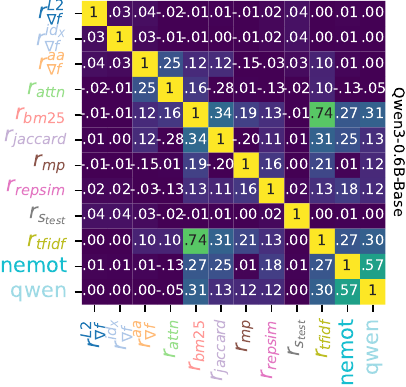} \\

\includegraphics[width=0.485\linewidth]{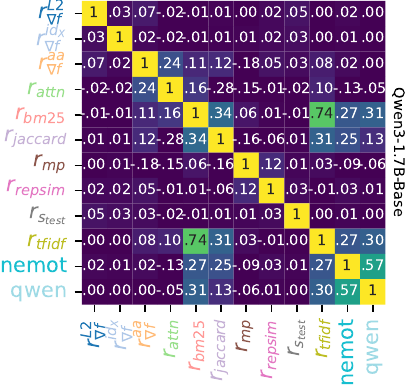} &
\includegraphics[width=0.485\linewidth]{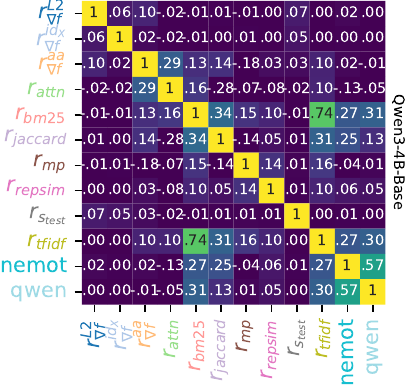} \\

\multicolumn{2}{c}{
\includegraphics[width=0.49\linewidth]{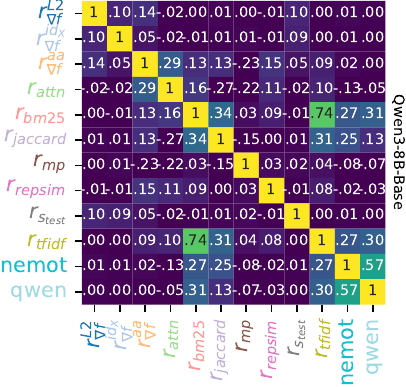}
}
\end{tabular}
\captionsetup{hypcap=false}
\captionof{figure}{Per-base-model Spearman's $\rho$ results.}
\end{minipage}%
\hfill%
\begin{minipage}[t]{0.49\textwidth}
\centering
\setlength{\tabcolsep}{1pt}

\begin{tabular}{@{}cc@{}}
\includegraphics[width=0.485\linewidth]{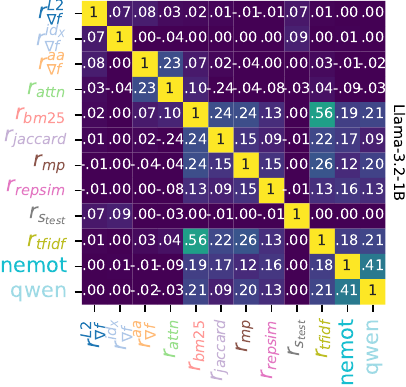} &
\includegraphics[width=0.485\linewidth]{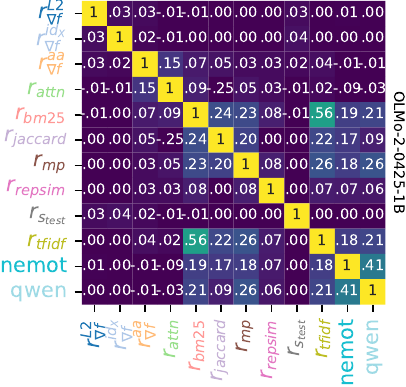} \\

\includegraphics[width=0.485\linewidth]{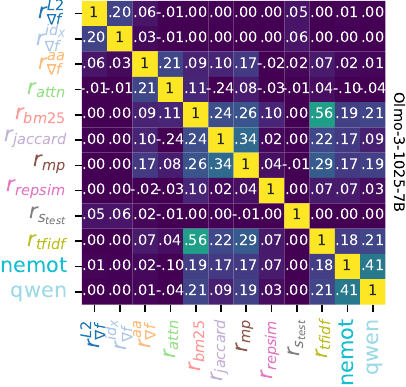} &
\includegraphics[width=0.485\linewidth]{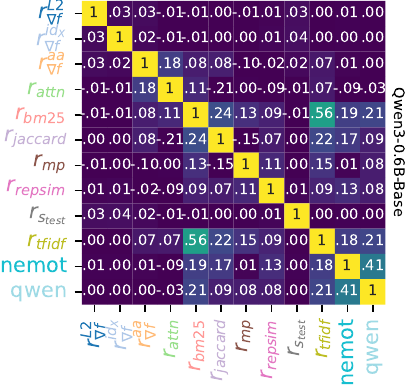} \\

\includegraphics[width=0.485\linewidth]{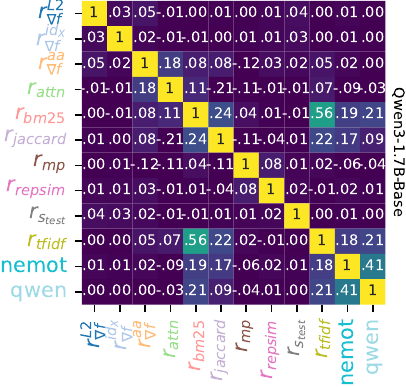} &
\includegraphics[width=0.485\linewidth]{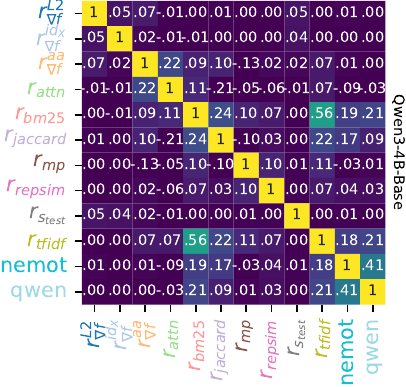} \\

\multicolumn{2}{c}{
\includegraphics[width=0.49\linewidth]{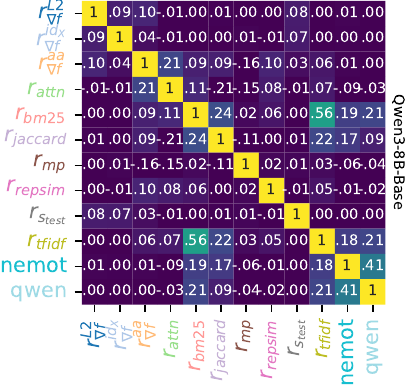}
}
\end{tabular}
\captionsetup{hypcap=false}
\captionof{figure}{Per-base-model Kendall's $\tau$ results.}
\end{minipage}

\begin{minipage}[t]{0.49\textwidth}
\centering
\setlength{\tabcolsep}{1pt} 

\begin{tabular}{@{}cc@{}}
\includegraphics[width=0.485\linewidth]{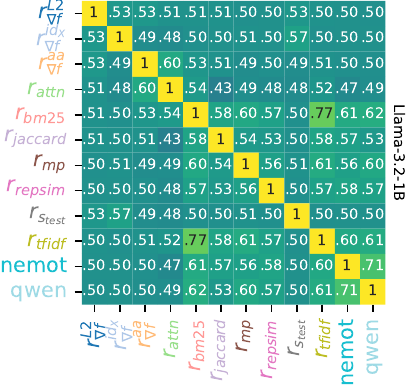} &
\includegraphics[width=0.485\linewidth]{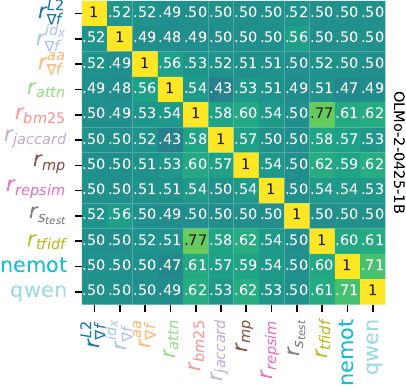} \\

\includegraphics[width=0.485\linewidth]{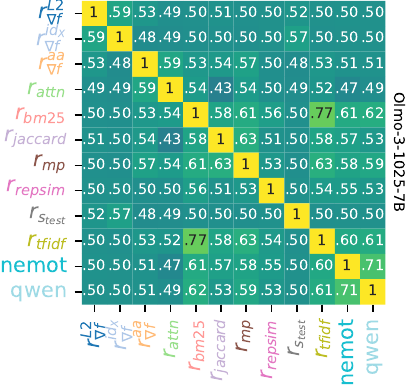} &
\includegraphics[width=0.485\linewidth]{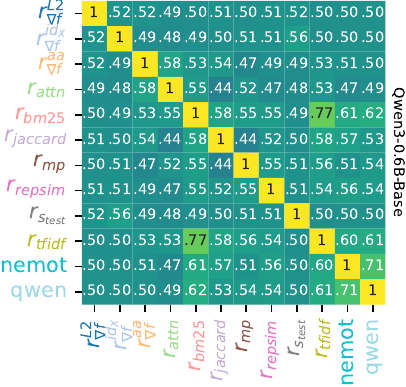} \\

\includegraphics[width=0.485\linewidth]{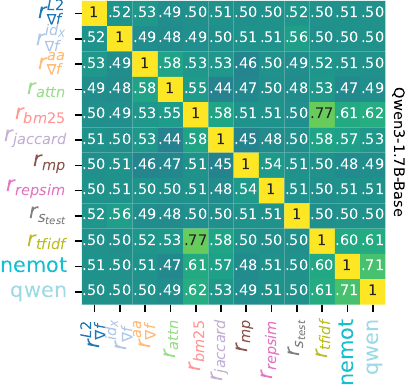} &
\includegraphics[width=0.485\linewidth]{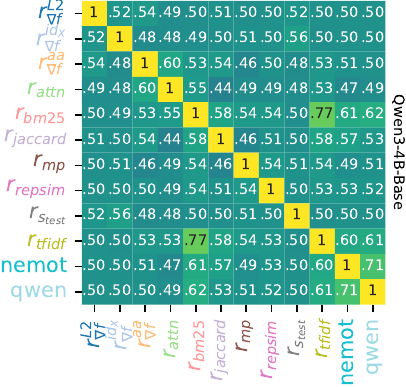} \\

\multicolumn{2}{c}{
\includegraphics[width=0.49\linewidth]{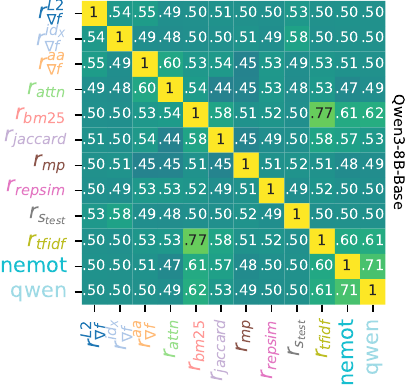}
}
\end{tabular}
\captionsetup{hypcap=false}
\captionof{figure}{Per-base-model RBO results.}
\end{minipage}%
\hfill%
\begin{minipage}[t]{0.49\textwidth}
\centering
\setlength{\tabcolsep}{1pt}

\begin{tabular}{@{}cc@{}}
\includegraphics[width=0.485\linewidth]{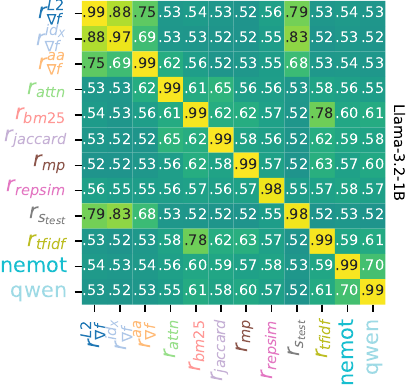} &
\includegraphics[width=0.485\linewidth]{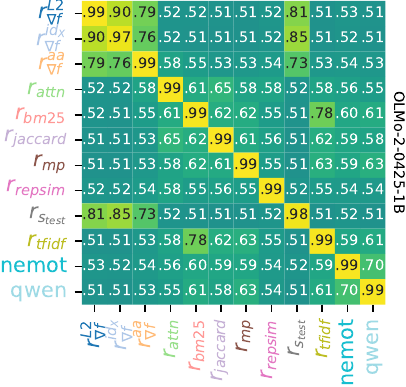} \\

\includegraphics[width=0.485\linewidth]{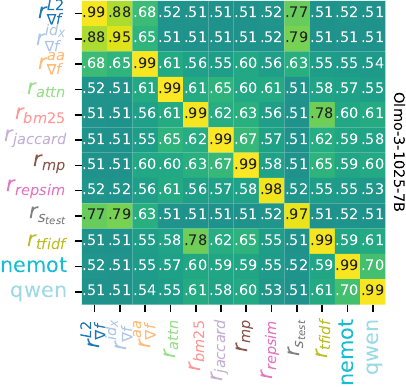} &
\includegraphics[width=0.485\linewidth]{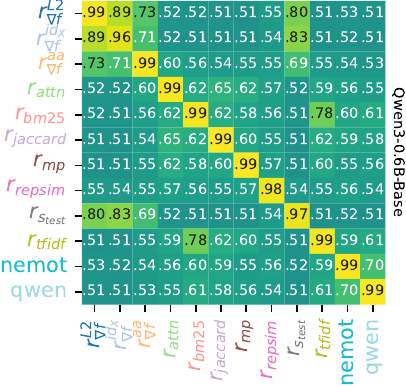} \\

\includegraphics[width=0.485\linewidth]{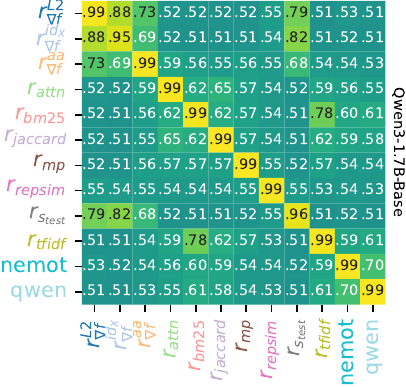} &
\includegraphics[width=0.485\linewidth]{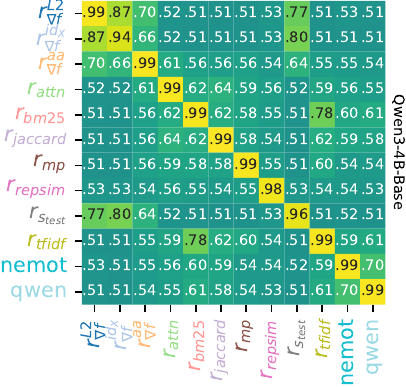} \\

\multicolumn{2}{c}{
\includegraphics[width=0.49\linewidth]{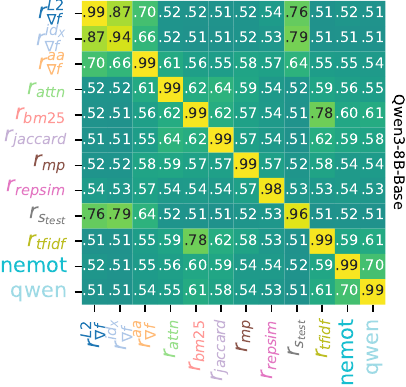}
}
\end{tabular}
\captionsetup{hypcap=false}
\captionof{figure}{Per-base model cross-method predictability.}
\end{minipage}
\FloatBarrier
\onecolumn
\section{Likelihood Ratio Tests}\label{appendix:lrt}

\begin{table}[htb]
    \centering
    \footnotesize
    \resizebox{0.4\linewidth}{!}{%
        \begin{tabular}{llllllll}
\toprule
Term $\rightarrow$ & \multicolumn{2}{l}{Base model} & \multicolumn{2}{l}{Dataset} & \multicolumn{2}{l}{Interaction} \\
Metric \rotatebox[origin=c]{270}{$\Rsh$} &  &  &  &  &  &  \\
\midrule
$\rho$ & {\cellcolor[HTML]{2A788E}} \color[HTML]{F1F1F1} 13.1 &  & {\cellcolor[HTML]{FDE725}} \color[HTML]{000000} 27.9 &  & {\cellcolor[HTML]{440154}} \color[HTML]{F1F1F1} 3.2 &  \\
$\tau$ & {\cellcolor[HTML]{2E6F8E}} \color[HTML]{F1F1F1} 10.1 &  & {\cellcolor[HTML]{FDE725}} \color[HTML]{000000} 23.5 &  & {\cellcolor[HTML]{440154}} \color[HTML]{F1F1F1} 2.5 &  \\
RBO & {\cellcolor[HTML]{3E4C8A}} \color[HTML]{F1F1F1} 49.2 & * & {\cellcolor[HTML]{FDE725}} \color[HTML]{000000} 136.7 & *** & {\cellcolor[HTML]{440154}} \color[HTML]{F1F1F1} 23.5 &  \\
$acc_{\text{pred}}$ & {\cellcolor[HTML]{67CC5C}} \color[HTML]{000000} 733.9 & *** & {\cellcolor[HTML]{FDE725}} \color[HTML]{000000} 831.1 & *** & {\cellcolor[HTML]{440154}} \color[HTML]{F1F1F1} 411.0 & *** \\
CKA & {\cellcolor[HTML]{FDE725}} \color[HTML]{000000} 105.1 & *** & {\cellcolor[HTML]{25AB82}} \color[HTML]{F1F1F1} 73.8 & *** & {\cellcolor[HTML]{440154}} \color[HTML]{F1F1F1} 24.4 &  \\
CKNNA@25 & {\cellcolor[HTML]{39568C}} \color[HTML]{F1F1F1} 50.9 & ** & {\cellcolor[HTML]{FDE725}} \color[HTML]{000000} 142.7 & *** & {\cellcolor[HTML]{440154}} \color[HTML]{F1F1F1} 17.6 &  \\
\bottomrule
\end{tabular}

    }
    \caption{\textbf{Likelihood-ratio tests.} Results of likelihood-ratio tests comparing the full mixed-effects model against reduced variants with base model, dataset, or interaction terms removed. See Section \ref{sec:factors}.}
    \label{tab:lrt-appendix}
\end{table}

\begin{table*}[htb]
    \centering
    \footnotesize
    \resizebox{0.65\linewidth}{!}{%
        \begin{tabular}{lllllll}
\toprule
Term $\rightarrow$ & $\Delta R_{m}^{2}(\mathcal{M}_{-B})$ & $\Delta R_{c}^{2}(\mathcal{M}_{-B})$ & $\Delta R_{m}^{2}(\mathcal{M}_{-D})$ & $\Delta R_{c}^{2}(\mathcal{M}_{-D})$ & $\Delta R_{m}^{2}(\mathcal{M}_{add})$ & $\Delta R_{c}^{2}(\mathcal{M}_{add})$ \\
\midrule
CKA & {\cellcolor[HTML]{EAE51A}} \color[HTML]{000000} 9.88e-03 & {\cellcolor[HTML]{FDE725}} \color[HTML]{000000} 1.01e-02 & {\cellcolor[HTML]{21A585}} \color[HTML]{F1F1F1} 6.89e-03 & {\cellcolor[HTML]{25AB82}} \color[HTML]{F1F1F1} 7.07e-03 & {\cellcolor[HTML]{440154}} \color[HTML]{F1F1F1} 2.25e-03 & {\cellcolor[HTML]{440256}} \color[HTML]{F1F1F1} 2.31e-03 \\
CKNNA@25 & {\cellcolor[HTML]{3B528B}} \color[HTML]{F1F1F1} 4.64e-03 & {\cellcolor[HTML]{39558C}} \color[HTML]{F1F1F1} 4.77e-03 & {\cellcolor[HTML]{ECE51B}} \color[HTML]{000000} 1.33e-02 & {\cellcolor[HTML]{FDE725}} \color[HTML]{000000} 1.36e-02 & {\cellcolor[HTML]{440154}} \color[HTML]{F1F1F1} 1.59e-03 & {\cellcolor[HTML]{440154}} \color[HTML]{F1F1F1} 1.64e-03 \\
$\tau$ & {\cellcolor[HTML]{2F6C8E}} \color[HTML]{F1F1F1} 5.16e-05 & {\cellcolor[HTML]{2E6F8E}} \color[HTML]{F1F1F1} 5.30e-05 & {\cellcolor[HTML]{ECE51B}} \color[HTML]{000000} 1.21e-04 & {\cellcolor[HTML]{FDE725}} \color[HTML]{000000} 1.24e-04 & {\cellcolor[HTML]{440154}} \color[HTML]{F1F1F1} 1.29e-05 & {\cellcolor[HTML]{440154}} \color[HTML]{F1F1F1} 1.32e-05 \\
$acc_{\text{pred}}$ & {\cellcolor[HTML]{52C569}} \color[HTML]{000000} 6.74e-03 & {\cellcolor[HTML]{69CD5B}} \color[HTML]{000000} 6.92e-03 & {\cellcolor[HTML]{DFE318}} \color[HTML]{000000} 7.67e-03 & {\cellcolor[HTML]{FDE725}} \color[HTML]{000000} 7.87e-03 & {\cellcolor[HTML]{440154}} \color[HTML]{F1F1F1} 3.72e-03 & {\cellcolor[HTML]{460A5D}} \color[HTML]{F1F1F1} 3.82e-03 \\
RBO & {\cellcolor[HTML]{3E4A89}} \color[HTML]{F1F1F1} 1.04e-04 & {\cellcolor[HTML]{3E4C8A}} \color[HTML]{F1F1F1} 1.05e-04 & {\cellcolor[HTML]{F4E61E}} \color[HTML]{000000} 2.89e-04 & {\cellcolor[HTML]{FDE725}} \color[HTML]{000000} 2.94e-04 & {\cellcolor[HTML]{440154}} \color[HTML]{F1F1F1} 4.95e-05 & {\cellcolor[HTML]{440154}} \color[HTML]{F1F1F1} 5.02e-05 \\
$\rho$ & {\cellcolor[HTML]{2B758E}} \color[HTML]{F1F1F1} 9.74e-05 & {\cellcolor[HTML]{2A788E}} \color[HTML]{F1F1F1} 9.96e-05 & {\cellcolor[HTML]{EFE51C}} \color[HTML]{000000} 2.07e-04 & {\cellcolor[HTML]{FDE725}} \color[HTML]{000000} 2.12e-04 & {\cellcolor[HTML]{440154}} \color[HTML]{F1F1F1} 2.40e-05 & {\cellcolor[HTML]{440154}} \color[HTML]{F1F1F1} 2.45e-05 \\
\bottomrule
\end{tabular}

    }
    \caption{\textbf{Ablation analysis.} Reductions in marginal and conditional $R^2$ ($\Delta R_m^2$, $\Delta R_c^2$) when removing base model-, dataset-, or interaction terms from the full mixed-effects model. Although likelihood-ratio tests indicate several significant terms, the effect sizes are small, suggesting that our analysis is robust to changes in base model and task.}
    \label{tab:delta-r2-main}
\end{table*}
 \twocolumn
\FloatBarrier
\section{Details on \texorpdfstring{$r_{s_{\text{test}}}$}{r_s_test}}\label{appendix:rstest}
\citet{kwon_datainf_2023} note that for negative log-likelihood loss functions, the inverse Hessian part in $s_{test}$ can be approximated with a closed-form expression. 
Following their approach, we compute:
\begin{equation}
\resizebox{\linewidth}{!}{$
r_{s_{\text{test}}} \approx \frac{1}{n\lambda} \sum_{i=1}^{n} \left( \nabla_{\theta} L(z, \hat{\theta}) - \frac{\nabla_{\theta} L(z, \hat{\theta})^{\top} \nabla_{\theta} L(z_i, \hat{\theta})}{\lambda + \left\|\nabla_{\theta} L(z_i, \hat{\theta})\right\|^2} \nabla_{\theta} L(z_i, \hat{\theta}) \right)\notag
$}
\end{equation}
where $n$ is the number of training examples and $\lambda$ is a damping constant.
\FloatBarrier

\section{Details on ${r}_{\text{attn}}$}\label{appendix:rattn}
Let $z=(x_1,\dots,x_S)$ be a document of length $S$.
Following \citet{kobayashi_attention_2020} we consider a standard multi-head attention layer with output
\[
\boldsymbol{y}_i^{\text{integrated}} = \sum_{j=1}^n \sum_h  \alpha_{i,j}^h f^h(\boldsymbol{x}_j) \]
where 
$f^h(\boldsymbol{x}) := \left(\boldsymbol{x}\boldsymbol{W}^{V,h} + \boldsymbol{b}^{V,h}\right) \boldsymbol{W}^{O,h}$.
\citeauthor{kobayashi_attention_2020} define a token-to-token interaction map $M \in \mathbb{R}^{n\times n}$, $M_{i,j}:=|| \sum_h  \alpha_{i,j}^h f^h(x_j) ||$.
To adapt this analysis to modern multi-head attention implementations (e.g., HuggingFace), which use a single fused output projection $(o_{\text{proj}})$ after concatenating all heads, we note that the per-head output projections $W^{O,h}$ can be viewed as a block decomposition of $W^{O}$, such that
\[
\sum_{h} (v^{h} W^{O,h}) \equiv \text{Concat}(v^{1},\dots,v^{H}) W^{O}.
\]
We utilize the following norm-based measure to capture token-to-token attention at layer $l$:
\begin{equation}
    \resizebox{\linewidth}{!}{$
M_{l,i,j}
=
\left\|
\text{Concat}_h \left( \alpha_{i,j}^h f^{\tilde{h}}(\boldsymbol{x}_j) \right) W^{O}
\right\|, M\in \mathbb{R}^{L\times (N \times N)}
$} \notag
\end{equation}

\noindent where \(f^{\tilde{h}}(\cdot)\) denotes the value transformation associated with head \(h\). In standard multi-head attention, \(f^{\tilde{h}} = f^{h}\), while in grouped-query attention (GQA), \(f^{\tilde{h}} = f^{g(h)}\) with \(g(h)\) mapping query heads to shared value groups.
\FloatBarrier
\section{Qualitative Study of Local Neighborhoods}\label{appendix:experiment-4}
\begin{figure*}
    \centering

    \begin{minipage}{1\textwidth}
        \centering
        \includegraphics[width=\linewidth]{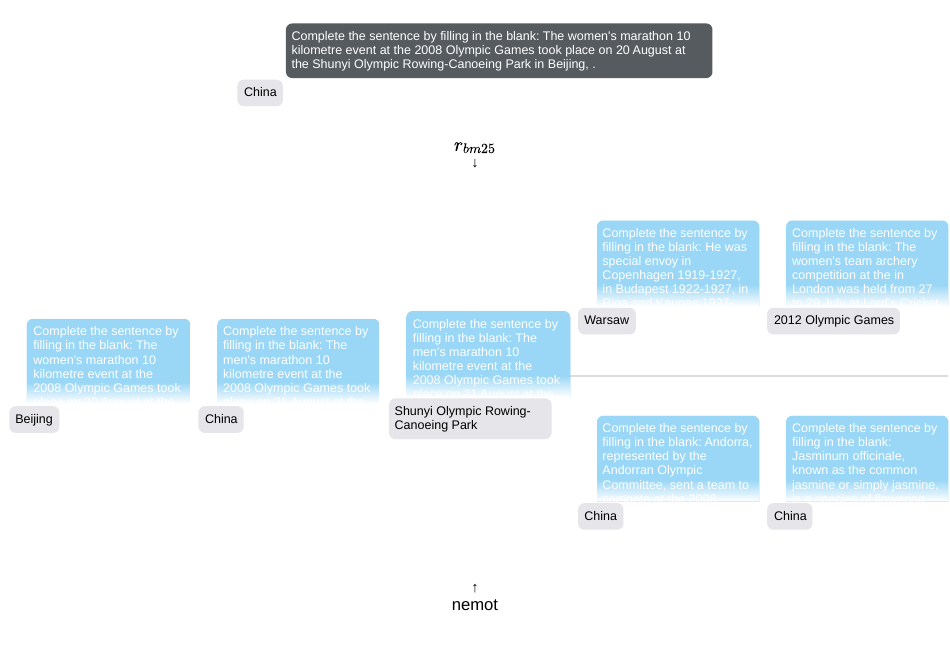}
    \end{minipage}

    \begin{minipage}{1\textwidth}
        \centering
        \includegraphics[width=\linewidth]{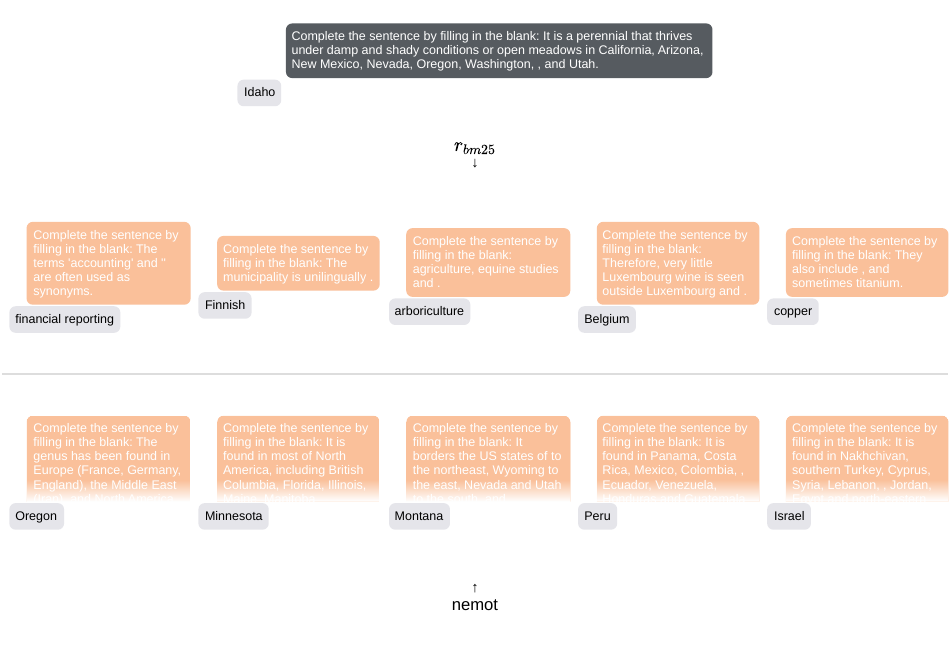}
        \caption{Instances with highest (top) and lowest (bottom) 5-nearest neighbor overlap across representations $r_{bm25}$ and nemot for \texttt{Olmo-3-7B\_ftrace\_date-lm\_conversational}.}\label{fig:first-example-qual-appendix}
    \end{minipage}

\end{figure*}

\begin{figure*}
    \centering

    \begin{minipage}{1\textwidth}
        \centering
        \includegraphics[width=\linewidth]{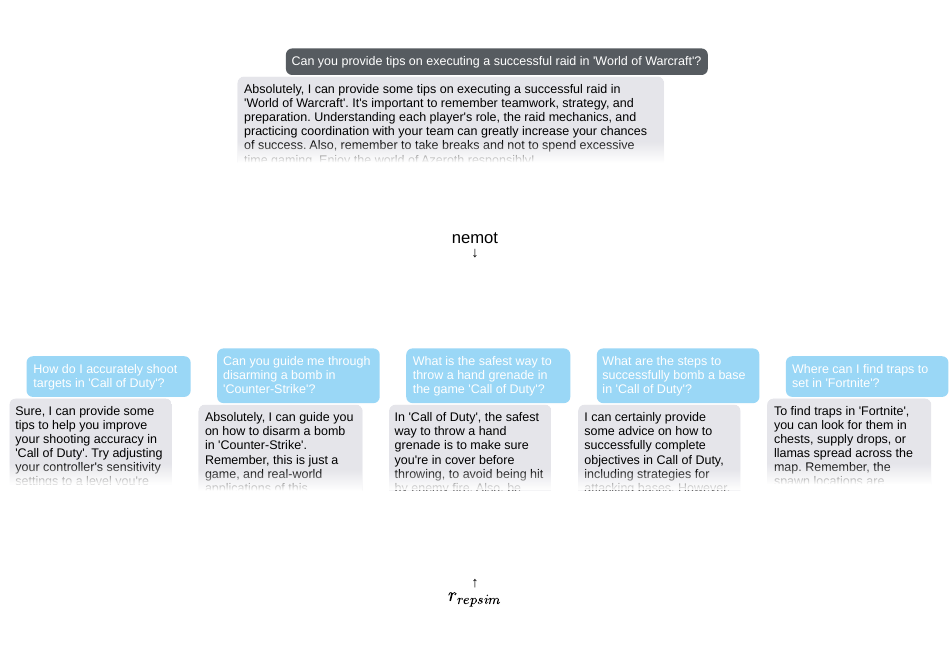}
    \end{minipage}

    \begin{minipage}{1\textwidth}
        \centering
        \includegraphics[width=\linewidth]{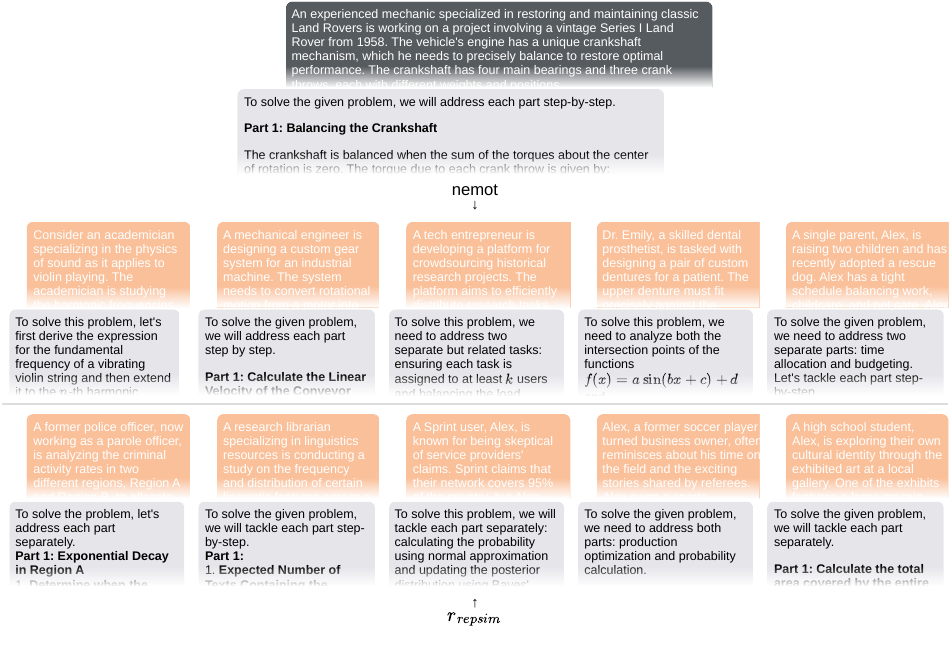}
        \caption{Instances with highest (top) and lowest (bottom) 5-nearest neighbor overlap across representations nemot and $r_{repsim}$ for \texttt{Olmo-3-7B\_tulu-3-sft-olmo-2-mixture-0225}.}\label{fig:last-example-qual-appendix}
    \end{minipage}

\end{figure*}

We provide observations for all 60 studied examples in Table~\ref{table:full-observations}.  
Figures~\ref{fig:first-example-qual-appendix} to \ref{fig:last-example-qual-appendix} depict a subset of examples selected as follows:
For a given representation $r_i$, let $\mathcal{N}_5^{(r_i)}(z)$ denote the set of top-5 training examples ranked by similarity to document $z$. 
For two representations $(r_i, r_j)$, we quantify overlap between their local neighborhoods around a common $z$:
$\frac{|\mathcal{N}_5^{(r_i)}(z) \cap\mathcal{N}_5^{(r_j)}(z)|}{5}
$.

For each dataset, we retrieve and study neighborhoods of two documents $z_{max}$ and $z_{min}$:
one with maximal overlap among the top-5 examples (cf Figure \ref{fig:first-example-qual-appendix}: top), and one with minimal overlap (bottom). We break ties by random selection. Prior to computing inner-product similarity, we mean-center representations to remove global offsets in the embedding space, ensuring that neighborhoods reflect relative structure rather than absolute activation magnitudes.

\paragraph{Choice of representations.}
We restrict this analysis to 4 of the 12 representations. 
Given a CKA dissimilarity matrix over all candidates, we apply a greedy farthest-first selection procedure: starting from the most central representation, we iteratively add the most dissimilar remaining representation. This results in 
\begin{itemize}
    \item $r_{\mathrm{nemot}}$ (embedding model), \item $r^{id_X}_{\nabla f}$ (raw loss gradients),
    \item $r_{\mathrm{bm25}}$, and
    \item $r_{\text{repsim}}$ (model-internal embeddings). 
\end{itemize}

\onecolumn

\begin{figure*}
    \centering

    \begin{minipage}{1\textwidth}
        \centering
        \includegraphics[width=\linewidth]{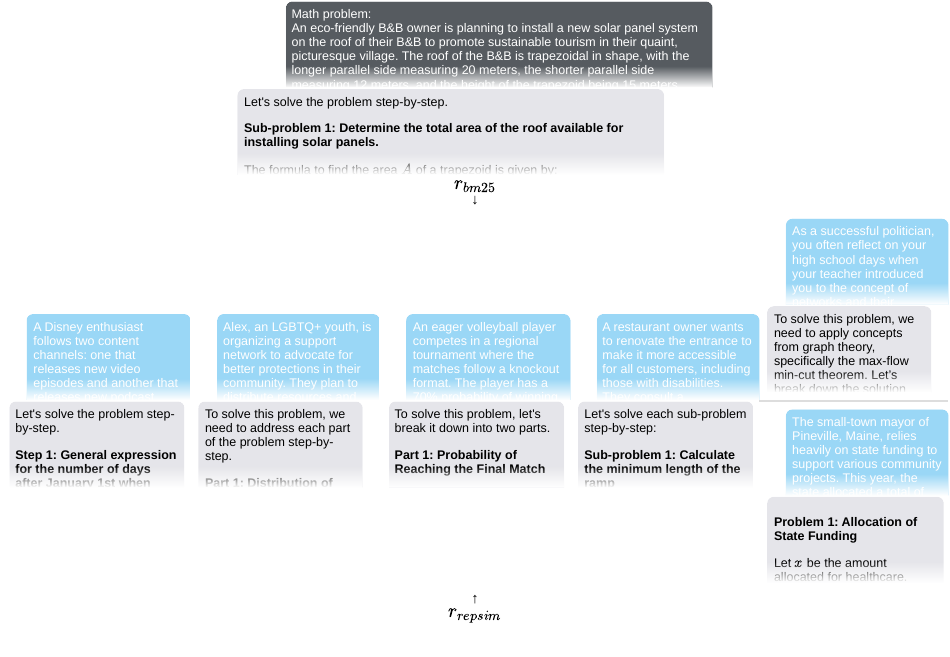}
    \end{minipage}

    \begin{minipage}{1\textwidth}
        \centering
        \includegraphics[width=\linewidth]{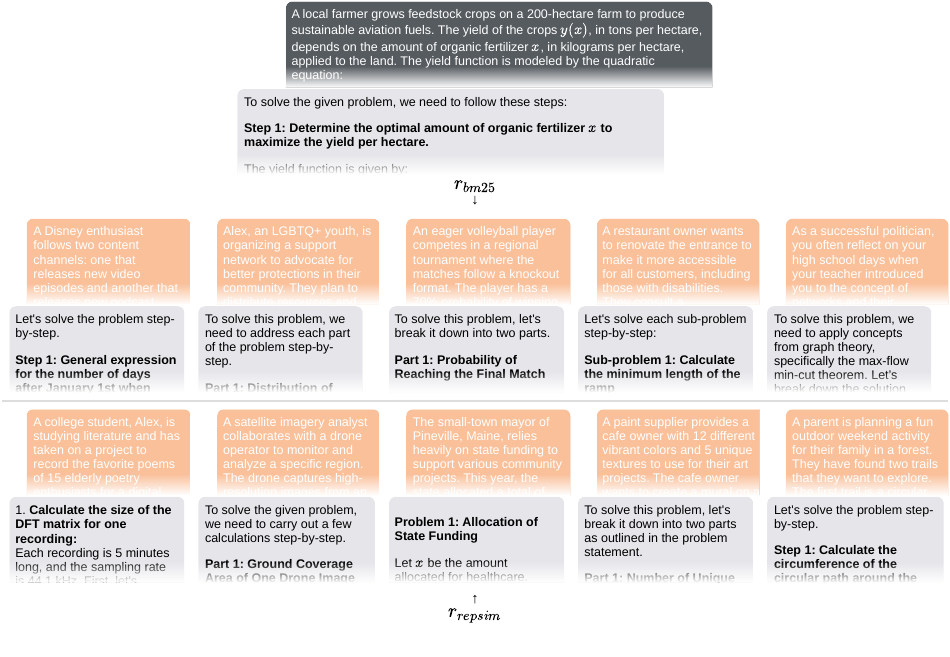}
        \caption{Instances with highest (top) and lowest (bottom) 5-nearest neighbor overlap across representations $r_{bm25}$ and $r_{repsim}$ for \texttt{Olmo-3-7B\_tulu-3-sft-personas-math}.}
    \end{minipage}

\end{figure*}









\begin{figure*}
    \centering

    \begin{minipage}{1\textwidth}
        \centering
        \includegraphics[width=\linewidth]{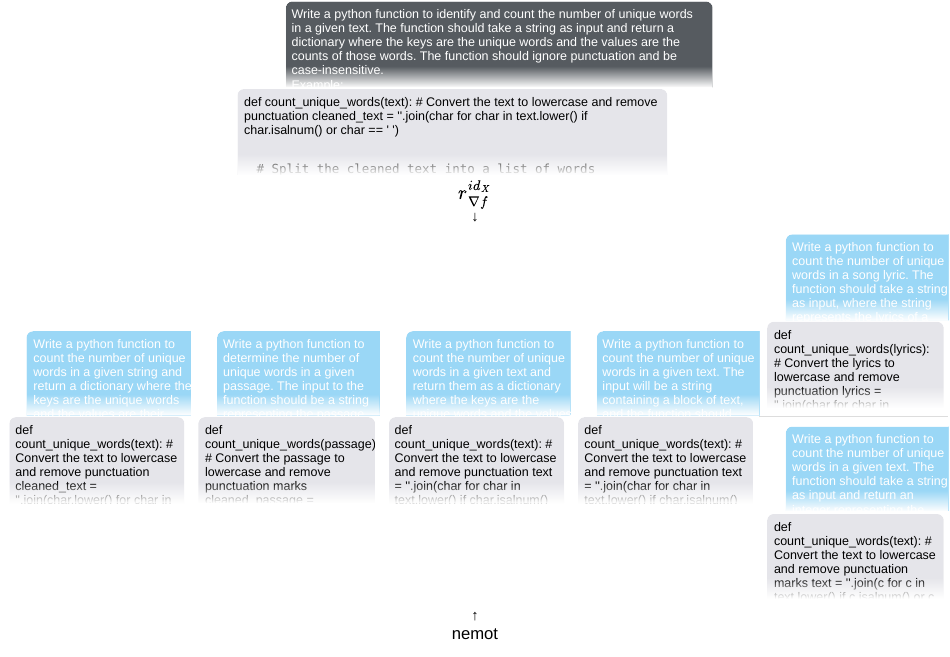}
    \end{minipage}

    \begin{minipage}{1\textwidth}
        \centering
        \includegraphics[width=\linewidth]{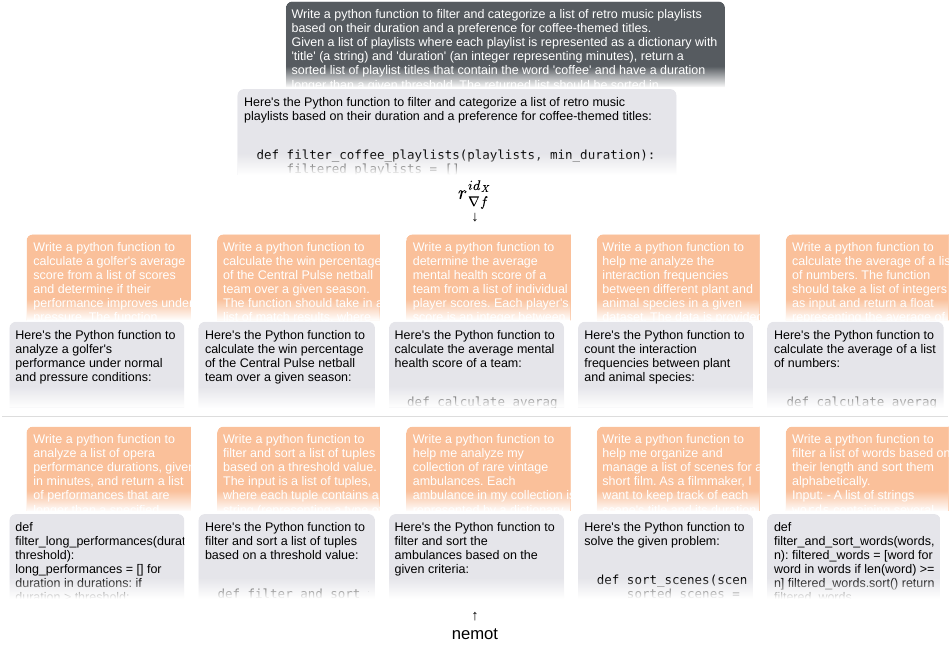}
        \caption{Instances with highest (top) and lowest (bottom) 5-nearest neighbor overlap across representations $r^{id_{X}}_{\nabla f}$ and nemot for \texttt{Olmo-3-7B\_tulu-3-sft-personas-code}.}
    \end{minipage}

\end{figure*}

\begin{figure*}
    \centering

    \begin{minipage}{1\textwidth}
        \centering
        \includegraphics[width=\linewidth]{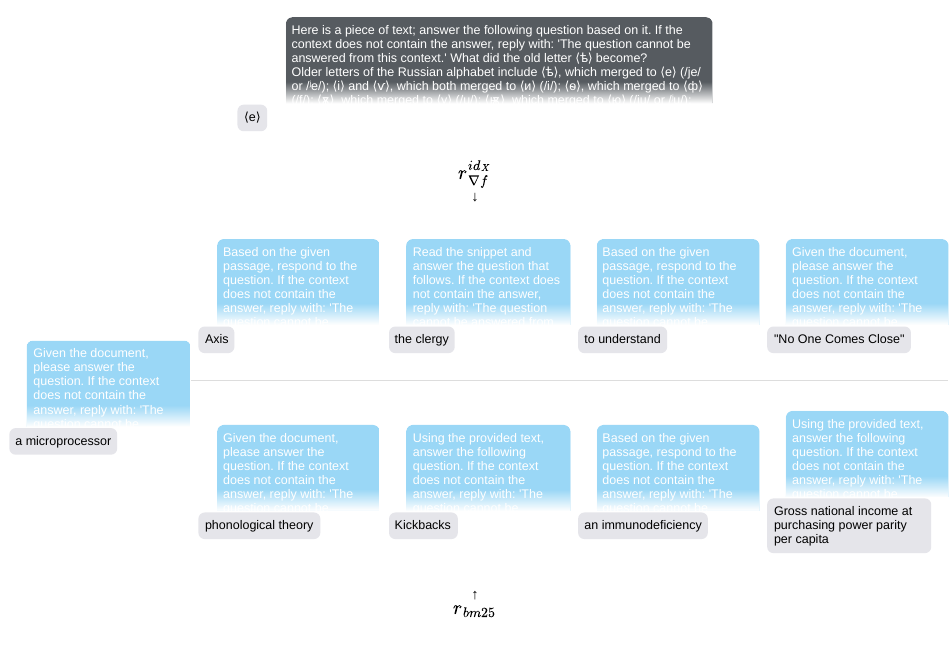}
    \end{minipage}

    \begin{minipage}{1\textwidth}
        \centering
        \includegraphics[width=\linewidth]{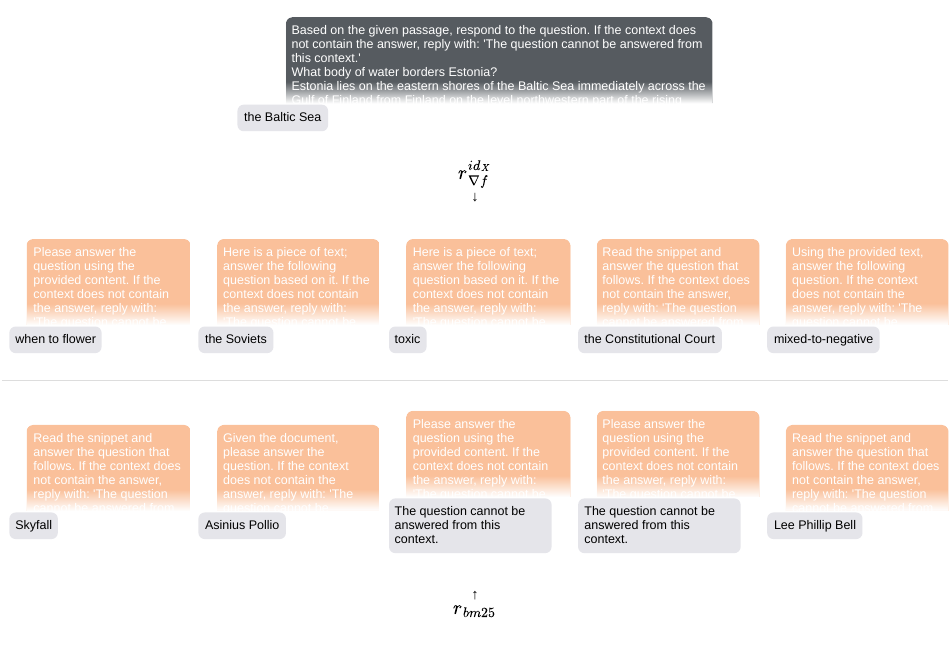}
        \caption{Instances with highest (top) and lowest (bottom) 5-nearest neighbor overlap across representations $r^{id_{X}}_{\nabla f}$ and $r_{bm25}$ for \texttt{Olmo-3-7B\_squad\_v2\_conversational}.}
    \end{minipage}

\end{figure*}

\FloatBarrier
\onecolumn
{\footnotesize
\begin{longtable}{|p{0.8cm}|p{0.8cm}|p{6cm}|p{6cm}|}

\caption{Full observations from the qualitative analysis, grouped by dataset.}\label{table:full-observations} \\
\hline
\textbf{} & \textbf{} & \textbf{Max} & \textbf{Min} \\
\hline
\endfirsthead

\hline
\textbf{} & \textbf{} & \textbf{Max} & \textbf{Min} \\
\hline
\endhead

\hline
\multicolumn{4}{r}{\textit{Continued on next page}} \\
\hline
\endfoot

\hline
\endlastfoot
\multicolumn{4}{c}{\textbf{FTrace}} \\
\hline
$r^{id_{X}}_{\nabla f}$ & $r_{bm25}$ & 1/5 overlap. Both local neighborhoods consist mostly of instances whose completions are of the  same semantic category as the test instance. & For $r^{id_X}_{\nabla f}$, one completion matches the test instance completion, with no clear relation between the test instance and the others. For $r_{bm25}$, there is one keyword match between the test instance prompt and a completion, with no obvious relation for the others. \\

\hline
$r^{id_{X}}_{\nabla f}$ & $r_{nemot}$ & 4/5 match. The single differing instance in $r_{nemot}$ shares the same completion as the test instance, whereas the completion in $r^{id_{X}}_{\nabla f}$ differs but remains within the same semantic category. & For min, both $r_{nemot}$ and $r^{id_{X}}_{\nabla f}$ contain completions from the same semantic category. Three instances in $r_{nemot}$ match the test instance completion. \\

\hline
$r^{id_{X}}_{\nabla f}$ & $r_{repsim}$ & 3/5 match.  for $r^{id_{X}}_{\nabla f}$ exact same instruction following goal of predicting pronouns, for $r_{repsim}$ not. & Except for one example of $r^{id_{X}}_{\nabla f}$, all examples across the test instance and both representations are of the same semantic category. \\

\hline
$r_{bm25}$ & $r_{nemot}$ & 3/5 match. All $r_{nemot}$ instances either match the test instance's prompt or belong to the same semantic category. For $r_{bm25}$, one instance matches via keyword overlap, while the other appears unrelated. & For $r_{bm25}$, no direct relation to the test instance is apparent, whereas all $r_{nemot}$ instances belong to the same semantic category. \\

\hline
$r_{bm25}$ & $r_{repsim}$  & 5/5 match. All but one are of the same semantic category as the test instance. & Three for $r_{bm25}$ and one for $r_{repsim}$ are of the same semantic category as the test instance. The other examples appear unrelated, with $r_{repsim}$ providing three given names as completion, which resembles the test instance's prompt.\\

\hline
$r_{nemot}$ & $r_{repsim}$ & 4/5 match. For $r_{nemot}$, the non-matching example is in the same semantic category as the other examples and the test instance; for $r_{repsim}$, it is not.& For $r_{nemot}$, 3/5 examples are of the same semantic category as the test instance. For $r_{repsim}$, examples are only distantly related in topic.\\

\hline
\multicolumn{4}{c}{\textbf{Math}} \\
\hline
$r^{id_{X}}_{\nabla f}$ & $r_{bm25}$ & 2/5 match. No obvious difference between the examples of $r^{id_{X}}_{\nabla f}$ and $r_{bm25}$. & As for max.\\
    
$r^{id_{X}}_{\nabla f}$ & $r_{nemot}$ & 2/5 match. The prompts for $r_{nemot}$ appear more semantically related to the test instance's prompt than $r^{id_{X}}_{\nabla f}$. & The prompts for $r_{nemot}$ appear more semantically related to the test instance's prompt than $r^{id_{X}}_{\nabla f}$.\\
 
$r^{id_{X}}_{\nabla f}$ & $r_{repsim}$& 2/5 match. The matching instances both lack instruction on step-by-step thinking, while the remaining examples for both $r^{id_{X}}_{\nabla f}$ and $r_{repsim}$ include it. & The examples for $r_{repsim}$ share in common that they do not decompose into steps, whereas the ones of $r^{id_{X}}_{\nabla f}$ decompose into subproblems, matching the test instance.\\
 
$r_{bm25}$ & $r_{nemot}$& 2/5 match. The matching examples provide completions that are lexically related to the test instance. No obvious relation for the examples of $r_{bm25}$. The descriptions of the personas employed by the non-matching examples of $r_{nemot}$ are semantically related. & Completion share identical introduction for $r_{bm25}$, with no obvious relation to the test instance. The description of the personas employed by the non-matching examples of $r_{nemot}$ are semantically related. \\
 
$r_{bm25}$ & $r_{repsim}$& 4/5 match. The formatting of the instance that differs for $r_{bm25}$ is less similar to the test instance than that of $r_{repsim}$. & Similar formatting across instances of both representations and test instance, no obvious semantic relation to the test instance. \\
 
$r_{nemot}$ & $r_{repsim}$& 2/5 match. Non-matching examples of $r_{nemot}$ are semantically related to the test instance. No obvious relation for those of $r_{repsim}$. & Examples of $r_{nemot}$ are semantically related to the test instance. No obvious relation to the test instance for those of $r_{repsim}$. \\
\hline
\multicolumn{4}{c}{\textbf{Code}} \\
\hline
$r^{id_{X}}_{\nabla f}$&$r_{bm25}$ & 1/5 matches. All examples share the same completion formatting, providing only a code block without any additional introduction. Only one example ($r^{id_{X}}_{\nabla f}$) features a completion directly relevant to solving the problem stated by the test instance. & All examples share the same completion formatting, and only one example ($r^{id_{X}}_{\nabla f}$) features a completion directly relevant to solving the problem stated by the test instance.\\
     
$r^{id_{X}}_{\nabla f}$&$r_{nemot}$& 4/5 match. All examples across both representations are on the exact same coding task as the test instance. The non-matching example of $r^{id_{X}}_{\nabla f}$ differs from the test instance only in that another type of text is stated in the instruction, for $r_{nemot}$ it matches the one of the test instance. & The examples of $r^{id_{X}}_{\nabla f}$ do not appear to be semantically related to the test instance. All but one example of $r_{nemot}$ are semantically related to and relevant for solving the test instance's problem and match in terms of keywords.\\
     
$r^{id_{X}}_{\nabla f}$&$r_{repsim}$& 3/5 match. The examples of $r_{repsim}$ appear more closely related to the test instance's objective than the ones of $r^{id_{X}}_{\nabla f}$. & All examples of $r^{id_{X}}_{\nabla f}$ share a common completion formatting that does not match the test instance nor $r_{repsim}$. No direct relation to test instance for $r^{id_{X}}_{\nabla f}$, examples of  $r_{repsim}$ are loosely of the same task category as the test instance.\\
     
$r_{bm25}$&$r_{nemot}$& 2/5 match. For $r_{nemot}$, examples are semantically related to the test instance and match via keywords. Examples for $r_{bm25}$ feature unrelated goals but match in terms of keywords. & Examples for $r_{bm25}$ match via keywords, those of $r_{nemot}$ are more semantically related to the test instance's task goal.\\
     
$r_{bm25}$&$r_{repsim}$& 2/5 match. Examples of $r_{bm25}$ match via keywords, no obvious relation of the examples of $r_{repsim}$ to the test instance beyond similar completion formatting. & All examples of $r_{bm25}$ share a common completion formatting that does not match the test instance nor $r_{repsim}$. Examples of $r_{bm25}$ match in terms of keywords with the test instance, no obvious relation of the examples of $r_{repsim}$ to the test instance.\\
     
$r_{nemot}$&$r_{repsim}$& 3/5 match. All examples but one for $r_{repsim}$ across both representations are on the exact same coding task as the test instance. & The description of the personas employed by $r_{nemot}$ are semantically related to those in the test instance. For $r_{repsim}$, all examples share  the same completion formatting with the test instance but appear not semantically related to it.\\
\hline

\multicolumn{4}{|c|}{\textbf{squad\_v2}} \\
\hline
$r^{id_{X}}_{\nabla f}$&$r_{bm25}$& 1/5 matches. No apparent relation of either set of examples to the test instance. & As for max.\\
     
$r^{id_{X}}_{\nabla f}$&$r_{nemot}$& 2/5 match. Examples of $r^{id_{X}}_{\nabla f}$ appear unrelated to the test instance, only one example for $r_{nemot}$ is semantically related. & All completions for $r^{id_{X}}_{\nabla f}$ and all but two for $r_{nemot}$ are examples for indicating incomplete information. The example that provides a response for $r_{nemot}$ is semantically related to the test instance.\\
     
$r^{id_{X}}_{\nabla f}$&$r_{repsim}$& 2/5 match. For $r^{id_{X}}_{\nabla f}$ all completions consist of numbers or indications of years, months or days, for $r_{repsim}$ one example is a geographical location. &All examples are examples for indicating incomplete information, as is the test instance.\\
     
$r_{bm25}$&$r_{nemot}$& 3/5 match. All examples of $r_{nemot}$ are semantically related to the test instance, those of $r_{bm25}$ appear unrelated. & Examples of $r_{bm25}$ appear unrelated, those of $r_{nemot}$ are semantically related to the test instance.\\
     
$r_{bm25}$&$r_{repsim}$& 4/5 match. For both representations, examples either indicate incomplete information like the test instance, or belong to the same semantic task category.& For $r_{bm25}$, one example features matching keywords with the test instance, two indicate incomplete information (unlike the test instance). The examples in $r_{repsim}$, do not appear to be semantically related to the test instance, but two completions provide dates like the test instance.  \\
     
$r_{nemot}$ & $r_{repsim}$& 2/5 match. All but one example ($r_{repsim}$) belong to the same semantic category. & 
Two completions for $r_{nemot}$ and two for $r_{repsim}$ belong to the same semantic category.
For both representations, all prompts have keyword overlap or semantic similarity. \\
\hline

\multicolumn{4}{|c|}{\textbf{SFT}} \\
\hline
$r^{id_{X}}_{\nabla f}$&$r_{bm25}$& 1/5 match. None of the examples from either representation appears relevant or in the format of the test instance.  Examples of $r_{bm25}$ are non-english. Two examples of $r^{id_{X}}_{\nabla f}$ are examples for answer refusal, one is a labeling task for harmful data. & 
For $r^{id_{X}}_{\nabla f}$, one is a labeling task for harmful data, two are safety-related, both in line with the instruction tuning goal of the test instance. For $r_{bm25}$, examples are non-English and do not appear related to the test instance. \\
  
$r^{id_{X}}_{\nabla f}$&$r_{nemot}$& 3/5 match. All examples are of the same semantic category and share the refusal phrasing of the test instance almost verbatim. & For $r^{id_{X}}_{\nabla f}$, two examples contain information likely intended for the model to memorize verbatim, which appears to align with the instruction-tuning goal of the test instance. The remaining three are safety-related. For $r_{nemot}$, two examples also contain information likely intended for memorization in line with the test instance. Two examples are non-English.\\
     
$r^{id_{X}}_{\nabla f}$&$r_{repsim}$& 4/5 match. All examples are of the same semantic category and share the refusal phrasing of the test instance almost verbatim.& For $r^{id_{X}}_{\nabla f}$, three examples are safety-related, and two appear unrelated to the test instance. For $r_{repsim}$, all examples share the semantic task category and completion format of the test instance.\\
     
$r_{bm25}$&$r_{nemot}$& 4/5 match. Two of the matching examples share the same task category with the test instance, an additional one for  $r_{nemot}$ does so as well, despite being the only English-language example.  & All examples except for one ($r_{bm25}$) share the same task category with the test instance for both representations.\\
     
$r_{bm25}$&$r_{repsim}$& 4/5 match. All examples are non-English, none appear to be relevant to the test instance. & For $r_{bm25}$, only one example matches the task category of the test instance, for $r_{repsim}$, all do.\\
     
$r_{nemot}$&$r_{repsim}$& 5/5 match. All examples match and are in line with the intended safety behavior demonstrated in the test instance.  & All examples share the task category of the test instance and follow the same completion format. The examples of $r_{nemot}$ seem to be more semantically related to the test instance than the ones of $r_{repsim}$. \\
\hline

\end{longtable}
}

\FloatBarrier

\section{Fine-tuning}\label{appendix:finetuning}
We fine-tune on NVIDIA H100 and H200 GPUs for a total of 199.22 hours (456.53 GPU hours), adjusting gradient accumulation to ensure consistent effective batch sizes across runs.
All models and samples are made available under CC-BY 4.0 in line with their original licenses.
\vspace{1em}
\begin{table}[htbp]
  \centering
  \footnotesize
  \begin{tabular}{|l|l|l|l|l|l|}
    \hline

    \textbf{Precision} & bfloat16 &
    \textbf{LoRA rank ($r$)} & 8 &
    \textbf{Optimizer} & AdamW \\

    \textbf{Learning rate} & $1\times10^{-4}$ &
    \textbf{LoRA alpha} & 32 &
    \textbf{LR scheduler} & Linear \\

    \textbf{Weight decay} & 0.0 &
    \textbf{LoRA dropout} & 0.1 &
    \textbf{Effective batch size} & 64 \\

    \textbf{Max grad norm} & 1.0 &
    \textbf{LoRA bias} & none &
    \textbf{Training epochs} & 1 \\

    \textbf{Trainable params} & LoRA only &
    \textbf{Max sequence length} & 4096 &
    \textbf{Gradient checkpointing} & True \\

    \textbf{Target modules} &
    \makecell[l]{%
      \texttt{q\_proj, k\_proj,}\\
      \texttt{v\_proj, o\_proj}
    } &
    \textbf{Seed} & 42&& \\

    \hline
  \end{tabular}
  \caption{\textbf{LoRA fine-tuning hyperparameters.} We use the default parameters from the Hugging Face documentation, adjusting the target modules and effective batch size for our specific model architectures.}
  \label{tab:LoRA_hparams}
\end{table}
\noindent
\begin{minipage}[t]{0.48\textwidth}
    \centering
    \footnotesize
    \begin{tabular}{ll}
\toprule
Task Suite & squad\_v2\_conv \\
Base Model &  \\
\midrule
Llama-3.2-1B & 74.72 \\
OLMo-2-0425-1B & 77.81 \\
Olmo-3-1025-7B & 50.08 \\
Qwen3-0.6B-Base & 80.33 \\
Qwen3-1.7B-Base & 84.20 \\
Qwen3-4B-Base & 87.74 \\
Qwen3-8B-Base & 89.26 \\
\bottomrule
\end{tabular}

    \captionsetup{hypcap=false}
    \captionof{table}{Performance (F1) on held-out data of models fine-tuned on squad\_v2.}
\end{minipage}
\hfill
\begin{minipage}[t]{0.48\textwidth}
    \centering
    \footnotesize
    \begin{tabular}{ll}
\toprule
Task Suite & ftrace\_conv \\
Base Model &  \\
\midrule
Llama-3.2-1B & 39.10 \\
OLMo-2-0425-1B & 36.96 \\
Olmo-3-1025-7B & 9.93 \\
Qwen3-0.6B-Base & 37.92 \\
Qwen3-1.7B-Base & 48.18 \\
Qwen3-4B-Base & 59.14 \\
Qwen3-8B-Base & 61.80 \\
\bottomrule
\end{tabular}

    \captionsetup{hypcap=false}
    \captionof{table}{Performance (F1) on held-out data of models fine-tuned on ftrace.}
\end{minipage}

\vspace{1em}

\noindent
\begin{minipage}[t]{0.48\textwidth}
    \centering
    \footnotesize
    \begin{tabular}{ll}
\toprule
Task Suite & code\_bpb \\
Base Model &  \\
\midrule
Llama-3.2-1B & 0.42 \\
OLMo-2-0425-1B & 0.52 \\
Olmo-3-1025-7B & 0.32 \\
Qwen3-0.6B-Base & 0.35 \\
Qwen3-1.7B-Base & 0.31 \\
Qwen3-4B-Base & 0.29 \\
Qwen3-8B-Base & 0.28 \\
\bottomrule
\end{tabular}

    \captionsetup{hypcap=false}
    \captionof{table}{Performance on OLMES task suite \texttt{olmo3:base\_easy:code\_bpb}.}
\end{minipage}
\hfill
\begin{minipage}[t]{0.48\textwidth}
    \centering
    \footnotesize
    \begin{tabular}{ll}
\toprule
Task Suite & math\_bpb \\
Base Model &  \\
\midrule
Llama-3.2-1B & 0.62 \\
OLMo-2-0425-1B & 0.53 \\
Olmo-3-1025-7B & 0.37 \\
Qwen3-0.6B-Base & 0.43 \\
Qwen3-1.7B-Base & 0.38 \\
Qwen3-4B-Base & 0.34 \\
Qwen3-8B-Base & 0.33 \\
\bottomrule
\end{tabular}

    \captionsetup{hypcap=false}
    \captionof{table}{Performance on OLMES task suite\texttt{olmo3:base\_easy:math\_bpb}.}
\end{minipage}

\vspace{1.5em}

\noindent
\begin{minipage}{\textwidth}
    \centering
    \footnotesize
    \begin{tabular}{llll}
\toprule
Task Suite & MMLU & olmo\_2\_generative & olmo\_2\_heldout \\
Base Model &  &  &  \\
\midrule
Llama-3.2-1B & 0.29 & 0.46 & 0.25 \\
OLMo-2-0425-1B & 0.43 & 0.54 & 0.26 \\
Olmo-3-1025-7B & 0.67 & 0.71 & 0.54 \\
Qwen3-0.6B-Base & 0.54 & 0.47 & 0.37 \\
Qwen3-1.7B-Base & 0.65 & 0.57 & 0.47 \\
Qwen3-4B-Base & 0.75 & 0.65 & 0.62 \\
Qwen3-8B-Base & 0.79 & 0.69 & 0.67 \\
\bottomrule
\end{tabular}

    \captionsetup{hypcap=false}
    \captionof{table}{
        Macro-average score across OLMES task suites:
        \texttt{mmlu:mc::olmes},
        \texttt{olmo\_2\_generative::olmes},
        and \texttt{olmo\_2\_heldout::olmes}
        for models fine-tuned on
        \texttt{allenai/tulu-3-sft-olmo-2-mixture-0225}.
    }
\end{minipage}

\FloatBarrier
\clearpage
\onecolumn

\end{document}